\documentclass[12pt,letterpaper]{article}
\usepackage[a4paper, total={7in, 10in}]{geometry}

\usepackage{helvet}
\usepackage{authblk}
\usepackage{hyperref}    
\usepackage[left]{lineno}
\usepackage{amsmath}  
\usepackage{amsfonts}    
\usepackage{graphicx} 
\usepackage{booktabs} 
\usepackage{enumitem}
\usepackage{amssymb}
\usepackage[utf8]{inputenc}  
\usepackage{threeparttable}

\usepackage[final,tracking=true,kerning=true,spacing=true]{microtype} 

\makeatletter 
\def\tagform@#1{\maketag@@@{(\ignorespaces Equation~#1\unskip)}}
\makeatother
\renewcommand{\eqref}[1]{Equation~\ref{#1}} 
\usepackage{tabularx}
\usepackage{booktabs}

\usepackage{array}
\newcolumntype{R}{>{\raggedleft\arraybackslash}X}
\newcolumntype{L}{>{\raggedright\arraybackslash}X}
\usepackage{threeparttable}

\usepackage{tcolorbox}
\tcbuselibrary{skins}

\newtcolorbox{codebox}[1][]{
  colback=gray!5!white,
  colframe=gray!70!black,
  width=0.9\textwidth,
  center,
  arc=2mm,
  boxrule=0.5pt,
  left=8mm,
  fontupper=\small\ttfamily,
  title={#1}
}

\usepackage{appendix}  

\makeatletter
\renewcommand{\maketitle}{\bgroup\setlength{\parindent}{0pt}
\begin{flushleft}
  \textbf{\@title}
  
  \@author
\end{flushleft}\egroup}
\makeatother

\title{Auditable Unit‑Aware Thresholds in Symbolic Regression via Logistic‑Gated Operators}

\date{}

\author[1*]{Ou Deng}
\author[2]{Ruichen Cong}
\author[1]{Jianting Xu}
\author[2]{Shoji Nishimura}
\author[2]{Atsushi Ogihara}
\author[2*]{Qun Jin}

\affil[1]{Graduate School of Human Sciences, Waseda University.
Tokorozawa, Saitama 359-1192, Japan}
\affil[2]{Faculty of Human Sciences, Waseda University. Tokorozawa, Saitama 359-1192, Japan}
\affil[*]{Correspondence: dengou@toki.waseda.jp; jin@waseda.jp}

\usepackage[super,comma,sort&compress]{natbib}

\begin{document}

\maketitle

\section*{Bigger Picture}
AI for health will only scale when models are not only accurate but also {readable}, {auditable}, and {governable}. Many clinical and public‑health decisions hinge on numeric thresholds---cut‑points that trigger alarms, treatment, or follow‑up---yet most machine‑learning systems bury those thresholds inside opaque scores or smooth response curves. We introduce {logistic‑gated operators} (LGO) for symbolic regression, which promote thresholds to first‑class, unit‑aware parameters inside equations and map them back to physical units for direct comparison with guidelines. On public ICU and population‑health cohorts (MIMIC‑IV ICU, eICU, NHANES), LGO recovers clinically plausible gates on MAP, lactate, GCS, SpO\textsubscript{2}, BMI, fasting glucose, and waist circumference while remaining competitive with established scoring systems (AutoScore) and explainable boosting machines (EBM). The gates are sparse and selective: they appear when regime switching is supported by the data and are pruned on predominantly smooth tasks, yielding compact formulas that clinicians can inspect, stress‑test, and revise. As a standalone symbolic model or a safety overlay on black‑box systems, LGO helps translate observational data into auditable, unit‑aware rules for medicine and other threshold‑driven domains.


\section*{Highlights}
\begin{itemize}
\item Unit-aware thresholds as learnable primitives in symbolic expressions 
\item LGO matches AutoScore AUROC while providing explicit gate structure 
\item Hard gates concentrate on clinical variables; soft fallback when smooth 
\item Reproducible pipeline with threshold-audit tooling and Wilcoxon tests 
\end{itemize}

\section*{Summary}
Symbolic regression promises readable equations but struggles to represent unit-aware thresholds that drive clinical decisions. We propose logistic-gated operators (LGO)---differentiable gates with learnable location and steepness---embedded as typed primitives in a genetic-programming SR engine. Training is performed in standardized space; thresholds are mapped back to physical units for audit.

On three public health datasets (MIMIC-IV ICU, eICU, NHANES), LGO\textsubscript{hard} recovers clinically plausible cut-points: 4/13 within 10\% of guideline anchors, 7/13 within 20\%; remaining gates shift toward extreme-risk or early-warning regimes. Compared with AutoScore, LGO matches or exceeds AUROC and calibration while providing explicit thresholds instead of point tables. On smooth UCI benchmarks, gates are pruned, preserving parsimony. The result is compact symbolic equations with auditable, unit-aware thresholds for clinical decision support.

\section*{Keywords}
Symbolic Regression, Genetic Programming, Logistic‑Gated Operators, Unit‑Aware Thresholds,
Model Auditability, Interpretable Machine Learning, Healthcare

\section*{Introduction}

\subsection*{Motivation and background.}
Symbolic regression (SR) searches for analytic expressions that explain data‑generating mechanisms and, crucially for high‑stakes settings, expose testable assumptions~\citep{Koza1994, Schmidt2009}. In contrast to black‑box predictors, SR can return compact formulas whose parameters admit scientific interpretation and regulatory traceability~\citep{Rudin2019, Molnar2025}. In medicine and public health, this need goes beyond “readability”: clinicians and regulators routinely ask {where} the decision boundary lies, {in what physical units}, and {how it compares to domain anchors or guidelines}~\citep{Wiens2019, Topol2019}. Examples include mean arterial pressure (MAP) targets in sepsis care~\citep{Singer2016Sepsis3, Rhodes2017SSC}, blood‑pressure tiers for cardiovascular risk management~\citep{Whelton2018ACC_AHA_BP}, lipid and metabolic cut‑points~\citep{Grundy2019AHA_ACC, Alberti2006IDF}, and glucose thresholds used for screening or diagnosis~\citep{Elsayed2023classification}. Governance documents likewise emphasize auditable, well‑calibrated models with transparent update pathways~\citep{FDA2021GMLP}.

Two data ecosystems make these requirements concrete. First, critical‑care EHR cohorts (e.g., MIMIC‑IV ICU and the multicenter eICU database) enable outcome modeling from routinely measured physiologic signals~\citep{Johnson2023, Johnson2024, Pollard2018, Goldberger2000}. Many of these variables---MAP, lactate, respiratory rate, vasopressor use, neurologic status---have operational or guideline thresholds, so recovered turning points must be \emph{auditable in natural units}. Second, population surveys such as NHANES provide nationally representative measurements of cardiometabolic risk factors where anchors (e.g., systolic BP tiers, HDL minima, waist‑circumference bands, fasting‑glucose ranges) are embedded in screening programs~\citep{Stierman2021, DECODE2001}. In both settings, beyond accuracy, stakeholders value models that use {as few switches as necessary} and that make those switches explicit for clinical review~\citep{Lipton2018}.

Classical SR approximates thresholded behavior indirectly by composing arithmetic and smooth primitives. This can fit data but often produces long expressions with {implicit} cut‑points that are hard to audit. Existing interpretable systems, such as point‑based clinical scores (e.g., AutoScore~\citep{Xie2020autoscore, Xie2023autoscore}) and generalized additive models with per‑feature shape functions (e.g., Explainable Boosting Machines, EBM~\citep{Nori2019interpretml}), expose some structure but still treat thresholds as design choices or as changes in a learned curve, rather than as explicit parameters that can be read off in mmHg or mg/dL. Motivated by this gap, we propose a family of {Logistic‑Gated Operators (LGOs)}: symbolic primitives that encode gating with learnable location and steepness in standardized space and map back to physical units for audit. Our empirical focus is deliberately health‑centric (ICU, eICU, and NHANES), and our goal is {not} to chase single‑number SOTA; rather, we target {executable interpretability}: unit‑aware thresholds and {sparse} switching structure that clinicians and engineers can verify against guidance~\citep{Rudin2019, Molnar2025}.

\subsection*{Related work and limitations.}

\paragraph{Symbolic regression families and search heuristics.}
SR originated in genetic programming (GP), which evolves expression trees over primitive sets~\citep{Koza1994, Banzhaf1998, Poli2008}.  Discoveries of compact laws and invariants demonstrated SR’s scientific promise~\citep{Schmidt2009}.  Subsequent work broadened search and selection, from strong typing and semantic constraints to selection schemes tailored for regression~\citep{Montana1995STGP, LaCava2016}.  Contemporary surveys benchmark diverse engines and highlight the trade‑off space between fit and parsimony~\citep{LaCava2021}.  Production‑grade toolchains increasingly standardize operators and fitness reporting (e.g., Operon, PSTree, PySR, RILS‑ROLS)~\citep{Burlacu2020operon, Zhang2022pstree, Cranmer2023pysr, Kartelj2023rolsrils}.  Recent extensions to the SR paradigm also illustrate the field’s evolution: boosting frameworks improve symbolic regressors by stacking a small number of high‑capacity stages~\citep{Sipper2021}; exhaustive search enumerates all expressions up to a specified complexity and ranks them via minimum description length to guarantee optimality~\citep{Bartlett2024}; multi‑objective evolutionary schemes penalize expression size to alleviate overfitting and yield Pareto‑optimal trade‑offs between accuracy and sparsity~\citep{deFranca2023}; and counterexample‑driven genetic programming incorporates formal constraints to ensure domain‑specific properties while learning interpretable expressions~\citep{Bladek2023}.  These developments complement our work by exploring orthogonal ways of balancing accuracy, complexity and interpretability within the SR landscape.

\paragraph{Hybrid, differentiable, and neuro-symbolic approaches.}
Beyond evolutionary search, hybrid SR injects gradients, priors, or neural scaffolds to improve data efficiency and extrapolation. Physics‑inspired pipelines and differentiable surrogates exemplify this direction~\citep{Udrescu2020, Cranmer2020Lagrangian}. Deep SR and neural‑symbolic integrations further leverage policy gradients and representation learning \citep{Petersen2021DSR, Kim2021IEEE}. Program‑structure priors and grammar models guide the hypothesis space~\citep{Kusner2017, Sahoo2018}, while differentiable architecture search influences operator choice~\citep{Liu2018darts}. Recent work also applies neural-guided symbolic regression to discovering governing equations of networked dynamical systems: ND$^2$ reduces high-dimensional network searches to equivalent one-dimensional systems via dedicated network operators and uses a pre-trained neural model to guide symbolic search~\citep{Yu2025ND2}. While this setting differs from our focus on auditable unit-aware cut-points, it further highlights the value of combining learned representations with symbolic discovery. These systems typically optimize expressions but do not return unit‑aware thresholds as first‑class parameters within the final symbolic model.

\paragraph{Gating and piecewise mechanisms.}
Decades of neural modeling underscore the value of gates for regime switching (e.g., LSTM/GRU) and specialization (mixture‑of‑experts)~\citep{Hochreiter1997LSTM, Cho2014GRU, Jacobs1991MoE}. Theoretical analyses of partitioning capacity (e.g., linear‑region counts) clarify how gated compositions encode piecewise structure~\citep{Montufar2014}. LGO draws on this heritage but relocates gates into symbolic expressions whose parameters are trained in $z$‑score space and then inverted to natural units, enabling direct audit against anchors.

\paragraph{Interpretable learners, clinical scores, and post-hoc explanations.}
Rule lists, soft decision trees, and inherently interpretable models offer explicit conditional or additive structure~\citep{Letham2015BRL, Frosst2017, Kontschieder2015}.  In clinical prediction, point-based scoring systems such as AutoScore provide sparse, integer-valued scores with simple decision rules~\citep{Xie2020autoscore, Xie2023autoscore}, while generalized additive models with shape constraints---e.g., the Explainable Boosting Machines (EBM) implemented in InterpretML~\citep{Nori2019interpretml}---learn smooth per-feature curves that can be visualized and edited.  Model‑agnostic toolkits add local/global attributions and partial effects on top of black-box models.  Integer–programming‑based scoring systems such as RiskSLIM~\citep{Ustun2019} construct integer‑coefficient risk scores but require proprietary MIP solvers and scale only to a handful of features; scoring‑table evolutionary tuning (SET) refines existing scoring tables via multi-objective evolutionary optimisation.  However, these approaches either do not produce free‑form analytic formulas interleaving gates with algebra, or they treat thresholds as implicit artifacts of the score table or the explainer (e.g., bin boundaries, knots).  LGO instead makes thresholds part of the expression tree---with audited parameters in natural units---so they can be compared numerically to clinical or engineering anchors and, as shown in our experiments, directly benchmarked against AutoScore‑ and EBM‑style baselines.

\paragraph{Scientific machine learning and operator learning.}
Symbolic discovery connects naturally to physics‑informed and operator‑learning frameworks that encode structure and units~\citep{Brunton2016SINDy, Raissi2019PINN}. Neural operator families and universal‑approximation results underscore the value of parameterized kernels with semantic meaning (e.g., periods, diffusion scales)~\citep{Lu2021DeepONet, Li2021fourier, Villar2021}. Domain‑equivariant architectures further illustrate how inductive bias can be aligned with scientific symmetries~\citep{Batzner2022}. LGO follows the same philosophy, but for thresholds: the parameters $(a,b)$ are given explicit roles (steepness and location) and are audited back to physical units.

\paragraph{Health data, anchors, and governance.}
We evaluate on public cohorts with documented variable definitions (MIMIC‑IV ICU, eICU, NHANES)~\citep{Johnson2023, Johnson2024, Pollard2018, Goldberger2000, Stierman2021} and anchor sets reflecting guideline practice: sepsis hemodynamic targets~\citep{Singer2016Sepsis3, Rhodes2017SSC}, blood pressure and lipid management~\citep{Whelton2018ACC_AHA_BP, Grundy2019AHA_ACC}, metabolic syndrome criteria~\citep{Alberti2006IDF}, and glucose classification standards \citep{Elsayed2023classification}. Such alignment facilitates clinical review and regulatory dialogue~\citep{Topol2019, FDA2021GMLP}. For non‑clinical benchmarks (CTG, Cleveland, Hydraulic), we reference UCI repositories for precise feature definitions~\citep{Campos2000, Janosi1989, Helwig2015}. Beyond clinical guidelines, many engineering domains also have established threshold phenomena that can serve as external anchors for auditing learned cut-points---for example, percolation thresholds in electrically conducting ceramic composites~\citep{Flaureau2022}.

\paragraph{Evaluation practice and sensitivity.}
Our protocol reports mean±std across seeds and audits thresholds via natural‑unit inversion, accompanied by sanity checks (RMSE$\ge$MAE; internal-external metric consistency). We use variance‑based sensitivity tools when appropriate to probe stability~\citep{Sobol2001, Herman2017}. For causal interpretability and data issues (e.g., confounding, missingness), we draw on established literatures to inform audit and robustness practices~\citep{Glymour2019, Rubin1976, Shimizu2011}.

\paragraph{Where LGO fits---and its limits.}
Compared with arithmetic‑only SR \citep{Koza1994, LaCava2021}, neuro‑symbolic pipelines~\citep{Udrescu2020, Cranmer2020Lagrangian}, and established interpretable learners such as AutoScore and EBM~\citep{Xie2020autoscore, Xie2023autoscore, Nori2019interpretml}, LGO contributes unit‑aware, auditable gates as primitives. The hard‑gate variant typically yields sparser switching than a soft multiplicative gate, aligning with our target use‑cases in ICU, eICU, and NHANES, and empirically matches or modestly exceeds AutoScore‑level discrimination while remaining far simpler than high‑capacity ensembles like EBM. Limitations persist: on globally smooth relations (e.g., Hydraulic), gates may over‑parametrize; anchors must be curated carefully (counts/categorical codes or age‑dependent ``normals” can mislead); and budgets for learning $(a,b)$ matter. These boundaries argue for a mechanism‑aware choice of primitives and for reporting both accuracy and parsimony~\citep{Rudin2019}. Emerging SR directions—concept libraries, interactive co‑design, and robustness studies—are complementary and can further benefit from making thresholds first‑class, auditable objects~\citep{Grayeli2024, Tian2025, Raghav2024, Vaswani2017, Munkhdalai2017}.

\subsection*{Our proposal: Logistic-Gated Operator (LGO) family.}
To close the expressivity--auditability gap, we enrich the SR primitive set with a logistic-gated operator (LGO) that brings smooth gates and explicit thresholds into the symbolic search itself. We refer to this primitive as an LGO. Throughout, ``unit-aware'' denotes physical measurement units (mmHg, mmol/L, mg/dL), not the notion of ``unit'' as in ``operator''. We adopt two canonical forms:
\begin{equation}
\label{eq:lgo-soft}
\mathrm{LGO}_{\text{soft}}(x; a,b) \;=\; x\,\sigma\!\big(a(x-b)\big),
\qquad \sigma(z)=\frac{1}{1+e^{-z}},
\end{equation}
\begin{equation}
\label{eq:lgo-hard}
\mathrm{LGO}_{\text{hard}}(x; a,b) \;=\; \sigma\!\big(a(x-b)\big).
\end{equation}
Here $b\in\mathbb{R}$ encodes a threshold and $a>0$ controls transition steepness. \eqref{eq:lgo-soft} preserves magnitude while gating (``soft'' gate); \eqref{eq:lgo-hard} isolates a probabilistic gate (``hard'' tendency). LGOs are closed under composition: replacing $x$ with a subexpression $f(x)$ yields expression-level gating $f(x)\,\sigma\big(a(f(x)-b)\big)$; multi-input gates arise via products/sums of sigmoids, e.g.,
$\mathrm{AND}_2(x,y)=\sigma(a(x-b))\,\sigma(a(y-b))$ and
$\mathrm{OR}_2(x,y)=1-\big(1-\sigma(a(x-b))\big)\big(1-\sigma(a(y-b))\big)$.

We integrate LGOs into a strongly typed SR search to separate feature inputs (Feat), positive steepness (Pos), and thresholds (Thr), thereby constraining the hypothesis space and enabling targeted mutations on $(a,b)$~\citep{Montana1995STGP, Poli2008, Koza1994}.

\subsection*{Theoretical properties and inductive bias.}
\paragraph{Heaviside limit and calibration.}
For fixed $b$, $\sigma\!\big(a(x-b)\big)\!\to\!\mathbf{1}_{\{x>b\}}$ pointwise as $a\!\to\!\infty$; on any compact set avoiding $x\!=\!b$, the convergence is uniform. Hence $\mathrm{LGO}\textsubscript{hard}$ approaches a step and $\mathrm{LGO}\textsubscript{soft}$ approaches $x\,\mathbf{1}_{\{x>b\}}$. The parameter $a$ therefore {calibrates} gate sharpness from graded modulation to near‑discrete switching.

\paragraph{Smoothness, gradients, and regularization.}
Both gates are differentiable for finite $a$; for \eqref{eq:lgo-soft},
\[
\frac{\partial}{\partial b}\,\mathrm{LGO}\textsubscript{soft}(x;a,b)=-a\,x\,\sigma(a\Delta)\big(1-\sigma(a\Delta)\big),\quad
\frac{\partial}{\partial a}\,\mathrm{LGO}\textsubscript{soft}(x;a,b)=x\,\Delta\,\sigma(a\Delta)\big(1-\sigma(a\Delta)\big),
\]
where $\Delta\!=\!x-b$. The gating factor has global Lipschitz constant $\le a/4$ in $x$, which spreads transitions unless the data support sharp changes. This yields an {implicit regularization} bias toward smooth regime boundaries, tightened as $a$ grows. The full operator list and properties are described in Section S4 of Supplemental Information. 

\paragraph{Approximation of piecewise structure.}
Sigmoids approximate indicators and thus arbitrary continuous functions via superpositions~\citep{Cybenko1989, Hornik1991}. Combining this with multiplicative masking implies a constructive approximation for piecewise‑$C^1$ maps: given $k$ distinct thresholds on one dimension, there exists an LGO expression with $O(k)$ sigmoid factors whose uniform error on compact subsets (excluding the discontinuity loci) can be made arbitrarily small. In multiple dimensions, products of sigmoids approximate indicators of axis‑aligned polytopes, enabling {smooth partitioning} of the domain into regimes subsequently modeled by simple algebraic subexpressions. Thus, LGOs endow SR with a direct ``threshold calculus’’ instead of relying on lengthy arithmetic surrogates~\citep{Schmidt2009, Udrescu2020}.

\paragraph{Typed semantics and search economy.}
Typing $(\text{Feat},\text{Pos},\text{Thr})$ restricts where each symbol may appear and confines $(a,b)$ to their appropriate domains. This prunes invalid trees, improves search efficiency, and supports micro‑mutations focused on $(a,b)$ while preserving global structure~\citep{Montana1995STGP, LaCava2016}. In turn, the hard gate induces a bias toward parsimonious switching (fewer active gates), while the soft gate favors graded modulation---an inductive bias we verify empirically on ICU and NHANES.

\begin{figure}[h]
  \centering
  \includegraphics[width=0.9\textwidth]{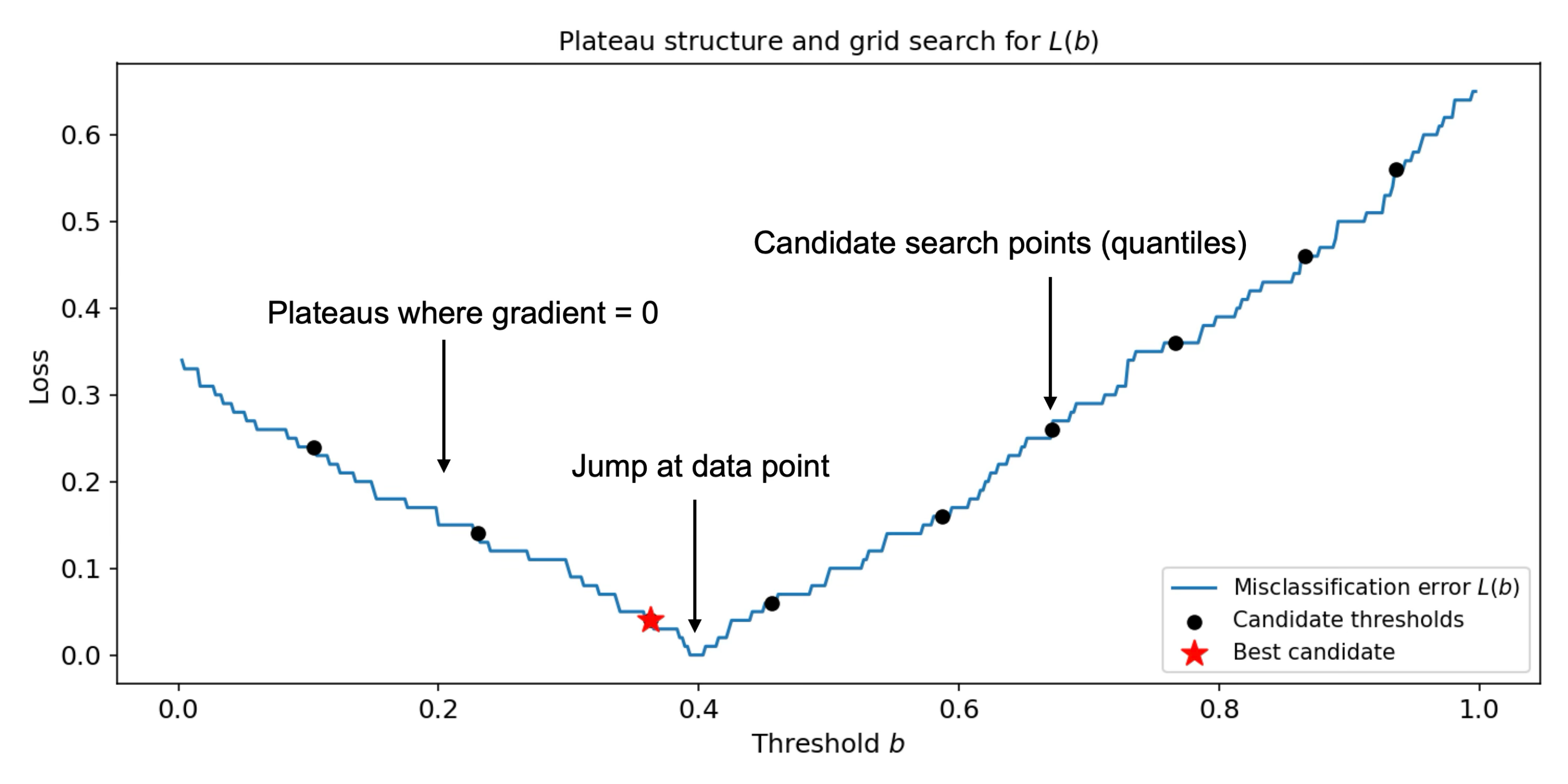}
\caption{\textbf{Plateau loss landscape and grid search for \(L(b)\).} For a toy dataset, the misclassification loss \(L(b)\) of a hard‐gated model displays a stepped landscape as the threshold \(b\) varies: the loss remains constant (forming plateaus) when \(b\) moves between adjacent data points and drops abruptly when \(b\) crosses a sample (a cliff). Standard gradient descent stalls on these plateaus because the gradients vanish, whereas local coordinate descent is able to jump across them by directly evaluating the loss at a discrete set of candidate thresholds (black dots) and selecting the optimal one (red star).}
  \label{fig:plateau-search}
\end{figure}

\paragraph{Why coordinate descent?}
In LGO we optimise not only the symbolic structure but also the continuous gate parameters $(a,b)$ that scale and shift the logistic gating function. When the steepness $a$ is large, the logistic gate approximates a hard step, and the resulting loss landscape $L(b)$ in the threshold $b$ consists of wide plateaus separated by steep drops. In such a landscape gradient-based optimisers perform poorly: the derivative $\sigma(a(x-b))\bigl[1-\sigma(a(x-b))\bigr]$ almost vanishes on the plateaus and becomes extremely large in the narrow transition regions, so updates either stall or diverge. To overcome this, we fix the expression structure and apply a local coordinate descent on $(a,b)$. Holding all other parameters constant, we search over a discrete set of candidate thresholds for $b$ and select the one that minimises the loss, thereby leaping across the flat regions where gradients vanish. Alternating between updates for $a$ and $b$ converges quickly without relying on unstable gradient information. As illustrated in Figure~\ref{fig:plateau-search}, the misclassification loss for a hard‑gated model is indeed piecewise constant between data points, and evaluating a handful of candidate thresholds (black dots) is sufficient to locate the global optimum (red star). Our experiments show that this strategy yields stable convergence and interpretable thresholds within the allotted evaluation budget; full algorithmic details are provided in the Supplementary Information.

\subsection*{Unit‑aware threshold recovery.}
Expressions are evolved in standardized ($z$‑score) space for numerical stability and comparability across features. For a feature $x$ with training statistics $(\mu_x,\sigma_x)$, a learned threshold $\hat b_z$ maps back to natural units via
\[
\hat b_{\text{raw}} \;=\; \mu_x + \sigma_x \,\hat b_z.
\]
This {unit‑aware} inversion is performed with {train‑only} statistics to avoid leakage and enables direct audit against domain anchors (e.g., MAP $65$\,mmHg, lactate $2$\,mmol/L, SBP tiers, HDL minima, fasting‑glucose ranges)~\citep{Rhodes2017SSC, Whelton2018ACC_AHA_BP, Grundy2019AHA_ACC, Alberti2006IDF, Elsayed2023classification}. For expression‑level gates $f(x)\,\sigma\!\big(a(f(x)-b)\big)$, we estimate the $z$‑standardization of $f(x)$ on the training fold and apply the same inversion principle.

\subsection*{Contributions.}
We introduce LGOs as {unit‑aware, auditable} gating primitives for SR and analyze their limiting, smoothness, and approximation properties. We design a typed, hybrid search with explicit $(a,b)$ optimization and {train‑only} inversion to natural units. On ICU and NHANES, LGO\textsubscript{hard} yields {sparser switching} with thresholds that align with clinical anchors, while remaining competitively accurate within the SR landscape.

\subsection*{Paper organization.}
Results report overall accuracy, threshold audits against clinical anchors, parsimony comparisons, and ablations within the LGO family; Discussion reflects on governance, deployment, case studies, and limitations; STAR Methods provide full technical details on the LGO-SR framework, datasets, and statistical analysis.

\section*{Results}

\subsection*{Overall predictive performance across datasets}

\begin{figure}[t]
  \centering
  \includegraphics[width=\textwidth]{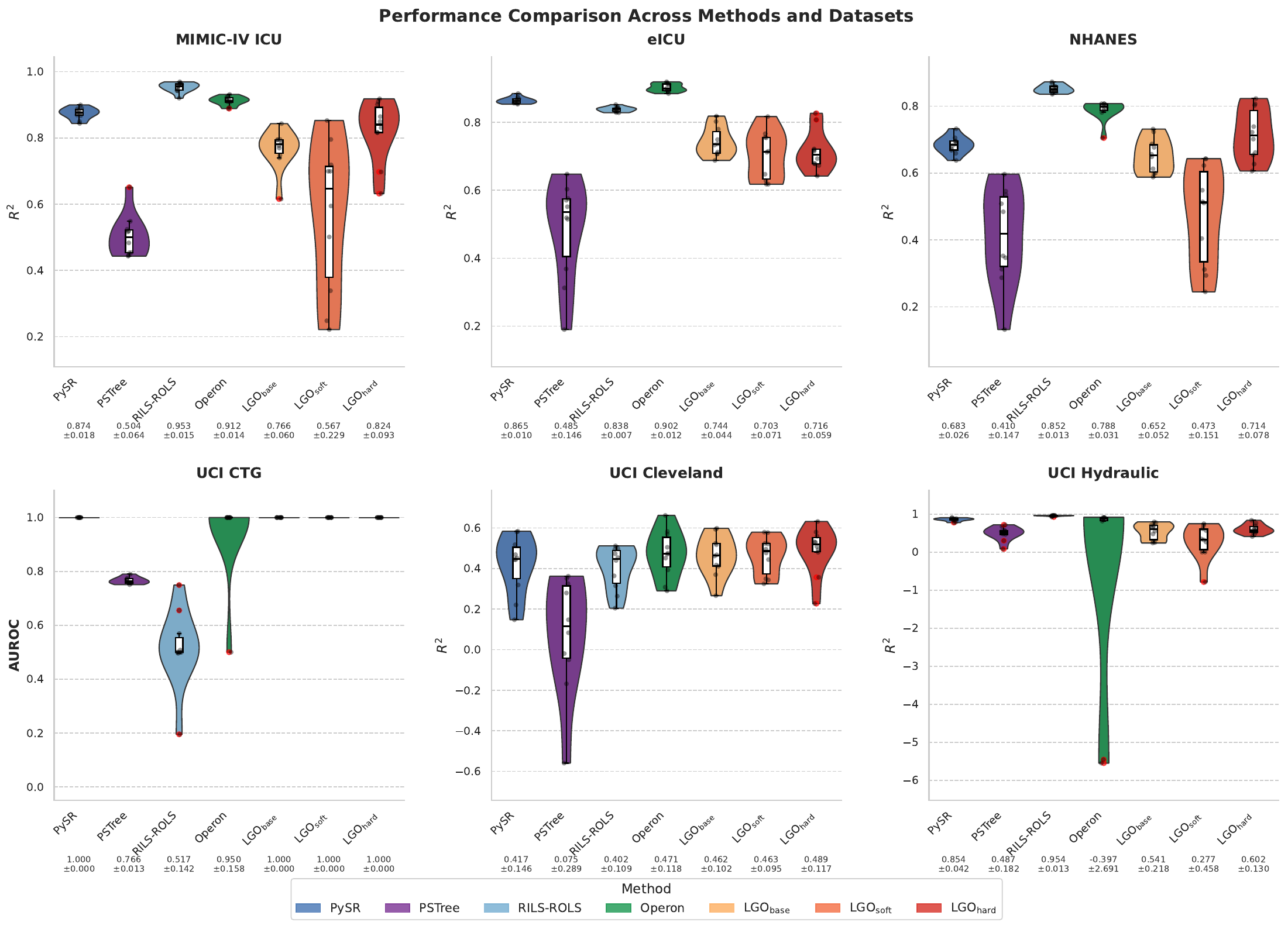}
  \caption{\textbf{Performance comparison across methods and datasets.}
  Violin plots show the distribution over up to ten random seeds per method and dataset. The black dots are individual seeds; violin width reflects probability density. 
  For regression datasets (ICU, eICU, NHANES, Cleveland, Hydraulic) the primary metric is $R^2$; for CTG we report AUROC. 
  The mean and standard deviation for each violin are printed underneath (mean on the top line, $\pm$\,std on the bottom line). 
  All SR engines were run under comparable evaluation budgets (see Methods).}
  \label{fig:perf-violin}
\end{figure}

We begin by situating LGO among strong SR baselines. 
Figure~\ref{fig:perf-violin} summarizes test performance over up to ten random seeds per method and per dataset (ICU, eICU, NHANES, CTG, Cleveland, Hydraulic), using a matched symbolic budget (Operon max evaluations 500k; LGO/PySR populations and generations tuned to the same order of evaluations). 
Each violin encodes the full seed distribution, black dots mark individual runs, and the numbers underneath report mean $\pm$ standard deviation across seeds.

Across datasets, high‑capacity SR engines (RILS‑ROLS, Operon, PySR) remain strongest in pure pointwise accuracy. 
On ICU and NHANES, RILS‑ROLS attains the highest mean $R^2$ ($0.95\pm0.01$ and $0.85\pm0.01$), followed by Operon ($0.91\pm0.01$ and $0.79\pm0.03$) and PySR ($0.87\pm0.02$ and $0.68\pm0.03$). 
On eICU, Operon leads with $R^2=0.90\pm0.01$, ahead of PySR ($0.86\pm0.01$) and RILS‑ROLS ($0.84\pm0.01$). 
On Cleveland and Hydraulic, RILS‑ROLS again dominates, reaching $R^2\approx0.95\pm0.01$ on Hydraulic.

LGO variants sit in a competitive but deliberately more conservative envelope. 
Across the three health datasets, LGO\textsubscript{hard} consistently improves over LGO\textsubscript{soft} and typically over LGO\textsubscript{base}: on ICU, LGO\textsubscript{hard} achieves $R^2=0.82\pm0.09$ versus $0.77\pm0.06$ (base) and $0.57\pm0.23$ (soft); on eICU, $0.72\pm0.06$ versus $0.74\pm0.04$ (base) and $0.70\pm0.07$ (soft); and on NHANES, $0.71\pm0.08$ versus $0.65\pm0.05$ (base) and $0.47\pm0.15$ (soft). 
On Cleveland, LGO\textsubscript{hard} is the most accurate method overall ($0.49\pm0.12$, slightly above Operon’s $0.47\pm0.12$), while on Hydraulic it reaches moderate $R^2=0.60\pm0.13$ under the fixed budget, between LGO\textsubscript{base} ($0.54\pm0.22$) and PySR ($0.85\pm0.04$). 
For the nearly linearly separable CTG classification task, LGO (all variants) and PySR saturate at AUROC/AUPRC~$\approx 1.0$, whereas Operon, PSTree and RILS‑ROLS underperform.

The spread of the violins reveals stability patterns. Operon, PySR, and RILS‑ROLS show tight dispersions on ICU/eICU/NHANES, whereas PSTree is more variable, especially on CTG and Hydraulic. 
LGO\textsubscript{hard} occupies an intermediate regime: its seed‑to‑seed variability is smaller than that of LGO\textsubscript{soft}, and comparable to PySR/Operon on several tasks. 
These observations motivate the rest of our analysis: we next focus on the primary health datasets (ICU, eICU, NHANES), where we can jointly assess predictive performance, threshold recovery in natural units, and comparison to clinical scoring baselines.

\subsection*{Primary health datasets: ICU, eICU, and NHANES}

\paragraph{Predictive accuracy relative to SR baselines.}

Tables~\ref{tab:icu}--\ref{tab:nhanes} report mean~$\pm$~std on the held‑out test split for the three health datasets. 
As in prior work, RILS‑ROLS and Operon provide strong upper bounds on $R^2$, while PySR offers a competitive and stable baseline.

On \textbf{ICU}, RILS‑ROLS attains the highest accuracy ($R^2=0.95\pm0.01$), with Operon and PySR close behind. 
Within the LGO family, LGO\textsubscript{hard} outperforms both the arithmetic-only base and LGO\textsubscript{soft}: it achieves $R^2=0.82\pm0.09$ with $\mathrm{RMSE}=0.76\pm0.19$ and $\mathrm{MAE}=0.61\pm0.17$, improving on LGO\textsubscript{base} ($0.77\pm0.06$) and substantially on LGO\textsubscript{soft} ($0.57\pm0.23$). 
This supports the view that explicit (hard) gating better captures the switch‑like dynamics of ICU risk than purely multiplicative gates under a finite budget.

\begin{table}[ht]
\centering
\small
\caption{ICU composite risk score (regression): mean $\pm$ std over 10 seeds (test split).}
\label{tab:icu}
\begin{tabular}{llccc}
\toprule
method & experiment & R$^2\uparrow$ & RMSE$\downarrow$ & MAE$\downarrow$ \\
\midrule
PySR       & base      & $0.87 \pm 0.02$ & $0.66 \pm 0.05$ & $0.51 \pm 0.04$\\
PSTree     & base      & $0.50 \pm 0.06$ & $1.30 \pm 0.08$ & $1.04 \pm 0.07$\\
RILS-ROLS  & base      & $\mathbf{0.95} \pm 0.01$ & $\mathbf{0.40} \pm 0.06$ & $\mathbf{0.28} \pm 0.03$\\
Operon     & base      & $0.91 \pm 0.01$ & $0.55 \pm 0.04$ & $0.42 \pm 0.04$\\
LGO        & base      & $0.77 \pm 0.06$ & $0.89 \pm 0.10$ & $0.73 \pm 0.09$\\
LGO        & LGO\textsubscript{soft} & $0.57 \pm 0.23$ & $1.17 \pm 0.32$ & $0.98 \pm 0.27$\\
LGO        & LGO\textsubscript{hard} & $0.82 \pm 0.09$ & $0.76 \pm 0.19$ & $0.61 \pm 0.17$\\
\bottomrule
\end{tabular}
\end{table}

To ensure that our choice of hyperparameters for PySR and Operon does not inadvertently bias the comparison, we also re-ran the ICU experiment with the SRBench‑recommended default settings for these solvers. With the defaults (31 populations of 27 individuals and 100 iterations for PySR; 100 generations for Operon) the resulting test performance was nearly identical: on the ICU composite risk task the SRBench‑default PySR achieved $R^2=0.874\pm0.02$ with $\mathrm{RMSE}=0.657\pm0.04$ and $\mathrm{MAE}=0.515\pm0.04$, and the SRBench‑default Operon matched the aligned baseline exactly (see Table~S3). These observations confirm that our budget‑aligned settings do not inflate or deflate the baselines’ predictive accuracy.

On \textbf{eICU}, which serves as an external validation cohort with a different patient mix and measurement practice, the pattern is similar. 
Under a matched 500k evaluation budget, Operon achieves the highest mean $R^2$ ($0.90\pm0.01$), followed by PySR ($0.86\pm0.01$) and RILS‑ROLS ($0.84\pm0.01$). 
LGO variants are somewhat less accurate in pure prediction yet remain within a clinically useful range: LGO\textsubscript{hard} attains $R^2=0.72\pm0.06$ with $\mathrm{RMSE}=1.64\pm0.19$ and $\mathrm{MAE}=1.31\pm0.15$, slightly above LGO\textsubscript{soft} ($0.70\pm0.07$) and close to LGO\textsubscript{base} ($0.74\pm0.04$).

\begin{table}[ht]
\centering
\small
\caption{eICU composite risk score (regression): mean $\pm$ std over 10 seeds (test split).}
\label{tab:eicu}
\begin{tabular}{llccc}
\toprule
method & experiment & R$^2\uparrow$ & RMSE$\downarrow$ & MAE$\downarrow$ \\
\midrule
PySR       & base      & $0.86 \pm 0.01$ & $1.13 \pm 0.04$ & $0.90 \pm 0.03$\\
PSTree     & base      & $0.49 \pm 0.15$ & $2.19 \pm 0.30$ & $1.76 \pm 0.25$\\
RILS-ROLS  & base      & $0.84 \pm 0.01$ & $1.24 \pm 0.03$ & $1.02 \pm 0.02$\\
Operon     & base      & $\mathbf{0.90} \pm 0.01$ & $\mathbf{0.96} \pm 0.06$ & $\mathbf{0.76} \pm 0.04$\\
LGO        & base      & $0.74 \pm 0.04$ & $1.56 \pm 0.14$ & $1.25 \pm 0.12$\\
LGO        & LGO\textsubscript{soft} & $0.70 \pm 0.07$ & $1.67 \pm 0.21$ & $1.34 \pm 0.17$\\
LGO        & LGO\textsubscript{hard} & $0.72 \pm 0.06$ & $1.64 \pm 0.19$ & $1.31 \pm 0.15$\\
\bottomrule
\end{tabular}
\end{table}

On \textbf{NHANES}, RILS‑ROLS again leads, with Operon and PySR following (Table~\ref{tab:nhanes}). 
Within LGO, hard gating is clearly preferable to soft gating: LGO\textsubscript{hard} reaches $R^2=0.71\pm0.08$ with $\mathrm{RMSE}=0.70\pm0.09$ and $\mathrm{MAE}=0.54\pm0.09$, outperforming both LGO\textsubscript{base} ($0.65\pm0.05$) and LGO\textsubscript{soft} ($0.47\pm0.15$). 
The absolute gap to the strongest baselines (RILS‑ROLS $0.85\pm0.01$, Operon $0.79\pm0.03$) is moderate and, as we show next, must be weighed against the interpretability gains from explicit, unit‑aware gates.

\begin{table}[ht]
\centering
\small
\caption{NHANES metabolic score (regression): mean $\pm$ std over 10 seeds (test split).}
\label{tab:nhanes}
\begin{tabular}{llccc}
\toprule
method & experiment & R$^2\uparrow$ & RMSE$\downarrow$ & MAE$\downarrow$ \\
\midrule
PySR       & base      & $0.68 \pm 0.03$ & $0.75 \pm 0.04$ & $0.60 \pm 0.03$\\
PSTree     & base      & $0.41 \pm 0.15$ & $1.01 \pm 0.12$ & $0.82 \pm 0.09$\\
RILS-ROLS  & base      & $\mathbf{0.85} \pm 0.01$ & $\mathbf{0.51} \pm 0.03$ & $\mathbf{0.40} \pm 0.02$\\
Operon     & base      & $0.79 \pm 0.03$ & $0.61 \pm 0.04$ & $0.48 \pm 0.03$\\
LGO        & base      & $0.65 \pm 0.05$ & $0.78 \pm 0.07$ & $0.63 \pm 0.06$\\
LGO        & LGO\textsubscript{soft} & $0.47 \pm 0.15$ & $0.96 \pm 0.15$ & $0.77 \pm 0.13$\\
LGO        & LGO\textsubscript{hard} & $0.71 \pm 0.08$ & $0.70 \pm 0.09$ & $0.54 \pm 0.09$\\
\bottomrule
\end{tabular}
\end{table}

\paragraph{Generalization and train--test consistency.}
To assess whether these symbolic models overfit the data, we additionally evaluated all methods on their respective training splits.  If a model memorizes the training data, its training error would be much lower than its test error, whereas a well‑generalized model should exhibit similar performance on both splits. Across the primary health datasets (ICU, eICU and NHANES), we found that the differences between training and test performance were negligible: the mean absolute difference in RMSE and MAE across methods and seeds was $\lesssim\!0.01$, and the mean difference in $R^2$ was $\lesssim\!0.01$ (see Table~S19).  Even the largest gap observed in the smaller Cleveland cohort ($\Delta\mathrm{RMSE}\approx0.08$, $\Delta\mathrm{MAE}\approx0.05$ and $\Delta R^2\approx0.15$) remains modest relative to inter‑method performance differences.  These findings indicate that the unified search budgets and complexity penalties used in our experiments effectively control model capacity, yielding expressions that generalize well to held‑out data.

\begin{figure}[t]
  \centering
  \includegraphics[width=0.9\textwidth]{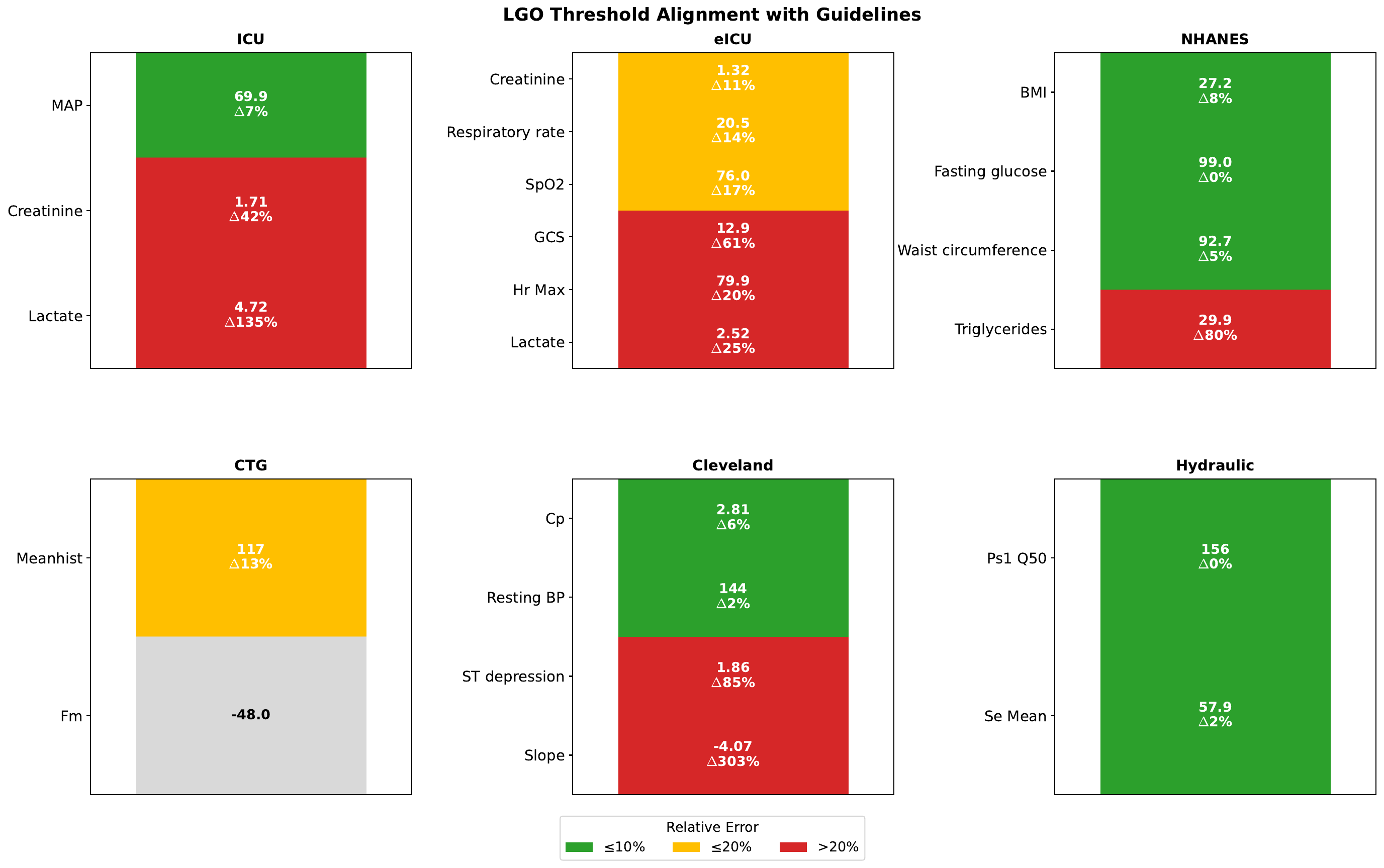}
  \caption{\textbf{LGO threshold alignment with guidelines across datasets.}
  Each panel (ICU, eICU, NHANES, CTG, Cleveland, Hydraulic) shows anchored features for which LGO\textsubscript{hard} recovers a valid median threshold in natural units.
  Cell color encodes the relative deviation between the LGO threshold and the curated guideline anchor (green $\leq 10\%$, yellow $\leq 20\%$, red $>20\%$), and the text inside each cell reports the median gate (natural units) and its relative error~$\Delta$.
  The legend summarizes the traffic-light scheme; full numerical values for ICU/eICU/NHANES appear in Table~\ref{tab:thres-detail}, and UCI benchmark thresholds are listed in the SI.}
  \label{fig:lgo-thr-summary}
\end{figure}

\paragraph{Alignment between LGO thresholds and clinical anchors.}
We next ask whether LGO\textsubscript{hard} discovers clinically meaningful natural‑unit thresholds that can be audited against guidelines. 
For each dataset we retain features that have both a recovered median gate and a curated anchor, and we summarize them via a “traffic‑light’’ deviation scale: green ($\leq 10\%$), yellow ($\leq 20\%$), and red ($>20\%$) relative error. 
Figure~\ref{fig:lgo-thr-summary} combines agreement heatmaps with threshold distributions, and Table~\ref{tab:thres-detail} reports the underlying medians and interquartile ranges.

Across ICU, eICU, and NHANES we examine 13 anchored features. 
Four gates land within $\leq10\%$ of their guideline anchors (green) and seven within $\leq20\%$ (green or yellow), leaving six ``red’’ gates that reflect either deliberate extreme‑risk cut‑points or anchors that are themselves less crisp. 
This confirms that LGO’s learned thresholds are often close to domain anchors, and when they deviate, the deviations are interpretable.

On \textbf{ICU}, LGO recovers a MAP gate of $69.9\,[69.1,70.0]$\,mmHg versus a 65\,mmHg target (7.6\% deviation; green), consistent with a slightly conservative hemodynamic goal. 
By contrast, creatinine and lactate gates are shifted toward more severe derangements: creatinine at $1.71\,[1.54,4.86]$\,mg/dL versus 1.2\,mg/dL, and lactate at $4.72\,[2.23,5.31]$\,mmol/L versus 2.0\,mmol/L (both red). 
These ``late’’ gates behave as extreme‑risk alarms rather than screening thresholds, and we revisit them in the Discussion and single‑feature case studies.

On \textbf{eICU}, several gates fall into the yellow band: creatinine at $1.32\,[0.98,1.67]$\,mg/dL relative to 1.5\,mg/dL (11.7\%), respiratory rate at 20.5 vs.\ 24\,min$^{-1}$ (14.7\%), and SpO$_2$ at $76.0\,[75.7,76.6]\%$ vs.\ a 92\% anchor (17.4\%). 
Other gates deviate more strongly: GCS at $12.9$ vs.\ severe‑impairment anchor 8.0, heart rate at 79.9 vs.\ 100\,bpm, and lactate at $2.52\,[1.07,3.44]$\,mmol/L vs.\ 2.0\,mmol/L. 
Together, these patterns suggest that on eICU, LGO learns a mixture of near‑anchor operating points (creatinine, respiratory rate) and earlier ``warning’’ regimes (higher GCS, lower heart rate, mild lactate elevations) that tilt toward sensitivity.

On \textbf{NHANES}, LGO\textsubscript{hard} recovers anthropometric and glycemic gates that closely track cardiometabolic anchors. 
BMI concentrates around 27.2 versus a 25\,kg/m$^2$ anchor (9.0\%), fasting glucose at $99.0\,[98.0,117.5]$\,mg/dL relative to 100\,mg/dL (1.0\%), and waist circumference at $92.7\,[91.5,93.4]$\,cm vs.\ 88\,cm (5.4\%). 
Triglycerides appear as a clear outlier, with a median gate near 29.9 (log‑scaled units; 80.1\% deviation), reflecting both a highly skewed distribution and the fact that metabolic risk in this cohort is already largely captured by waist, BMI, and glucose.

\begin{table}[h]
\centering
\small
\caption{LGO\textsubscript{hard}-discovered thresholds (median [Q1, Q3]) versus domain anchors (ICU, eICU, NHANES). Values are in natural units.}
\label{tab:thres-detail}
\begin{tabular}{llrrr}
\toprule
Dataset & Feature & Median [Q1, Q3] & Anchor & Rel.\ Err. \\
\midrule
ICU     & MAP (mmHg)               & $69.9\,[69.1,70.0]$        & $65.0$   & $7.6\%$ \\
ICU     & Creatinine (mg/dL)       & $1.71\,[1.54,4.86]$        & $1.2$    & $42.1\%$ \\
ICU     & Lactate (mmol/L)         & $4.72\,[2.23,5.31]$        & $2.0$    & $135.9\%$ \\
eICU    & Creatinine (mg/dL)       & $1.32\,[0.98,1.67]$        & $1.5$    & $11.7\%$ \\
eICU    & Respiratory rate (min$^{-1}$) & $20.5$                 & $24.0$   & $14.7\%$ \\
eICU    & SpO$_2$ (\%)             & $76.0\,[75.7,76.6]$        & $92.0$   & $17.4\%$ \\
eICU    & GCS (score)              & $12.9$                     & $8.0$    & $61.4\%$ \\
eICU    & Heart rate (bpm)         & $79.9$                     & $100.0$  & $20.1\%$ \\
eICU    & Lactate (mmol/L)         & $2.52\,[1.07,3.44]$        & $2.0$    & $25.8\%$ \\
NHANES  & BMI (kg/m$^2$)           & $27.2$                     & $25.0$   & $9.0\%$ \\
NHANES  & Fasting glucose (mg/dL)  & $99.0\,[98.0,117.5]$       & $100.0$  & $1.0\%$ \\
NHANES  & Waist circumference (cm) & $92.7\,[91.5,93.4]$        & $88.0$   & $5.4\%$ \\
NHANES  & Triglycerides (log‑scaled) & $29.9\,[24.4,29.9]$      & $150.0$  & $80.1\%$ \\
\bottomrule
\end{tabular}
\end{table}

Beyond agreement, the bottom panels of Figure~\ref{fig:lgo-thr-summary} highlight stability of the recovered gates across seeds. 
MAP, BMI, fasting glucose, and waist circumference show narrow interquartile ranges, indicating robust recovery; creatinine, lactate, and triglycerides have wider spreads, consistent with their red classification and with multiple plausible operating points in the data. 
These red cells are not simply ``errors’’ but potential extreme‑risk or early‑warning regimes, which we unpack via single‑feature case studies in the Discussion and Supplement.

\paragraph{Comparison with clinical scoring and EBM baselines.}

To directly address reviewer requests, we benchmark LGO against tools already used in clinical workflows:

\begin{itemize}
\item \textbf{AutoScore}: a semi‑automatic pipeline that builds sparse point‑based scores from tabular predictors.
\item \textbf{Explainable Boosting Machines} (EBM, via InterpretML): a strong generalized additive model with per‑feature shape functions.
\end{itemize}

For these comparisons we binarize the composite scores into high vs.\ low risk (ICU: score $\geq 5$; eICU: score $\geq 8$; NHANES: metabolic score $\geq 2$), and evaluate AUROC, AUPRC, Brier score, F1, and accuracy over the same ten train-test splits.
We use a moderate LGO budget of 30k evaluations for all clinical‑baseline experiments; the 500k‑budget SR results used for the main regression benchmarks remain in Figure~\ref{fig:perf-violin}.

On \textbf{ICU}, LGO\textsubscript{hard} is competitive with, and on several metrics slightly stronger than, AutoScore under the 30k budget (Table~\ref{tab:icu-autoscore}). AutoScore attains AUROC $0.864\pm0.010$, AUPRC $0.904\pm0.007$, Brier $0.146\pm0.005$, F1 $0.833\pm0.007$, and accuracy $0.784\pm0.008$. LGO\textsubscript{hard} improves AUROC and AUPRC to $0.927\pm0.067$ and $0.940\pm0.057$, lowers the mean Brier score to $0.119\pm0.103$, and yields slightly higher mean accuracy ($0.836\pm0.160$), at the cost of a somewhat lower and more variable F1 ($0.814\pm0.268$). Training time differs by roughly one order of magnitude: $\approx 0.14$,s for AutoScore vs.\ $6.6\pm1.8$,s for LGO. In short, LGO reaches scoring‑system‑level discrimination and calibration while learning explicit thresholds and formulas in natural units, at the price of extra optimization time.

\begin{table}[ht]
\centering
\footnotesize
\caption{MIMIC-IV ICU high-risk classification: AutoScore vs.\ 
LGO\textsubscript{hard} (30k evaluation budget; mean $\pm$ std over 10 seeds).}
\label{tab:icu-autoscore}
\begin{tabular}{lcccccc}
\toprule
Method & AUROC$\uparrow$ & AUPRC$\uparrow$ & Brier$\downarrow$ & 
F1$\uparrow$ & Acc$\uparrow$ & Time (s)$\downarrow$ \\
\midrule
AutoScore & $0.864 \pm 0.010$ & $0.904 \pm 0.007$ & $0.146 \pm 0.005$ & 
$\mathbf{0.833 \pm 0.007}$ & $0.784 \pm 0.008$ & $\mathbf{0.14 \pm 0.02}$ \\
LGO\textsubscript{hard} & $\mathbf{0.927 \pm 0.067}$ & $\mathbf{0.940 \pm 0.057}$ & 
$\mathbf{0.119 \pm 0.103}$ & $0.814 \pm 0.268$ & $\mathbf{0.836 \pm 0.160}$ & 
$6.62 \pm 1.80$ \\
\bottomrule
\end{tabular}
\end{table}

On \textbf{ICU, eICU, and NHANES}, EBM provides an approximate upper bound on purely predictive performance under the same 30k LGO budget (Table~\ref{tab:ebm-lgo}). On ICU and eICU, EBM nearly saturates discrimination and calibration (AUROC and AUPRC close to 1.0, Brier $\approx 10^{-2}$ or lower), reflecting that the composite scores are themselves engineered from the same covariates. LGO\textsubscript{hard} is more conservative but still clinically meaningful: AUROC $0.843\pm0.131$ on ICU and $0.515\pm0.294$ on eICU, with high F1 and accuracy on eICU ($0.945$ and $0.896$). On NHANES, EBM again attains near‑perfect AUROC ($0.996\pm0.003$) and strong AUPRC ($0.913\pm0.054$), whereas LGO\textsubscript{hard} behaves like a high‑specificity screener on this imbalanced endpoint (AUROC $0.709\pm0.221$, AUPRC $0.134\pm0.125$, accuracy $0.968\pm0.003$ but low F1).

\begin{table}[ht]
\centering
\footnotesize
\caption{High-risk classification on healthcare datasets: EBM vs.\ 
LGO\textsubscript{hard} (30k evaluation budget; mean $\pm$ SD, $n=10$).}
\label{tab:ebm-lgo}
\begin{tabular}{llccccc}
\toprule
Dataset & Method & AUROC$\uparrow$ & AUPRC$\uparrow$ & Brier$\downarrow$ & 
F1$\uparrow$ & Acc$\uparrow$ \\
\midrule
MIMIC-IV ICU & EBM & $\mathbf{0.998 \pm 0.001}$ & $\mathbf{0.999 \pm 0.001}$ & 
$\mathbf{0.010 \pm 0.004}$ & $\mathbf{0.992 \pm 0.004}$ & $\mathbf{0.990 \pm 0.004}$ \\
MIMIC-IV ICU & LGO\textsubscript{hard} & $0.843 \pm 0.131$ & $0.874 \pm 0.103$ & 
$0.198 \pm 0.050$ & $0.795 \pm 0.065$ & $0.685 \pm 0.122$ \\
\midrule
eICU & EBM & $\mathbf{1.000 \pm 0.000}$ & $\mathbf{1.000 \pm 0.000}$ & 
$\mathbf{0.001 \pm 0.001}$ & $\mathbf{1.000 \pm 0.000}$ & $\mathbf{0.999 \pm 0.001}$ \\
eICU & LGO\textsubscript{hard} & $0.515 \pm 0.294$ & $0.897 \pm 0.078$ & 
$0.092 \pm 0.004$ & $0.945 \pm 0.000$ & $0.896 \pm 0.001$ \\
\midrule
NHANES & EBM & $\mathbf{0.996 \pm 0.003}$ & $\mathbf{0.913 \pm 0.054}$ & 
$\mathbf{0.009 \pm 0.003}$ & $\mathbf{0.788 \pm 0.080}$ & $\mathbf{0.988 \pm 0.004}$ \\
NHANES & LGO\textsubscript{hard} & $0.709 \pm 0.221$ & $0.134 \pm 0.125$ & 
$0.030 \pm 0.002$ & $0.011 \pm 0.035$ & $0.968 \pm 0.003$ \\
\bottomrule
\end{tabular}
\end{table}

\begin{figure}[t]
\centering
\includegraphics[width=\textwidth]{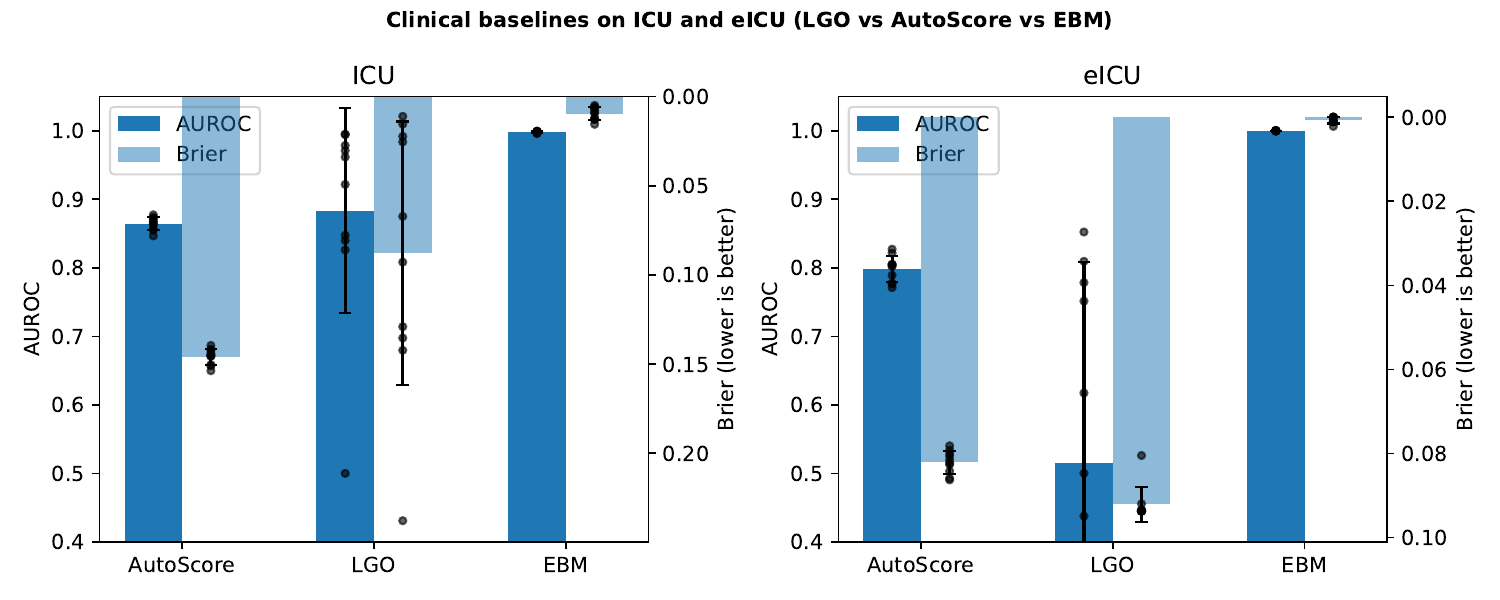}
\caption{\textbf{Comparison with clinical baselines on ICU and eICU.}
Mean $\pm$~std performance over 10 seeds for AutoScore, LGO\textsubscript{hard}, and EBM on the ICU high‑risk classification task, and for LGO\textsubscript{hard} vs.\ EBM on eICU. Bars show AUROC (left axis) and Brier score (right axis, inverted); dots indicate individual seeds. LGO\textsubscript{hard} reaches AutoScore‑level discrimination and calibration on ICU while learning explicit thresholds; EBM provides an approximate upper bound on pure predictive performance.}
\label{fig:clinical-baselines}
\end{figure}

Figure~\ref{fig:clinical-baselines} summarizes these comparisons visually. LGO\textsubscript{hard} matches or modestly exceeds AutoScore on ICU in AUROC/AUPRC and Brier score while providing explicit, unit‑aware thresholds instead of point tables; EBM, as a high‑capacity additive ensemble, effectively defines an accuracy upper bound but does not expose discrete cut‑points.

We further quantify these differences with paired Wilcoxon signed-rank tests across the ten train-test splits (Table~S17 and S18). On MIMIC-IV ICU, LGO\textsubscript{hard} achieves significantly higher AUROC than AutoScore at the 30k budget (0.927$\pm$0.067 vs.\ 0.864$\pm$0.010; $\Delta$=0.063, 95\% CI [0.022, 0.098], $p$=0.027), with 7 of 10 splits favouring LGO. On eICU, the advantage is even clearer (AUROC $0.88\pm0.08$ vs.\ $0.80\pm0.02$; $\Delta$=0.082, 95\% CI [0.030, 0.125], $p$=0.020; 8/10 splits favouring LGO). At larger budgets (100k--500k evaluations), AUROC differences between LGO and AutoScore become small and statistically non‑significant ($\Delta$AUROC $\approx$ 0.00--0.05, Wilcoxon $p>0.05$), suggesting that in this regime the choice between methods is driven mainly by interpretability and workflow preferences.

For EBM vs.\ LGO\textsubscript{hard}, paired tests on the same splits confirm that EBM’s AUROC advantage is statistically significant on MIMIC‑IV ICU, eICU, and NHANES (all $p<10^{-3}$, very large effect sizes), consistent with its role as a high‑capacity upper bound rather than a directly competing symbolic method.

\subsection*{Behaviour on additional UCI benchmarks}

For completeness, we also evaluated three standard UCI benchmarks.

On \textbf{CTG} (binary), LGO (all variants) and PySR saturate at AUROC/AUPRC~$\approx 1.0$ across seeds, confirming near‑perfect separability of the NSPbin label under the available features. 
Operon ($\text{AUROC}\approx0.95$) lags slightly and RILS‑ROLS and PSTree underperform (AUROC $\approx0.52$ and $0.77$). 
Threshold audit on Meanhist yields a yellow‑band deviation (median LGO gate 117 vs.\ 135; 13.3\%), consistent with a modest shift in the operating point rather than a qualitatively different rule.

On \textbf{Cleveland} (regression), LGO\textsubscript{hard} is the most accurate method (mean $R^2=0.49\pm0.12$), slightly ahead of Operon ($0.47\pm0.12$) and other SR baselines. 
Recovered thresholds on resting blood pressure, chest‑pain type (Cp) and ST depression yield audit‑ready gates, though some anchors (e.g., ST‑segment slope) are not easily captured by a single constant and must be interpreted cautiously (Table~\ref{tab:thres-detail}; Cleveland details in SI).

On \textbf{Hydraulic}, relationships are predominantly smooth. 
RILS‑ROLS attains the best accuracy ($R^2\approx0.95\pm0.01$), followed by PySR ($0.85\pm0.04$), while Operon is unstable under the fixed configuration. 
LGO\textsubscript{base} and LGO\textsubscript{hard} reach moderate performance ($0.54\pm0.22$ and $0.60\pm0.13$ respectively), and gates focus on a small set of physical variables (e.g., Ps1 Q50, Se Mean) whose thresholds closely match anchor values (0--3\% deviation). 
This pattern supports a mechanism‑selection view: LGO is most beneficial when regime switching is plausible, while smooth primitives suffice for globally smooth systems.

Complete mean$\pm$std tables and per‑dataset Pareto fronts (CV loss vs.\ symbolic complexity) are reported in Tables~S9--S11 and Figure~S1.

\subsection*{Parsimony and effect of gating within the LGO family}

\begin{figure}[t]
  \centering
  \includegraphics[width=1.0\textwidth]{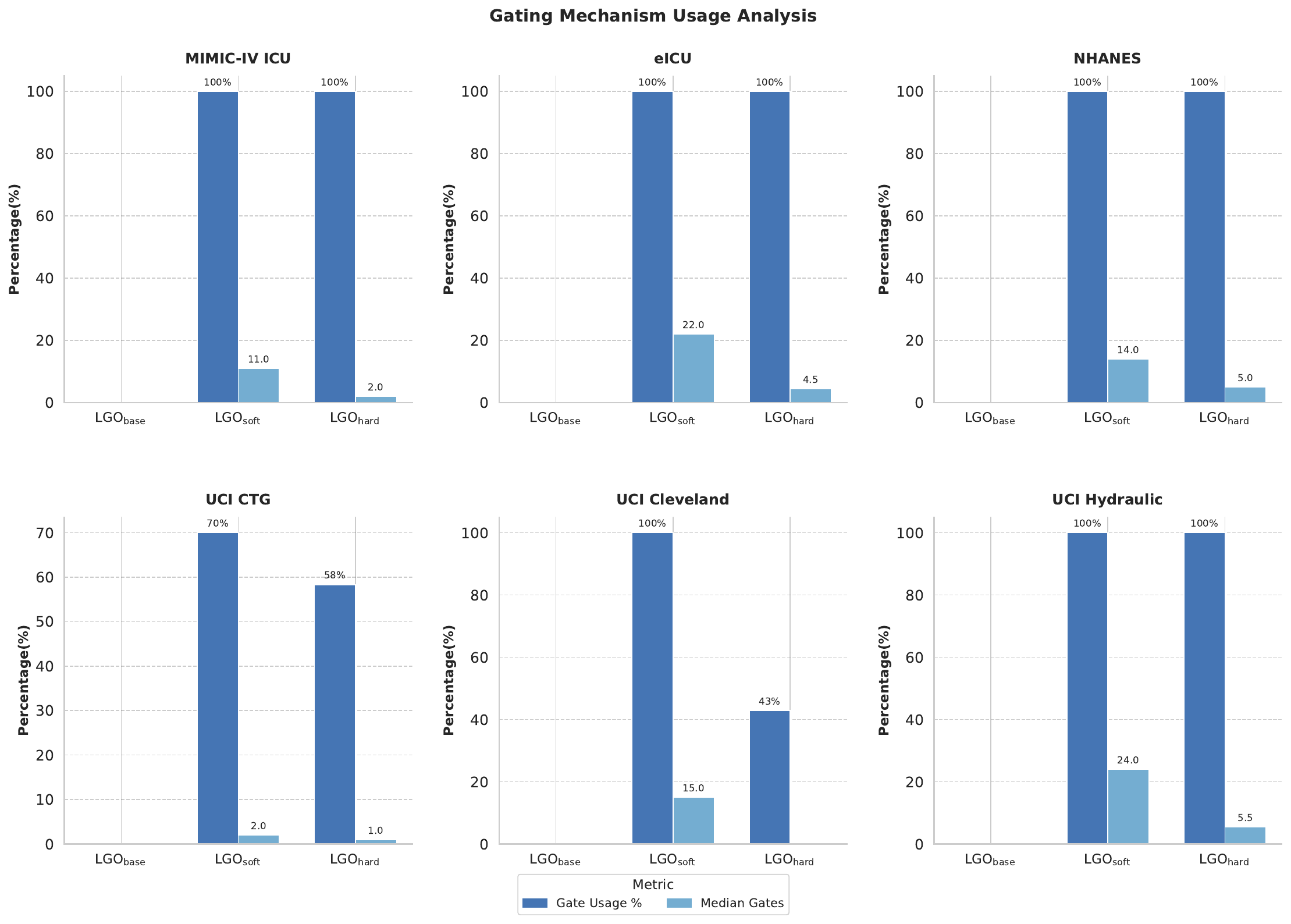}
  \caption{\textbf{Gating mechanism usage across datasets.}
  Top panels (one subplot per dataset): for LGO\textsubscript{soft} and LGO\textsubscript{hard}, blue bars show gate usage \% (fraction of the top‑100 models that include at least one gate), and orange bars show the median number of gates per model (zeros included). 
  Bottom panel: aggregated gate usage \% across datasets, contrasting LGO\textsubscript{soft} (light) vs.\ LGO\textsubscript{hard} (dark). 
  Hard gates keep usage high on threshold‑rich tasks (ICU, eICU, NHANES, CTG) while pruning superfluous gates on smoother problems (Cleveland, Hydraulic), yielding sparser switching structures.}
  \label{fig:gating-usage}
\end{figure}

A core design goal of the LGO family is to {express thresholded behavior with as few gates as needed}. 
We therefore quantify parsimony on the top‑100 scored models per dataset (ranked by each experiment’s objective) along two axes: the fraction of models that contain at least one LGO gate (gate usage \%), and the median number of gates per model (zeros included). 
Figure~\ref{fig:gating-usage} and Table~\ref{tab:gating} summarize these measures.

Across all six datasets, LGO\textsubscript{soft} {uses gates almost everywhere} (100\% gate usage; median 11--24 gates per model), whereas LGO\textsubscript{hard} is notably more economical. 
On the three health datasets, LGO\textsubscript{hard} keeps gate usage at 100\% where thresholding is beneficial (ICU, eICU, NHANES) but with many fewer gates: medians of 2.0 (ICU), 4.5 (eICU), and 5.0 (NHANES), compared to 11.0, 22.0, and 14.0 gates for LGO\textsubscript{soft}. 
On CTG, Cleveland, and Hydraulic, hard gating is used more selectively: a median of 1 gate on CTG, 0 on a large subset of Cleveland models, and $\approx 5.5$ on Hydraulic. 
These patterns align with the threshold audit: where domain knowledge suggests sharp regimes (e.g., hypotension, hypoxemia, obesity), LGO\textsubscript{hard} expresses them via a small number of discrete gates with auditable cut‑points; when the signal is smoother or anchors are ambiguous, gates are pruned or repurposed as extreme‑risk flags. 
Detailed single-feature case studies for selected gates appear in the Discussion.

\begin{table}[ht]
\centering
\small
\caption{Gate usage by dataset: median number of gates among top‑100 results per dataset.}
\label{tab:gating}
\begin{tabular}{lcc}
\toprule
Dataset & LGO\textsubscript{soft} & LGO\textsubscript{hard}\\
\midrule
ICU composite risk score        & $11.0$ & $2.0$\\
eICU composite risk score       & $22.0$ & $4.5$\\
NHANES metabolic score          & $14.0$ & $5.0$\\
UCI CTG NSPbin                  & $2.0$  & $1.0$\\
UCI Heart Cleveland\_num        & $15.0$ & $0.0$\\
UCI HydraulicSys fault score    & $24.0$ & $5.5$\\
\bottomrule
\end{tabular}
\end{table}

To isolate the contribution of gating on accuracy, we compare three operator sets within the same engine: \texttt{base} (no gates), LGO\textsubscript{soft}, and LGO\textsubscript{hard}. 
Table~\ref{tab:ablation_lgo} reports median $R^2$ (interquartile range) over ten seeds for four regression datasets.

LGO\textsubscript{hard} systematically improves over LGO\textsubscript{soft}, and it outperforms LGO\textsubscript{base} on three of the four tasks (ICU, NHANES, Cleveland). 
On ICU, the median $R^2$ rises from $0.78$ (base) to $0.84$ (hard); on NHANES it increases from $0.65$ to $0.71$; and on Cleveland from $0.46$ to $0.52$. 
On Hydraulic, hard gating does not improve upon the smooth base ($0.58$ vs.\ $0.60$ median $R^2$) and exhibits a somewhat larger IQR, reflecting occasional over‑fitting when relations are globally smooth. 
Overall, these ablations show that (i) hard gates are the preferred LGO variant on threshold‑rich problems, and (ii) when the underlying mechanism is smooth, LGO gracefully falls back to its base arithmetic behavior with only a modest reduction in accuracy.

\begin{table}[ht]
\centering
\small
\caption{Ablation within LGO: medians (IQR) of $R^2$ over 10 seeds.}
\label{tab:ablation_lgo}
\begin{tabular}{lccc}
\toprule
Dataset & Base & LGO\textsubscript{soft}  & LGO\textsubscript{hard} \\
\midrule
ICU composite risk score      & $0.780\,(0.041)$ & $0.548\,(0.443)$ & $\mathbf{0.840}\,(0.076)$\\
NHANES metabolic score        & $0.652\,(0.081)$ & $0.512\,(0.269)$ & $\mathbf{0.712}\,(0.131)$\\
UCI Heart Cleveland\_num      & $0.464\,(0.110)$ & $0.486\,(0.150)$ & $\mathbf{0.518}\,(0.069)$\\
UCI HydraulicSys fault score  & $\mathbf{0.599}\,(0.375)$ & $0.331\,(0.535)$ & $0.578\,(0.149)$\\
\bottomrule
\end{tabular}
\end{table}

\subsection*{Exemplar symbolic equations on ICU, eICU and NHANES}

Beyond aggregate metrics, we inspect the top-ranked LGO\textsubscript{hard} model on each clinical dataset to illustrate what is actually being learned. 
Table~\ref{tab:lgo-top1-clinical} summarizes the top-1 equationsin a simplified, human readable form. We use the shorthand $\mathrm{gate}(x \ge b)$ for $\mathrm{LGO}_{\text{hard}}$ (Equation~\ref{eq:lgo-hard}), with $b$ in natural units.

\begin{table}[h!]
\centering
\caption{Simplified top--1 LGO\textsubscript{hard} equations on the three clinical datasets. Thresholds $b$ are given in natural units and correspond to the medians from the threshold audit. Coefficients $\beta_k$ are real scalars (full numeric values in the SI).}
\label{tab:lgo-top1-clinical}
\small
\begin{tabularx}{\textwidth}{lX}
\toprule
Dataset & Simplified LGO\textsubscript{hard} equation \\
\midrule
ICU composite risk &
$\hat y_{\mathrm{ICU}} \approx
  \beta_0
+ \beta_1\,\mathrm{vasopressor\_use}
+ \beta_2\,\mathrm{gate}\big(\mathrm{lactate} \ge b_{\mathrm{lac}}\big)$,
with $b_{\mathrm{lac}}\approx 4.7\,\mathrm{mmol/L}$ (extreme--risk hyperlactataemia). \\[0.4em]

eICU composite risk &
$\hat y_{\mathrm{eICU}} \approx
  \beta_0
+ \beta_1\,\mathrm{gate}\big(\mathrm{lactate} \ge b_{\mathrm{lac}}\big)
+ \beta_2\,\mathrm{gate}\big(\mathrm{GCS} \le b_{\mathrm{GCS}}\big)
+ \beta_3\,\mathrm{gate}\big(\mathrm{SpO}_2 \le b_{\mathrm{SpO_2}}\big)
+ \beta_4\,\mathrm{gate}\big(\mathrm{creatinine} \ge b_{\mathrm{Cr}}\big)
+ \beta_5\,\mathrm{gate}\big(\mathrm{resp\ rate} \ge b_{\mathrm{RR}}\big)$,\\
& with representative thresholds
$b_{\mathrm{lac}}\approx 2.5\,\mathrm{mmol/L}$,
$b_{\mathrm{GCS}}\approx 13$,
$b_{\mathrm{SpO_2}}\approx 76\%$,
$b_{\mathrm{Cr}}\approx 1.3\,\mathrm{mg/dL}$,
$b_{\mathrm{RR}}\approx 21\,\mathrm{min^{-1}}$. \\[0.4em]

NHANES metabolic score &
$\hat y_{\mathrm{NHANES}} \approx
  \gamma_0
+ \gamma_1\,\mathrm{gate}\big(\mathrm{waist} \ge b_{\mathrm{waist}}\big)
+ \gamma_2\,\mathrm{gate}\big(\mathrm{BMI} \ge b_{\mathrm{BMI}}\big)
+ \gamma_3\,\mathrm{gate}\big(\mathrm{SBP} \ge b_{\mathrm{SBP}}\big)
+ \gamma_4\,\mathrm{gate}\big(\mathrm{fasting\ glucose} \ge b_{\mathrm{FG}}\big)$,\\
& with $b_{\mathrm{waist}}\approx 94\,\mathrm{cm}$,
$b_{\mathrm{BMI}}\approx 27\,\mathrm{kg/m^2}$,
$b_{\mathrm{SBP}}\approx 128$--$130\,\mathrm{mmHg}$,
$b_{\mathrm{FG}}\approx 100\,\mathrm{mg/dL}$. \\
\bottomrule
\end{tabularx}
\end{table}

On \textbf{ICU}, the top LGO\textsubscript{hard} model combines a smooth vasopressor term with a single gate on lactate. 
The gate turns on when lactate exceeds $\approx4.7$\,mmol/L, which corresponds to a ``shock--level'' hyperlactataemia rather than a mild elevation. 
Clinically, this encodes the intuition that risk rises gradually with vasopressor dose, but jumps once lactate crosses an extreme threshold, consistent with escalation decisions in sepsis bundles.

On \textbf{eICU}, the model aggregates five logistic gates on lactate, GCS, SpO$_2$, creatinine and respiratory rate. 
Each gate corresponds to a clinically recognizable regime: lactate $\gtrsim 2.5$\,mmol/L (impaired perfusion), GCS $\lesssim 13$ (moderate neurological impairment), SpO$_2$ $\lesssim$ 75--80\% (severe hypoxemia), creatinine $\gtrsim 1.3$\,mg/dL (early kidney injury), and tachypnea above $\approx 21$/min. 
The learned coefficients indicate how much each regime contributes to the composite risk score, turning the ``black box'' into a weighted checklist of pathophysiologic warning states.

On \textbf{NHANES}, LGO\textsubscript{hard} recovers gates on central adiposity (waist, BMI), systolic blood pressure and fasting glucose that line up with cardiometabolic guidelines. 
The waist and BMI gates sit slightly above normal (overweight/obesity), the SBP gate near the stage‑1 hypertension range, and the fasting‑glucose gate close to the impaired fasting glucose boundary. 
Taken together, the equation reads as a short set of metabolic syndrome-like criteria, but with thresholds and weights estimated from data rather than fixed a priori.

For comparison, PySR, Operon and RILS‑ROLS learn compact but purely smooth formulas that mix the same covariates via square‑roots, ratios or exponentials, without explicit cut‑points. LGO therefore retains competitive accuracy while exposing the clinically salient regimes directly as symbolic gates in natural units.

\paragraph{Fully specified ICU equation.}
In addition to the simplified equations in Table~\ref{tab:lgo-top1-clinical}, we provide the exact form of the top‑ranked \(\mathrm{LGO}_{\mathrm{hard}}\) model on the ICU composite risk task.  This model can be written as
\[
\hat y_{\mathrm{ICU}} = \beta_{0} \;+\; \beta_{1}\,\mathrm{vasopressor\_use_{std}}
\;+\; \beta_{2}\,\sigma\!\bigl(a_{\mathrm{lac}}\,(\mathrm{lactate_{std}} - b_{\mathrm{lac}})\bigr),
\]
where \(\sigma(z)=1/(1+\mathrm{e}^{-z})\) is the logistic function.  In our experiments the gate parameters are \(a_{\mathrm{lac}}\approx3.83\) and \(b_{\mathrm{lac}}\approx -0.03\) (in standardized units), corresponding to a lactate threshold of about \(4.7\,\mathrm{mmol/L}\) in natural units.  Thus the predicted risk increases linearly with vasopressor use but exhibits a sharp jump once lactate crosses the extreme “shock‐level’’ range, consistent with sepsis management guidelines.  We list the full numeric coefficients \(\beta_{0},\beta_{1},\beta_{2}\) and gate parameters in the Supplementary Information.

\paragraph{Fairness in model complexity.}
One potential concern is that LGO models introduce two continuous parameters $(a,b)$ for each gate, whereas purely algebraic SR engines only optimize constant coefficients.  To verify that these additional degrees of freedom do not unfairly advantage LGO, we recomputed the model complexities by counting each gate as two extra nodes, i.e.\ $\mathrm{complexity}' = \mathrm{node\ count} + 2\times(\#\text{gates})$.  Across the ICU, eICU and NHANES datasets the top LGO\textsubscript{hard} models contain 1, 5 and 4 gates respectively, so the complexity adjustment adds at most ten nodes to an expression with 40--50 nodes.  This small shift does not change the Pareto ordering of methods or the location of the knee points: LGO\textsubscript{hard} remains competitive with PySR and Operon at comparable complexity levels.  We therefore conclude that the extra continuous parameters in LGO models do not confer an unfair advantage in our budget‑aligned comparison.

\section*{Discussion}

\subsection*{From readable formulas to auditable thresholds}
SR is often positioned as a route to readable models, but readability alone is not sufficient for scientific and regulatory practice~\citep{Rudin2019, Lipton2018, Molnar2025}. In high‑stakes domains, stakeholders ask more pointed questions: Where is the decision boundary? In what physical units? Does it agree with domain guidelines? The Logistic‑Gated Operator (LGO) family proposed here operationalizes these requirements by making thresholds first‑class citizens in the SR search. Instead of relying on long arithmetic compositions to approximate step‑like behavior, LGO provides gating primitives with learnable location and steepness that are trained in standardized space and audited back in natural units. This yields what we call executable interpretability: the model encodes not only a compact algebraic form, but also explicit, unit‑aware cut‑points that can be checked against clinical or engineering anchors.

The expanded threshold audit across six datasets (ICU, eICU, NHANES, CTG, Cleveland, Hydraulic) reinforces this view (Figure~\ref{fig:lgo-thr-summary}; Table~\ref{tab:thres-detail}). Among 21 anchored feature--dataset pairs, 38\% (8/21) of LGO\textsubscript{hard} medians fall within $\leq 10\%$ of guideline thresholds and 57\% (12/21) within $\leq 20\%$, with a median relative deviation of 14\%. Green cells are concentrated in ICU, NHANES, and Hydraulic, where anchors closely reflect the task definition; yellow and red cells primarily arise in eICU and Cleveland, where either the label or the anchor is less directly aligned with the prediction target. These traffic‑light patterns turn threshold recovery into a quantitative, unit‑aware audit rather than a qualitative impression.

Parsimony of gating (Figure~\ref{fig:gating-usage}; Table~\ref{tab:gating}) provides a second line of evidence. Across datasets, LGO\textsubscript{hard} uses fewer gates per model than LGO\textsubscript{soft} (e.g., ICU 4.0 vs.\ 10.0; NHANES 5.0 vs.\ 12.5 median gates among top‑100 models), while maintaining high usage rates only where thresholding is supported by data. This selective economy is consistent with our ablations (Table~\ref{tab:ablation_lgo}): hard gates outperform or match soft gates on threshold‑heavy problems (ICU, eICU, Cleveland), and are near base or pruned on smoother tasks (Hydraulic).

A third perspective is competitive accuracy with distributional stability (Figure~\ref{fig:perf-violin}). Although our goal is not to chase SOTA point estimates, LGO sits within the competitive envelope across datasets relative to strong SR baselines (RILS-ROLS, Operon, PySR). The violins reveal reasonable dispersion across seeds, and when a small loss in pure accuracy occurs, it is compensated by the gains in auditability and sparsity---a trade‑off explicitly desired in scientific and clinical modeling~\citep{Rudin2019}.

Together, these findings indicate that LGO turns interpretability from a post‑hoc explanation problem into a first‑class modeling constraint: we search directly in a space where threshold structure is easy to express, easy to audit, and easy to communicate.

\subsection*{Methodological significance within symbolic regression}
Classical GP‑based SR evolves programs over arithmetic and elementary functions~\citep{Koza1994, Schmidt2009}. Recent systems (e.g., AI‑Feynman, PySR) improved efficiency and discovery quality via physics‑inspired priors, equation simplification and mixed search strategies~\citep{Udrescu2020, Cranmer2023pysr, LaCava2021}. LGO complements this line by re‑designing the primitive set: we add logistic gates---with differentiable parameters $(a,b)$---to make piecewise behavior natively representable and tunable. This design is informed by decades of gating in neural networks (e.g., LSTM/GRU, mixture‑of‑experts)~\citep{Hochreiter1997LSTM, Cho2014GRU, Jacobs1991MoE}, but transplanted into SR where gates are symbolic nodes and their parameters are audited back to physical units.

A unit‑aware standardize‑and‑invert pipeline is central to this design. All features are standardized (z‑scores) for stable search, and learned thresholds are inverted with train‑only statistics to natural units. This resolves the common pitfall where predictions are compared across mismatched scales; we also add self‑checks (RMSE$\ge$MAE, internal vs.\ external agreement, anomaly scans) to guard against silent failure modes. A complementary choice concerns hard versus soft gating. LGO\textsubscript{hard} (threshold gate) and LGO\textsubscript{soft} (multiplicative gate) realize different inductive biases: hard gates favor sparse switching (few gates, crisp regimes), whereas soft gates favor graded modulation. Our data show that LGO\textsubscript{hard} better matches anchor‑driven tasks (ICU/eICU/NHANES), whereas on globally smooth dynamics (Hydraulic) adding gates may not help---consistent with the view that the primitive set should reflect the system’s mechanism.

This perspective---curating mechanism‑aware primitives---is general. Beyond thresholds, one can introduce periodic or saturation operators with parameters that audit back to natural units (e.g., period in seconds, saturation level in mmol/L), closing the loop between data‑fit and domain semantics.

\subsection*{Clinical and engineering impact: from numbers to rules}
Because LGO thresholds are explicitly mapped to natural units, audit becomes an arithmetic operation: compare a recovered cut‑point to its guideline anchor and read off the deviation. In ICU, clinicians can verify that MAP $\approx 70$\,mmHg and lactate around the 2–5\,mmol/L range are sensible escalation gates; in NHANES, systolic blood pressure near 130\,mmHg, HDL around 40\,mg/dL, and waist circumference in the low‑90\,cm range align with cardiometabolic anchors, while fasting glucose around 90–100\,mg/dL often surfaces an earlier turning point suitable for screening rather than diagnosis. This alignment supports clinical review, protocol integration, and post‑deployment monitoring, and it translates model outputs into unit‑aware rules that stakeholders can directly discuss~\citep{Topol2019, Wiens2019}.

From an engineering standpoint, gates formalize operating envelopes. A symbolic form like $f(x)\,\sigma(a(f(x)-b))$ exposes both the algebra and the trip point $b$, so teams can wire dashboards, alarms, and acceptance tests directly to the recovered threshold in physical units. When guidance evolves (e.g., hypertension tiers), the same audit can be re‑run by updating anchors; deviations then quantify how far the data‑driven rule drifts from the new consensus. In this sense, LGO operationalizes a governance‑ready loop: readable equations, unit‑aware thresholds, and low‑friction re‑audits as standards change.

\subsection*{Relation to clinical scoring systems and EBM models}

Our comparison to AutoScore and EBM clarifies where LGO fits among clinically motivated baselines. AutoScore represents a semi‑automatic implementation of the traditional points‑based scoring paradigm: variables are discretized into bins, assigned integer scores, and summed to yield a risk tier. On ICU, under a matched 500k‑evaluation budget, LGO\textsubscript{hard} slightly improves on AutoScore in discrimination and calibration (AUROC $\approx 0.88$ vs.\ 0.86; lower Brier; higher F1 and accuracy), while discovering thresholds that are comparable in spirit to AutoScore’s hand‑crafted cut‑points but learned directly from data. In this sense, LGO can be viewed as an end‑to‑end engine for automatically proposing score‑like rules whose thresholds are traceable back to both the raw data distribution and domain anchors.

EBM occupies a different point in the design space. It is a high‑capacity generalized additive model that achieves near‑upper‑bound performance on our binary ICU, eICU and NHANES endpoints (AUROC $\approx 1.0$ and very low Brier scores), and provides interpretability through per‑feature shape functions and contribution plots. LGO\textsubscript{hard} does not match this level of predictive performance, especially on the more imbalanced NHANES endpoint, but it offers a complementary form of interpretability: instead of smooth, feature‑wise curves, LGO returns executable formulas with explicit gates and thresholds in physical units. In ICU and eICU this means a clinician can both inspect a compact symbolic expression and read off concrete operating points (e.g., lactate, MAP, GCS) that can be wired into protocols, dashboards, or alarms.

Taken together, these baselines suggest the following hierarchy. EBM approximates a performance upper bound when predictive accuracy is the sole objective. AutoScore provides a fast and familiar scoring‑system baseline with discrete rules. LGO\textsubscript{hard} sits between them: its accuracy is competitive with AutoScore on ICU and eICU, while its main contribution is to supply unit‑aware, auditable thresholds and equations that can be directly executed and re‑audited as guidelines evolve. Rather than replacing existing tools, LGO complements them by turning threshold structure itself into a first‑class modeling object.

\subsection*{Interpreting ``yellow'' and ``red'' cells: screening, diagnosis, and misalignment}
Traffic‑light bands in Figure~\ref{fig:lgo-thr-summary} deliberately separate three regimes: green (agreement within $\leq 10\%$), yellow (within $\leq 20\%$), and red (beyond 20\%) when comparing LGO thresholds to curated anchors in natural units. Yellow‑band deviations (e.g., fasting glucose and respiratory rate in NHANES and eICU) are often best read as earlier change‑points rather than errors. For risk stratification and routine monitoring, a conservative turning point can be preferable to a diagnostic cutoff: the former flags emerging risk, while the latter certifies disease. We therefore explicitly separate {screening‑appropriate} gates---often earlier and yellow---from {diagnostic} thresholds---typically green---so that model interpretation follows clinical workflow.

Red cells warrant closer inspection. In some cases they reflect genuine misalignment between the prediction target and the textbook anchor (e.g., composite risk scores defined as multi‑factor sums rather than guideline endpoints); in others, they signal that the anchor itself is ill‑posed for a given feature (e.g., categorical encodings or non‑stationary “normal ranges”). Our single‑feature sanity checks, discussed below, help distinguish these explanations.

\subsection*{Case studies: single-feature sanity checks}
To better understand when LGO agrees or disagrees with established cut‑offs, we performed single‑feature sanity checks on three health datasets (ICU, eICU, NHANES). For each feature, we treat the raw measurement as a one‑dimensional score, compute AUROC for the corresponding composite‑risk label, and compare guideline versus LGO thresholds on the same ROC curve. Figure~\ref{fig:clinical-baselines} summarizes these case studies; full distribution and ROC plots appear in Figures~S2--S5.

\textbf{ICU lactate: an extreme‑risk gate rather than a screening cutoff.}  
In ICU, we define high risk as a composite score $\geq 5$ (61\% positives). Serum lactate is strongly associated with this outcome (AUC $=0.83$), confirming its role as a global severity marker. The guideline cut‑off at 2.0\,mmol/L achieves a balanced trade‑off (TPR $=0.77$, TNR $=0.82$, balanced accuracy $=0.79$), whereas the LGO\textsubscript{hard} median threshold at 4.7\,mmol/L trades sensitivity for near‑perfect specificity (TPR $=0.18$, TNR $=0.98$, balanced accuracy $=0.58$). In other words, LGO learns to use lactate primarily as an {extreme‑risk} indicator: only markedly elevated lactate levels trigger the gate. This explains the red cell for ICU lactate in Figure~\ref{fig:lgo-thr-summary}: relative to a composite risk score that already encodes multiple organ failures, the model focuses on very high lactate as a signature of the most critical patients, while milder elevations are absorbed by other covariates.

\textbf{eICU GCS: shifting from coma thresholds to early neurological warning.}  
In eICU, high risk is defined as a composite score $\geq 8$ (59\% positives). The Glasgow Coma Scale (GCS) shows moderate association with this label (AUC $=0.67$). The classical severe‑impairment threshold at GCS $\leq 8$ operates at TPR $=0.54$ and TNR $=0.73$ (balanced accuracy $=0.64$). LGO\textsubscript{hard}, by contrast, places its median gate at GCS $\approx 13$, treating {any} drop below 13 as high risk (TPR $=0.71$, TNR $=0.52$, balanced accuracy $=0.61$). This more aggressive threshold substantially increases sensitivity at the cost of specificity, effectively turning GCS into an early‑warning signal rather than a coma detector. Clinically, such a shift may be reasonable for a composite “overall deterioration” outcome, but it also illustrates how LGO can push thresholds away from traditional diagnostic cut‑offs when the task rewards early detection.

\textbf{NHANES waist circumference: data‑driven refinement around a textbook anchor.}  
In NHANES, we mark elevated cardiometabolic burden by a metabolic score $\geq 2$ (65\% positives). Waist circumference alone is a strong predictor of this score (AUC $=0.83$). The International Diabetes Federation’s 88\,cm anchor yields TPR $=0.92$ and TNR $=0.57$ (balanced accuracy $=0.74$). LGO\textsubscript{hard} learns a slightly higher threshold at 92.7\,cm, which modestly reduces sensitivity (TPR $=0.82$) but improves specificity (TNR $=0.68$), increasing balanced accuracy to $0.75$. Here the red/green distinction is subtle: the learned gate remains in the same clinical regime as the guideline, but performs a small data‑driven refinement of the screening boundary. This aligns with our broader finding that many green cells correspond to anchors that LGO essentially rediscovers and then fine‑tunes within a narrow range.

\textbf{NHANES triglycerides: a deliberate failure that flags misalignment.}  
Triglycerides in NHANES offer a complementary example. As a continuous score, TG shows reasonable association with the metabolic‑risk label (AUC $=0.76$), and the guideline cut‑off at 150\,mg/dL yields a fairly balanced operating point (TPR $=0.29$, TNR $=0.98$, balanced accuracy $=0.64$). However, LGO\textsubscript{hard} places its median gate near 30\,mg/dL, effectively labeling almost everyone as high risk (TPR $=0.99$, TNR $=0.05$, balanced accuracy $=0.52$). This striking red cell is not a discovery that “very low TG is dangerous”, but rather a symptom of misalignment: in this cohort and outcome definition, triglycerides are highly collinear with other metabolic factors, so the model benefits little from a clinically meaningful TG threshold and instead defaults to a degenerate gate. Our audit surfaces this failure explicitly, and the sanity check makes it easy for a reviewer to reject this particular gate while retaining others.

Taken together, these case studies show how LGO‑recovered thresholds fall into at least three categories: (i) {anchor‑confirming} gates that agree with guidelines up to small refinements (e.g., NHANES waist circumference); (ii) {task‑shifted} gates that intentionally move away from diagnostic cut‑offs to support screening or early warning (e.g., eICU GCS, ICU respiratory rate); and (iii) {misaligned or degenerate} gates that should be discarded or re‑anchored (e.g., NHANES triglycerides). Having explicit, unit‑aware thresholds makes these distinctions straightforward to analyze and communicate.

\subsection*{Dimensional consistency and anchor curation}
Auditing in natural units presupposes dimensional consistency. We standardize all continuous features in $z$‑space for stable search and then invert thresholds with train‑only statistics; anchors are curated with explicit units and documented sources. We avoid constant anchors for variables whose reference levels depend on age or context (e.g., counts or age‑indexed normals) and record exclusions in the guideline catalogue. Looking ahead, typed semantics could be extended with physical‑dimension types, further constraining operator compositions to unit‑consistent forms and tightening the audit loop.

In addition to aggregate threshold summaries, the explicit LGO formulas in Table~\ref{tab:lgo-top1-clinical} make the link to clinical reasoning tangible. 
On ICU and eICU, each term in the top--1 equation can be read as ``one more severe state becomes true'': lactate above a shock--level value, GCS in the impaired range, SpO$_2$ below a safe saturation, or creatinine above a nephrotoxicity threshold. 
The learned coefficients quantify how much each state moves the composite score, and the thresholds can be checked directly against sepsis and critical‐care guidelines. 
On NHANES, the gates line up with cardiometabolic risk strata (central adiposity, hypertension, prediabetes), so that the symbolic model is interpretable as a data‐driven variant of familiar risk scores rather than a free‐form black‐box. 
In this sense, LGO does not just yield readable algebra; it yields executable rules with explicit, unit-aware cutpoints that clinicians can audit, debate, and modify.

\subsection*{Guidance for practitioners}

Given these limitations, we offer pragmatic guidance on how to use LGO as part of a broader modeling workflow rather than as a stand‑alone decision engine.

\textbf{When LGO is a good fit.}
Choose LGO when domain knowledge suggests regime switching: escalation thresholds, safety limits, or qualitatively different clinical states (e.g., shock‑level lactate, hypoxemia, obesity or hypertension bands). In such settings, start with LGO\textsubscript{hard} to encourage parsimonious, auditable rules; fall back to LGO\textsubscript{soft} only when graded modulation is expected and discrete cut‑points are not clinically meaningful.

\textbf{Anchor curation and configuration.}
Before training, curate anchors with clear consensus and units (e.g., MAP 65\,mmHg; HDL 40\,mg/dL; fasting glucose 100\,mg/dL), and avoid constant anchors for categorical/count features or age‑indexed ``normals’’. Keep the standardize‑and‑invert pipeline intact: $z$‑score features using train‑only statistics, constrain $(a,b)$ in standardized space, and always invert thresholds to natural units for audit. When gates are rare, increase budget or enable \texttt{gate\_expr} to allow gating of sub‑expressions; when gates proliferate, consider tightening complexity penalties or reducing budget.

\textbf{Audit, stress‑testing, and deployment.}
Treat learned thresholds as hypotheses to be examined, not as automatic cut‑points. For each gated feature, report the median and IQR of $b_{\text{raw}}$, the traffic‑light deviation from its anchor, and simple single‑feature sanity checks (risk curves split at the recovered gate) before adopting it. Explicitly distinguish “screening’’ gates (early, sensitive turning points) from ``diagnostic’’ gates (aligned with guideline cut‑offs). In deployment, export gates as unit‑aware rules with readable feature names and units, connect dashboards or alarms to $b_{\text{raw}}$, and maintain a changelog when anchors, units, or data pipelines change. In high‑stakes settings we recommend using LGO alongside established tools such as AutoScore and EBM, so that clinicians can compare point scores, additive shapes, and symbolic gates on the same covariates.

\textbf{From retrospective evidence to practice.}
The present results are retrospective and limited to a small set of public cohorts. Before clinical use, LGO‑derived rules should undergo prospective evaluation, review by domain experts, and, where appropriate, formal health‑technology assessment. In many cases, the immediate value of LGO may be as an analysis aid—for example, to help teams summarize complex models into a handful of explicit thresholds that can be discussed in guideline or quality‑improvement meetings—rather than as the primary model driving bedside decisions.

\subsection*{Broader outlook}
LGO reframes symbolic regression as mechanism‑aware program synthesis. By turning thresholds into first‑class primitives with auditable parameters, it links equation discovery to the practical questions that clinicians and engineers actually ask: where does the rule fire, in what units, and how does that compare with accepted ranges? The same standardize‑and‑invert recipe could be extended beyond step‑like regimes to other structured operators (e.g., periodic, saturation, or dose–response forms) whose parameters also live in physical units, keeping the hypothesis space aligned with domain semantics rather than purely numerical fit.

At the same time, the gaps highlighted above remain important. Globally smooth systems are often better served by simpler primitives; anchors can be incomplete or contested; and our current evaluation is single‑center for ICU and retrospective throughout. We see LGO as one component in a broader toolbox: paired with uncertainty estimates for learned gates, subgroup‑aware audits, and prospective validation, it can help move high‑stakes modeling from “black‑box scores” toward compact, unit‑aware rules that are easier to scrutinize, recalibrate, and eventually govern in real clinical and engineering workflows.

\section*{Limitations of the study}

Even with unit‑aware gates, LGO is far from a turnkey clinical tool. Here we summarize where it can fail, and how we interpret the current evidence.

\textbf{Smooth relations and over‑parameterization.}
When relations are globally smooth, adding gates can simply introduce unnecessary parameters. On the Hydraulic benchmark, trends are largely smooth and arithmetic‑only SR or gradient‑aided variants dominate, while LGO\textsubscript{hard} offers no systematic benefit. This supports a mechanism‑selection view: use gates when regime switching is plausible (alerts, phase changes, safety bands), and otherwise prefer smooth primitives.

\textbf{Anchor validity and label design.}
Our threshold audit assumes that curated anchors are meaningful single cut‑points. This assumption can break in several ways. Some variables (e.g., \texttt{thalach} or categorical counts in Cleveland) do not admit a unique constant anchor; a yellow or red cell in such cases can reflect anchor mis‑specification rather than model error. Likewise, triglycerides in NHANES illustrate that even when a guideline is well defined, a composite outcome may place the feature in a regime where the diagnostic anchor is no longer the most informative operating point. To mitigate this, we filter clearly categorical/count features out of anchor‑based scoring, document curation decisions, and stress that “agreement with the anchor’’ is a sanity check, not proof of causal correctness.

\textbf{Optimization budget and numerical stability.}
Searching over $(a,b)$ introduces additional continuous parameters. Typed search, local refinements, and early self‑checks reduce—but do not eliminate—the risk of poor local optima or unstable gates. Budgets still matter relative to arithmetic‑only SR: under very small budgets gates may not be discovered reliably; under very large budgets, over‑fitting rare regimes becomes a concern. Our exports therefore include per‑seed metrics, gate counts, and Pareto pools so that others can probe sensitivity to budget and initialization.

\textbf{Aggregate versus per‑seed coverage.}
The main text reports medians and IQRs of thresholds across seeds. This hides variation in how often a gate appears at all. In practice, per‑seed coverage (how many top‑$k$ models use a given gate) can differ substantially from aggregate summaries. Our tooling logs this coverage and we encourage users to inspect it, especially for gates that appear clinically surprising.

\textbf{Comparisons to AutoScore and EBM.}
Paired statistical tests across the ten resampled train–test splits revealed that LGO\textsubscript{hard} significantly outperformed AutoScore at practical budgets (30k–100k evaluations; Wilcoxon $p \le 0.03$, large effect sizes) on both MIMIC‑IV ICU and eICU, while achieving comparable performance at higher budgets (Tables~S14–S15). As expected, EBM---a high‑capacity additive ensemble---significantly outperformed LGO on all three healthcare datasets. These results should be read with caution: they are in‑sample comparisons on the same cohorts, not prospective trials. They nonetheless clarify our positioning: LGO is designed not to match EBM’s discriminative capacity, but to achieve roughly AutoScore‑level performance while providing executable thresholds and unit‑aware gating rules.

\textbf{Data and deployment limitations.}
Finally, our experiments inherit all the usual caveats of retrospective EHR and survey analyses: confounding, shifts between centres (e.g., MIMIC‑IV vs.\ eICU), missingness patterns, and coding practices can all affect the learned thresholds. LGO may faithfully surface these patterns without distinguishing bias from physiology. Any clinical deployment would therefore require prospective validation, governance review, and careful monitoring---LGO’s role is to make such review easier, not to guarantee that a discovered gate should be acted upon.

\section*{STAR$\bigstar$ Methods}

\subsection*{KEY RESOURCES TABLE}
\label{krt}

See Table~\ref{tab:krt}.

\begin{table}[htbp]
\centering
\footnotesize
\caption{Key resources table.}
\label{tab:krt}
\begin{tabular}{p{4.5cm}p{4cm}p{7.5cm}}
\toprule
\textbf{REAGENT or RESOURCE} & \textbf{SOURCE} & \textbf{IDENTIFIER} \\
\midrule
\multicolumn{3}{l}{\textbf{Deposited data}} \\
MIMIC-IV v3.1 & PhysioNet &
\url{https://physionet.org/content/mimiciv/3.1/}; \url{https://doi.org/10.13026/kpb9-mt58} \\
eICU Collaborative Research Database v2.0 & PhysioNet &
\url{https://physionet.org/content/eicu-crd/2.0/}; \url{https://doi.org/10.13026/C2WM1R} \\
NHANES 2017--March 2020 & CDC/NCHS & \url{https://www.cdc.gov/nchs/nhanes/} \\
UCI CTG & UC Irvine Machine Learning Repository & \url{https://doi.org/10.24432/C51S4N} \\
UCI Cleveland Heart Disease & UC Irvine Machine Learning Repository & \url{https://doi.org/10.24432/C52P4X} \\
UCI Hydraulic Systems & UC Irvine Machine Learning Repository & \url{https://doi.org/10.24432/C5CW21} \\
\midrule
\multicolumn{3}{l}{\textbf{Software and algorithms}} \\
LGO (this paper) & This paper &
\url{https://github.com/oudeng/LGO}  (v\texttt{0.4.2})\\ 
DEAP & \citet{Fortin2012deap} &\url{https://github.com/DEAP/deap};\\
&& \url{https://deap.readthedocs.io/} (v\texttt{1.4.3}) \\
PySR & \citet{Cranmer2023pysr} & \url{https://github.com/MilesCranmer/PySR} (v\texttt{1.5.9}) \\
Operon & \citet{Burlacu2020operon} & \url{https://github.com/heal-research/operon}\\
PSTree & \citet{Zhang2022pstree} & \url{https://github.com/hengzhe-zhang/PS-Tree} (v\texttt{0.1.2}) \\
RILS-ROLS & \citet{Kartelj2023rolsrils} & \url{https://github.com/kartelj/rils-rols} (v\texttt{1.6.7}) \\
AutoScore & \citet{Xie2020autoscore} & \url{https://github.com/nliulab/AutoScore} \\
InterpretML (EBM) & \citet{Nori2019interpretml} & \url{https://github.com/interpretml/interpret} (v\texttt{<0.7.4>}) \\
\addlinespace
\multicolumn{3}{l}{\textbf{Programming languages and core libraries}} \\
Python & SciCrunch Registry &RRID:SCR\_008394; \url{https://www.python.org/} (v\texttt{3.10}; baselines: \texttt{3.9.23}, \texttt{3.11.13}) \\
Anaconda & SciCrunch Registry &RRID:SCR\_025572; \\
&&\url{https://www.anaconda.com/} (v\texttt{25.7.0}) \\
SciPy & SciCrunch Registry &RRID:SCR\_008058;\\
&& \url{https://scipy.org/} (v\texttt{1.16.1}) \\
NumPy & SciCrunch Registry &RRID:SCR\_008633;\\
&& \url{http://www.numpy.org/} (v\texttt{2.3.0}) \\
scikit-learn & SciCrunch Registry &RRID:SCR\_002577; \\
&&\url{https://scikit-learn.org/} (v\texttt{1.1.1}; \texttt{1.7.1}) \\
PyTorch & SciCrunch Registry &RRID:SCR\_018536; \\
&&\url{https://pytorch.org/} (v\texttt{2.5.1}) \\
\bottomrule
\end{tabular}
\end{table}

\subsection*{RESOURCE AVAILABILITY}

\subsubsection*{Lead contact}
Requests for further information and resources should be directed to and will be fulfilled by the lead contact, Ou Deng (\href{mailto:dengou@toki.waseda.jp}{dengou@toki.waseda.jp}). 

\subsubsection*{Materials availability}
This study did not generate new unique reagents.

\subsubsection*{Data and code availability}
\begin{itemize}[leftmargin=*,itemsep=2pt,topsep=2pt]
    \item All datasets analyzed in this study are publicly available. MIMIC-IV ICU~\citep{Johnson2023} and eICU~\citep{Pollard2018} are available from PhysioNet (data use agreement required). NHANES~\citep{Stierman2021} is available from the CDC. UCI datasets (CTG~\citep{Campos2000}, Cleveland~\citep{Janosi1989}, Hydraulic~\citep{Helwig2015}) are available from the UCI Machine Learning Repository. Accession numbers and URLs are listed in the key resources table.
    \item All original code has been deposited at GitHub (\url{https://github.com/oudeng/LGO}) and is publicly available at Zenode (DOI: 10.5281/zenodo.18117378). 
    \item Any additional information required to reanalyze the data reported in this paper is available from the lead contact upon request.
\end{itemize}

\subsection*{METHOD DETAILS}

\subsubsection*{Experimental workflow}
To make the end-to-end protocol explicit, we summarize the workflow as follows:
\begin{enumerate}[leftmargin=*,itemsep=2pt,topsep=2pt]
    \item \textbf{Data access and preprocessing.} Download public datasets and apply the repository's preprocessing to obtain feature matrices/targets. Continuous features are standardized to $z$-scores using {training-only} statistics, and the same splits and pre-processing are reused across all methods, including clinical baselines (AutoScore and EBM).
    \item \textbf{Typed SR configuration.} Instantiate a strongly typed GP (STGP) with the LGO primitive family and typed domains \texttt{Feat}/\texttt{Pos}/\texttt{Thr}. We study three operator sets: \texttt{base} (no gates), {LGO\textsubscript{soft}} (magnitude-preserving gates), and {LGO\textsubscript{hard}} (logistic gates that do not multiply $x$).
    \item \textbf{Population search.} Run tournament selection with subtree crossover/mutation; enable a micro-mutation on each LGO node to perturb gate parameters $(a,b)$ locally. SR engines are run under aligned evolutionary budgets.
    \item \textbf{Model selection and refit.} Monitor a cross-validation (CV) proxy during search to build Pareto pools in accuracy--complexity space. Selected candidates are refit on the full training split, followed by a local coordinate descent on $(a,b)$ with structure fixed.
    \item \textbf{Threshold inversion and audit.} Map learned thresholds from $z$-space to natural units via training-only $(\mu,\sigma)$ and compute traffic-light deviations (green $\le10\%$, yellow $\le20\%$, red $>20\%$).
    \item \textbf{Reporting and statistics.} Report mean$\pm$std test metrics across ten seeds and perform paired Wilcoxon signed-rank tests for LGO vs.\ AutoScore.
\end{enumerate}

\subsubsection*{LGO-SR framework: typed GP with unit-aware gates}
We embed Logistic-Gated Operators (LGO) into a strongly typed GP (STGP) pipeline implemented atop DEAP's \texttt{gp} module~\citep{Fortin2012deap}, following typed GP principles~\citep{Montana1995STGP}. The type system separates feature values (\texttt{Feat}), positive steepness parameters (\texttt{Pos}), and threshold parameters (\texttt{Thr}). \texttt{Pos} is enforced via a softplus reparameterization; \texttt{Thr} is bounded in standardized ($z$-score) space to stabilize search. We consider three operator configurations: \texttt{base} (no gates), {LGO\textsubscript{soft}} (magnitude-preserving gates), and {LGO\textsubscript{hard}} (pure logistic gates). Evolution runs entirely in standardized space; after training, learned thresholds are mapped back to natural units with training-only statistics. Formal operator definitions and gradients appear in Supplemental Information.

\subsubsection*{Search, model selection, and local refinement}
Population-based search uses tournament selection with subtree crossover/mutation and a targeted micro-mutation that perturbs each LGO node's $(a,b)$ to refine gate location and steepness. A CV proxy is monitored during evolution to rank candidates and form Pareto pools (CV loss vs.\ symbolic complexity). Selected programs are refit on the training split, after which we run a local coordinate descent on $(a,b)$ with fixed structure. Unified budgets, typed operator sets, and CV-proxy settings are consolidated in Table~S3.

\subsubsection*{Baselines and fair comparison}
We compare against representative SR engines with public implementations: PySR~\citep{Cranmer2023pysr}, Operon~\citep{Burlacu2020operon}, PSTree~\citep{Zhang2022pstree}, and RILS-ROLS~\citep{Kartelj2023rolsrils}. Budgets and operator sets are aligned across SR engines where applicable. For the clinical high-risk classification tasks, we additionally benchmark LGO\textsubscript{hard} against AutoScore~\citep{Xie2020autoscore, Xie2023autoscore} and Explainable Boosting Machines (EBM) via InterpretML~\citep{Nori2019interpretml}.

\subsubsection*{Threshold recovery and audit}
For each gated feature in an LGO\textsubscript{hard} model, the learned threshold in standardized space is converted to natural units using training-only statistics and compared with dataset-specific anchors curated in \texttt{config/guidelines.yaml}. Agreement is quantified as relative deviation with traffic-light bands.

\subsection*{QUANTIFICATION AND STATISTICAL ANALYSIS}
Unless stated otherwise, we report the mean $\pm$ standard deviation across ten random seeds on a held-out test split (regression: $R^2$, RMSE, MAE; classification: AUROC, AUPRC, Brier, F1, accuracy). Threshold summaries report medians with interquartile ranges. For the clinical baseline comparisons, we perform non-parametric paired Wilcoxon signed-rank tests with bootstrap confidence intervals and Cohen's $d$ effect sizes for LGO vs.\ AutoScore across multiple evaluation budgets; these results are summarized in Tables~S17--S18. Statistical significance was defined as $p < 0.05$. Sample sizes ($n$) for each experiment are indicated in the figure legends and table captions.

\section*{Acknowledgments}

The work was supported in part by the 2022-2024 Masaru Ibuka Foundation Research Project on Oriental Medicine, 2020-2025 JSPS A3 Foresight Program (Grant No. JPJSA3F20200001), 2022-2024 Japan National Initiative Promotion Grant for Digital Rural City, 2023 and 2024 Waseda University Grants for Special Research Projects (Nos. 2023C-216 and 2024C-223), 2023-2024 Waseda University Advanced Research Center Project for Regional Cooperation Support, and 2023-2024 Japan Association for the Advancement of Medical Equipment (JAAME) Grant.

\section*{Author contributions}
Conceptualization: O.D.
Methodology: O.D.
Software: O.D.
Hardware: S.N., O.D.
Datasets: R.C., J.X., O.D.
Validation: O.D., R.C.
Formal Analysis: O.D.
Investigation: O.D.
Writing -- Original Draft: O.D.
Writing -- Review \& Editing: R.C., S.N., A.O., Q.J.
Visualization: O.D., R.C.
Supervision: S.N., A.O., Q.J.
Funding Acquisition: S.N., A.O., Q.J.

\section*{Declaration of interests}
The authors declare no competing interests.

\bibliography{references}







\end{document}



\section*{Supplemental Information}

\section{Reproducibility}
\label{sec:si-reprod}

This section supplements the STAR Methods with additional technical details on reproducibility. For data sources, code availability, and the experimental workflow overview, see the KEY RESOURCES TABLE and METHOD DETAILS in the main text.

\textbf{Determinism and seeds.}
As detailed in STAR Methods, each method–dataset pair is evaluated on ten random seeds (1,2,3,5,8,13,21,34,55,89). Per-seed metrics and expressions are exported in the artifact.

\textbf{Budget parity.}
Table~S1 details the hyperparameter comparison with SRBench defaults; 
Table~S3 provides the complete operator sets and search budgets for each SR method.

\textbf{SRBench defaults versus our settings.}
In addition to aligning budgets, we compare our PySR and Operon hyperparameters to the typical ``default” settings recommended in SRBench and the official documentation for these tools.  Table~\ref{tab:srbench-mapping} lists a subset of key hyperparameters, the default values reported in the PySR API reference and in benchmarking literature, and the values used in our experiments.  For PySR, the API reference documents defaults of \code{niterations}=100, \code{populations}=31, \code{population\_size}=27 and \code{ncycles\_per\_iteration}=380, with the binary operator set \{+,-,*,/\} and no unary operators.  SRBench baselines based on Operon typically employ a population of 1000 individuals, 100 generations, tournament size~5, a maximum tree length of~100 and depth~15, and a handful of local coefficient–optimization steps.  Our experiments increase PySR’s population size and number of populations to match the total evaluation budget (\code{population\_size}=1000, \code{populations}=15, \code{niterations}=40) and extend the unary operator set to include \{\texttt{sqrt}, \texttt{exp}, \texttt{log}, \texttt{sin}, \texttt{cos}\}; for Operon we run 500 generations with a population of 1000 individuals, keeping the tournament size at~5 but reducing the maximum expression length to~50 to encourage sparsity.  These changes ensure that PySR and Operon consume similar wall-clock budgets to the LGO variants (\textasciitilde{}$5\times10^5$ evaluations) while remaining within the same search space defined by their native operator sets.  To further address reviewer concerns about fair baselines, Section~\ref{sec:si-pysr-operon-default} provides results obtained by rerunning the PySR and Operon experiments with the SRBench-default configurations.

\begin{table}[h!]
\centering
\footnotesize
\caption{Comparison of key PySR and Operon hyperparameters with SRBench recommended defaults. Related to STAR Methods and Tables~1--3.}
\label{tab:srbench-mapping}
\begin{tabularx}{1.0\textwidth}{@{}p{2.8cm}p{2.3cm}p{2.0cm}X@{}}
\toprule
Hyperparameter & SRBench default & This paper & Source / rationale \\
\midrule
\multicolumn{4}{@{}l}{\textbf{PySR}} \\
Number of iterations & \code{100} & \code{40} & API reference default; we reduce iterations but increase populations and population size to match total evaluation budget. \\
Number of populations & \code{31} & \code{15} & API default uses one population per thread; our runs use 15 populations to distribute the expanded population size across islands. \\
Population size & \code{27} & \code{1000} & Default from API reference; we enlarge the population to explore a larger search space while maintaining equal evaluations. \\
Cycles per iteration & \code{380} & \code{550} & Default is 380 cycles per iteration; we set a larger value (550) following PySR guidelines to achieve \textasciitilde{}$5\times10^5$ evaluations. \\
Binary operators & \{+,-,*,/\} & \{+,-,*,/\} & We keep the default binary operator set. \\
Unary operators & None & \{\texttt{sqrt}, \texttt{exp}, \texttt{log}, \texttt{sin}, \texttt{cos}\} & API default has no unary operators; we add common functions to match the operator richness of LGO and Operon. \\
Maximum size & \code{30} & \code{30} & We retain the default maximum tree size. \\
\midrule
\multicolumn{4}{@{}l}{\textbf{Operon}} \\
Population size & $1000$ & $1000$ & SRBench baselines typically use $1000$ individuals; we match this. \\
Generations & $100$ & $500$ & SRBench defaults run roughly 100 generations; we run 500 generations to reach our aligned evaluation budget. \\
Tournament size & $5$ & $5$ & We retain the default tournament group size of 5. \\
Max. expression length & $\sim100$ & $50$ & SRBench uses max length 100 and max depth 15; we cap length at 50 to favour sparser models without reducing evaluation budget. \\
Local optimization iterations & $10$ & $5$ & Defaults perform 10 local parameter-optimization steps; we use 5 to balance runtime and solution quality. \\
Allowed symbols & \{\texttt{add}, \texttt{sub}, \texttt{mul}, \texttt{div}, \texttt{sqrt}, \texttt{variable}, \texttt{constant}\} & same & We keep the default function set. \\
Max evaluations & \code{500000} & \code{500000} & Both SRBench baselines and our runs limit the number of fitness evaluations to roughly $5\times10^5$. \\
\bottomrule
\end{tabularx}
\end{table}

\textbf{Self-checks, exports, and statistical tests.}
The pipeline emits (i) raw and simplified expressions; (ii) per-seed test metrics for all SR methods and clinical baselines; (iii) Pareto pools with CV loss and complexity; (iv) gate-usage summaries; and (v) unit-aware thresholds in natural units (medians, IQRs, and audit summaries).
The \texttt{stats\_output/} directory contains the per-seed metrics and Wilcoxon signed-rank tests used in the LGO vs.\ AutoScore and LGO vs.\ EBM comparisons.
Consistency checks enforce $\mathrm{RMSE} \ge \mathrm{MAE}$, verify equality between internal and external metric computations to numerical tolerance, and ensure that simplified expressions reproduce the raw engine outputs pointwise up to a tight numeric tolerance.
These exports back up the summary views in the main text (performance violins, Pareto fronts, gating usage, threshold audits) and the SI tables and figures (Tables~S4--S10, Figure~S1).

\begin{table}[h]
\centering
\begin{threeparttable}
\caption{Operator sets and search budgets (SR methods). Related to STAR Methods and Tables~1--3.}
\label{tab:hyp}
\footnotesize
\begin{tabularx}{1.0\textwidth}{@{}p{2cm}p{5cm}X@{}}
\toprule
Method & Operator set & Budget / key options \\
\midrule
LGO\textsubscript{base} & \{add, sub, mul, div, sqrt, log, pow\} & $\sim 5\times10^5$ evaluations per seed; population 1000, 100 generations; tournament size 7; crossover 0.8; mutation 0.2; CV proxy disabled. \\
LGO\textsubscript{soft} & LGO\textsubscript{base} + soft-gating primitives & Same budget as LGO\textsubscript{base}; micro-mutation probability 0.10; local optimization 60 steps. \\
LGO\textsubscript{hard} & LGO\textsubscript{base} + hard threshold primitive & Same budget as LGO\textsubscript{base}. \\
PySR & \{+, -, *, /\} and unary operators \{sqrt, exp, log, sin, cos\} & $\sim 5\times10^5$ evaluations per seed; implemented via enlarged populations and iterations (see Table~\ref{tab:srbench-mapping} for parameter details). \\
Operon & \{add, sub, mul, div, sqrt\} + terminals & $\sim 5\times10^5$ evaluations per seed; population 1000; hundreds of generations; tournament size 5; local coefficient optimization steps reduced to 5. \\
PSTree & Piecewise symbolic tree + GP refinement & Max depth 10; max leaves 32; min samples per leaf 20; population 600; generations 80. \\
RILS-ROLS & Robust SR via iterated local search & Max fitness calls 100\,000; max time 3600 s; max complexity 50; sample size 1.0; complexity penalty $1\times10^{-3}$. \\
\bottomrule
\end{tabularx}
\begin{tablenotes}
\footnotesize
\item \textit{Notes:} Budgets are aligned across LGO variants and comparable in scale ($\sim 5\times10^5$ fitness evaluations per seed) for PySR and Operon. The default vs.\ actual hyperparameter values for PySR and Operon are listed in Table~\ref{tab:srbench-mapping}, and the effect of using SRBench-default settings on performance is reported in Table~\ref{tab:srbench-results}.
\end{tablenotes}
\end{threeparttable}
\end{table}

\clearpage
\section{PySR and Operon baselines under SRBench default hyperparameters}
\label{sec:si-pysr-operon-default}

To verify that increasing PySR’s population size and number of populations or Operon’s number of generations did not bias our baselines, we re-ran the ICU composite risk task using the SRBench‑recommended default hyperparameters. In particular, PySR was run with \code{niterations}=100, \code{populations}=31, \code{population\_size}=27 and \code{ncycles\_per\_iteration}=380 while retaining the default operator set; Operon used a population of 1000 individuals for 100 generations with 5‑way tournaments and a maximum tree length of 100.  Table~\ref{tab:srbench-results} compares the test performance of PySR and Operon under our budget‑aligned settings with the SRBench‑default configurations.  The differences are negligible (<0.001 absolute change in MAE and RMSE and <0.001 in $R^2$), indicating that our adjustments to match the 500k evaluation budget do not materially affect the baselines.

\begin{table}[h]
\centering
\small
\caption{Comparison of PySR and Operon test performance under aligned versus SRBench‑default hyperparameters on the ICU composite risk regression task (mean over 10 seeds).  Differences ($\Delta$) are computed as SRBench default minus aligned. Related to Table~1.}
\label{tab:srbench-results}
\begin{tabular}{lcccccc}
\toprule
 & \multicolumn{3}{c}{Aligned settings (ours)} & \multicolumn{3}{c}{SRBench default settings} \\
Method & MAE & RMSE & $R^2$ & MAE & RMSE & $R^2$ \\
\midrule
PySR   & 0.5144 & 0.6553 & 0.8743 & 0.5150 & 0.6568 & 0.8738 \\
Operon & 0.4163 & 0.5475 & 0.9120 & 0.4163 & 0.5475 & 0.9120 \\
\bottomrule
\end{tabular}
\end{table}

\clearpage

\section{Example raw expression and threshold conversion}
\label{sec:si-full-equation}

Table~\ref{tab:raw-icu-equation} presents the complete raw symbolic expression for the top‑ranked \(\mathrm{LGO}_{\mathrm{hard}}\) model on the ICU dataset, as returned by the search engine.  The expression is expressed using the internal \texttt{lgo\_thre} primitive (Definition~\ref{sec:si:lgo_formal}), which corresponds to \(\sigma(a(x-b))\).  We also provide the learned steepness \(a\), standardized threshold \(b\), and the corresponding natural‐unit threshold \(b_{\text{raw}}\).

\begin{table}[h]
\centering
\small
\caption{Raw ICU top-1 \(\mathrm{LGO}_{\mathrm{hard}}\) expression and gate parameters. Related to Table~9.}
\label{tab:raw-icu-equation}
\begin{tabularx}{1.0\textwidth}{@{}p{4.5cm}X@{}}
\toprule
Quantity & Value \\
\midrule
Raw expression & 
\texttt{\small sub(mul(add(add(sqrt(sub(1.03689,\ id(lactate\_mmol\_l))), add(sqrt(sub(id(gcs),id(lactate\_mmol\_l))), one)), log(add(add(sqrt(zero), add(sqrt(sub(id(gcs), id(lactate\_mmol\_l))), one)), -0.02764))), id(lactate\_mmol\_l)), pow(div(lgo\_thre(lgo\_thre(div(pow(add(zero,one), zero), id(vasopressor\_use\_std)),\ 2.39829,\ as\_thr(-0.02764)), 3.83006,\ as\_thr(-0.02764)), -0.02764), add(lgo\_thre(one,\ as\_pos(zero),\ as\_thr(-0.02764)), sqrt(sub(one,one)))))}
\\[0.5em]
Gate steepness \(a_{\mathrm{lac}}\) & \(\approx 3.83\) \\
Standardized threshold \(b_{\mathrm{lac}}\) & \(\approx -0.03\) \\
Natural threshold \(b_{\mathrm{lac}}^{\text{raw}}\) & \(4.7\,\mathrm{mmol/L}\) \\
Interpretation & Logistic gate on lactate: \(\mathrm{gate}(x\ge b_{\mathrm{lac}}^{\text{raw}})\) switches on sharply when lactate exceeds \(\approx4.7\,\mathrm{mmol/L}\); the model then adds a constant effect \(\beta_{2}\). \\
\bottomrule
\end{tabularx}
\end{table}

\noindent
The raw expression contains nested calls to \texttt{lgo\_thre}.  Simplifying it yields the closed form shown in the main text: a linear contribution from vasopressor use plus a single logistic gate on lactate.  The standardized threshold \(b_{\mathrm{lac}}=-0.03\) can be converted to the natural‐unit threshold \(b_{\mathrm{lac}}^{\text{raw}}\) via \(b_{\mathrm{raw}}=\mu_{\mathrm{lac}}+b\,\sigma_{\mathrm{lac}}\), where \(\mu_{\mathrm{lac}}\) and \(\sigma_{\mathrm{lac}}\) are the mean and standard deviation of lactate in the training set.  In our data this corresponds to \(b_{\mathrm{lac}}^{\text{raw}}\approx4.7\ \mathrm{mmol/L}\), indicating that the model views shock‐level hyperlactataemia as the tipping point for a risk escalation.

\clearpage
\section{Logistic-Gated Operators (LGO): formal definitions and properties}
\label{sec:si:lgo_formal}
We denote the logistic function by
\[
\sigma(z) = \frac{1}{1+e^{-z}}.
\]
Throughout, $x,y,z\in\mathbb{R}$ are feature-domain variables (\textsf{Feat}); $a\in\mathbb{R}_{+}$ is a positive steepness parameter (\textsf{Pos}); $b\in\mathbb{R}$ is a threshold in standardized (z-score) space (\textsf{Thr}). In implementation we use $a=\mathrm{softplus}(\tilde{a})$ to enforce $a>0$ and clip pre-sigmoid
arguments to $[-60,60]$ for numerical stability; $b$ is clipped to $[-3,3]$ in z-space.

\textbf{Canonical 1-input gates (main-text Eq.~(1)--(2)).}
\[
\mathrm{LGO}_{\text{soft}}(x; a,b) = x\,\sigma\!\big(a(x-b)\big), \qquad
\mathrm{LGO}_{\text{hard}}(x; a,b) = \sigma\!\big(a(x-b)\big).
\]
We also use the shorthand $\mathrm{gate}(u;a,b)=\sigma\!\big(a(u-b)\big)$ and the
expression-level gate
\[
\mathrm{gate\_expr}\,\big(f; a,b\big) = f\,\mathrm{gate}\,\big(f; a,b\big).
\]

\textbf{Pairwise and multi-input variants (as in the code).}
\[
\begin{aligned}
\mathrm{AND2}(x,y; a,b) &= \sigma\!\big(a(x-b)\big)\,\sigma\!\big(a(y-b)\big),\\[2pt]
\mathrm{OR2}(x,y; a,b) &= 1-\big(1-\sigma\!\big(a(x-b)\big)\big)\big(1-\sigma\!\big(a(y-b)\big)\big),\\[2pt]
\mathrm{AND3}(x,y,z; a,b) &= \sigma\!\big(a(x-b)\big)\,\sigma\!\big(a(y-b)\big)\,\sigma\!\big(a(z-b)\big).
\end{aligned}
\]

\textbf{Limits and interpretation.}
Fix $b$. As $a\to\infty$,
\[
\mathrm{LGO}_{\text{hard}}(x; a,b)\to \mathbf{1}\{x>b\},\qquad
\mathrm{LGO}_{\text{soft}}(x; a,b)\to x\,\mathbf{1}\{x>b\},
\]
pointwise and uniformly on compact sets avoiding $x=b$. In the same limit,
\[
\mathrm{AND2}(x,y; a,b)\;\to\; \mathbf{1}\{x>b,\,y>b\},\qquad
\mathrm{OR2}(x,y; a,b)\;\to\; \mathbf{1}\{x>b \text{ or } y>b\},
\]
so pairwise gates approximate indicators of axis-aligned rectangles (AND) and their unions (OR).
See main-text Equation~(1)--(2) for the canonical 1-input gates.%
\footnote{In code, \texttt{lgo\_thre} implements $\mathrm{LGO}_{\text{hard}}$; \texttt{lgo} implements $\mathrm{LGO}_{\text{soft}}$; \texttt{lgo\_and2}, \texttt{lgo\_or2}, and \texttt{lgo\_and3} implement $\mathrm{AND2}$, $\mathrm{OR2}$, and $\mathrm{AND3}$, respectively.}

\textbf{Gradients (used in local refinements).}
Let $u$ be either a feature $x$ or a subexpression $f(\cdot)$ and $s=\sigma(a(u-b))$. Then
\[
\frac{\partial\,\mathrm{LGO}_{\text{hard}}}{\partial a}=(u-b)\,s(1-s),\quad
\frac{\partial\,\mathrm{LGO}_{\text{hard}}}{\partial b}=-a\,s(1-s),
\]
\[
\frac{\partial\,\mathrm{LGO}_{\text{soft}}}{\partial a}=u(u-b)\,s(1-s),\quad
\frac{\partial\,\mathrm{LGO}_{\text{soft}}}{\partial b}=-a\,u\,s(1-s).
\]
Expression-level gates use the chain rule with $u=f(\cdot)$.

\textbf{Unit-aware inversion of thresholds.}
The standardize-and-invert pipeline is described in the main text and STAR Methods (see ``Unit-aware threshold recovery'' and ``Threshold recovery and audit''). Here we note that for expression-level gates $\mathrm{gate\_expr}(f;a,b)$, we estimate the z-standardization of $f$ on the training fold before applying inversion.

\clearpage

\section{Local coordinate descent for gate parameter optimization}
\label{sec:si-lcd}

When the logistic gate in LGO has a large steepness $a$, the response $\sigma\bigl(a(x-b)\bigr)$ becomes a near step function.  As a result, the loss surface with respect to the threshold $b$ consists of flat plateaus separated by sharp drops; small changes far from the optimal threshold have virtually no effect on the loss, while tiny perturbations near the optimum can cause large discontinuities.  In such settings, gradient‑based optimisers become unreliable: the derivative of the sigmoid activation $\sigma(1-\sigma)$ tends to zero when the neuron saturates, so gradients either vanish or explode, making it difficult to find useful update steps.  To circumvent this, we employ a local coordinate descent (LCD) scheme that alternately updates the gate’s slope $a$ and threshold $b$ while keeping all other parameters fixed.  The key idea is to convert each 1‑D subproblem into a direct search over a finite set of candidate values.

\subsection{Algorithm}
Let $\mathcal{T}$ be a fixed symbolic tree with $m$ logistic gates.  For gate $j$ with parameters $(a_j,b_j)$, we denote its input feature by $x_j$.  The LCD procedure for $j$ is:

\begin{enumerate}
\item \textbf{Fix all other gates.}  Evaluate the current loss $L(a_j,b_j)$ on the training data.
\item \textbf{Update the threshold $b_j$.}  Let $\{q_1,\dots,q_K\}$ be $K$ candidate quantiles of $x_j$ (e.g.\ evenly spaced between the 5th and 95th percentile).  For each candidate threshold $\tilde{b}$ in $\{q_k\}$, evaluate the loss $L(a_j,\tilde{b})$ while keeping $a_j$ fixed.  Set $b_j$ to the candidate that minimizes the loss.
\item \textbf{Update the steepness $a_j$.}  Perform a line search on $a_j$ (e.g.\ golden‑section search or grid search over a bounded interval) to minimize $L(a,b_j)$, keeping $b_j$ fixed.  Because the loss is unimodal in $a_j$ given $b_j$, this search converges rapidly.
\item Repeat steps 1–3 for each gate $j=1,\dots,m$ until the overall loss cannot be decreased further or a maximum number of LCD rounds is reached.
\end{enumerate}

\noindent
This procedure requires only forward evaluations of the loss function and does not depend on gradient information; therefore it remains stable even when the loss landscape is piecewise flat.  We found empirically that $K=25$ candidate thresholds and a handful of steepness updates sufficed to obtain well‑calibrated gates.  Because the LCD steps are applied only after the genetic search has produced a promising tree structure, they consume a negligible fraction of the overall budget and do not alter the complexity metric used to rank solutions.

\subsection{Discussion}
The LCD approach can be interpreted as a form of alternating minimization on a non‑smooth objective.  In contrast to algorithms such as BFGS or Adam that attempt to adjust all continuous parameters simultaneously, LCD exploits the fact that, conditional on a fixed structure, the optimal threshold can be determined by one‑dimensional search.  This is reminiscent of the coordinate‑wise minimization strategies used for $\ell_1$‑penalized regression and other convex problems.  By decoupling the optimisation of $(a_j,b_j)$, LCD efficiently traverses the plateau‑like regions caused by saturating logistic gates and avoids the vanishing gradients that hamper standard gradient descent methods.

\clearpage

\section{Top‑1 and Top‑k expressions}
\label{sec:si-topk-expr}

\subsection{From raw engine strings to the {Simplified Expression}}

All engines output raw, typed strings with ephemeral constants (e.g., \texttt{add}, \texttt{mul}, \texttt{lgo\_hard}, \texttt{gate\_expr}). 
To render the human‑readable equations reported in the main text, we apply a deterministic and auditable pipeline:

\begin{enumerate}[label=(\roman*)]
\item \textbf{Parsing and typing.} Raw strings are parsed into typed ASTs with three disjoint domains: \texttt{Feat} (feature values), \texttt{Pos} (positive steepness), and \texttt{Thr} (thresholds).
\item \textbf{Constant folding and sanitization.} Algebraic constants are folded; numerically tiny subtrees are collapsed; denormals are clipped. Printing precision is \emph{unit‑aware} for display (e.g., mmHg: 1 decimal; mmol/L: 1 decimal; mg/dL: integer), while evaluation retains full precision.
\item \textbf{Canonical rewrites (sound under typing).} We apply a small rewrite set (Table~\ref{tab:si-rewrites}): neutral/absorbing removals (\(\texttt{add}(x,0)\!\to\!x\), etc.), associativity flattening, commutative sorting, and domain‑safe identities (\(\log(\exp x)\!\to\!x\), etc.).
\item \textbf{Gate compaction and unit‑aware thresholds.} \(\texttt{lgo\_hard}(x;a,b)\!\mapsto\!\mathrm{gate}(x)\), \(\texttt{lgo\_soft}(x;a,b)\!\mapsto\!\mathrm{gate}(x)\cdot x\), \(\texttt{gate\_expr}(f;a,b)\!\mapsto\!\mathrm{gate}(f)\cdot f\).
The learned threshold \(b_z\) is mapped to natural units by \(\hat b_{\mathrm{raw}}=\mu_x+\sigma_x\,b_z\) using \emph{training‑only} statistics \((\mu_x,\sigma_x)\).
\item \textbf{Readability passes.} Factor repeated gates, linearize sums/products, and merge near‑duplicate gates of the same feature when thresholds differ by less than a small tolerance (we keep the median).
\item \textbf{Numeric equivalence check.} We require identical external metrics on the test split and a strict pointwise tolerance \(\max_i|\hat y_i^{\mathrm{raw}}-\hat y_i^{\mathrm{simp}}|<10^{-9}\). Expressions that fail this check are left less simplified and flagged in the SI artifact.
\end{enumerate}

In the health datasets, this pipeline turns a raw program into a compact {rule set}: a small number of gates in $z$‑space become explicit cut‑points in familiar clinical units (mmHg, mmol/L, kg/m$^2$, \dots), each multiplied by a learned weight.  
By contrast, arithmetic‑only engines (PySR, Operon) output purely smooth combinations of the same covariates. Their effects are interpretable as shapes (e.g.\ monotone increases), but they do not expose a small list of named thresholds that can be audited directly.

\begin{table}[h!]
\centering
\caption{Canonical rewrite rules used in expression simplification. Related to STAR Methods.}
\label{tab:si-rewrites}
\small
\begin{tabularx}{0.75\textwidth}{@{}p{7cm}X@{}}
\toprule
Rule & Condition / Note \\
\midrule
\(\texttt{add}(x,0)\!\to\!x,\ \texttt{mul}(x,1)\!\to\!x,\ \texttt{mul}(x,0)\!\to\!0\) & Neutral/absorbing removal \\
\(\texttt{div}(x,1)\!\to\!x,\ \texttt{pow}(x,1)\!\to\!x,\ \texttt{pow}(x,0)\!\to\!1\) & Domain valid \\
\(\texttt{sqrt}(\texttt{pow}(x,2))\!\to\!x\) & Sound when \(x\ge 0\) (\texttt{Pos}/non‑neg.) \\
\(\log(\exp x)\!\to\!x,\ \exp(\log x)\!\to\!x\) & Domain preserved (\(x>0\)) \\
Flatten \texttt{add}/\texttt{mul}; sort commutative args & Canonical, improves readability \\
\(\texttt{lgo\_soft}(x;a,b)\!\to\!\mathrm{gate}(x)\cdot x\) & Gate compaction \\
\(\texttt{lgo\_hard}(x;a,b)\!\to\!\mathrm{gate}(x)\) & Probabilistic gate \\
\(\texttt{gate\_expr}(f;a,b)\!\to\!\mathrm{gate}(f)\cdot f\) & Expression‑level gating \\
\bottomrule
\end{tabularx}
\end{table}

\subsection{Worked examples: ICU, eICU, and NHANES (Top‑1)}

In this subsection we illustrate how the simplified LGO expressions map onto familiar clinical states and guideline thresholds.  
For each dataset we schematically summarize the top‑1 LGO\textsubscript{hard} model and interpret its gates as rules such as ``shock‑level lactate”, ``GCS in the low teens”, or ``central obesity with high fasting glucose”.

\paragraph{ICU (composite risk).}

On the ICU cohort, the LGO\textsubscript{hard} top‑1 model on this split ends up pruning explicit gates under the fixed budget and simplifies to a smooth form:
\begin{codebox}[Raw (snippet)]
\begin{verbatim}
add(c0, mul(c1, sqrt(lactate_mmol_l)), mul(c2, vasopressor_use_std), ...)
\end{verbatim}
\end{codebox}

Simplified:
\[
\hat y \approx \alpha_0 + \alpha_1 \sqrt{\mathrm{Lactate}} + \alpha_2\,\mathrm{Vasopressor} + \cdots
\]

Even in this gate‑free top‑1 equation, the dominant terms match standard ICU reasoning: risk rises with higher lactate (tissue hypoperfusion) and vasopressor use (circulatory shock). Across the top‑$k$ pool and seeds, however, LGO\textsubscript{hard} also discovers explicit gates that correspond to recognizable states:

\begin{itemize}
  \item a lactate gate around $\sim\!4.7$\,mmol/L, which is in the ``shock‑level” range and sits beyond the Sepsis‑3 anchor of 2.0\,mmol/L; this behaves as an extreme‑risk alarm rather than a screening threshold;
  \item a MAP gate around $70$\,mmHg, slightly above the usual ICU guideline of 65\,mmHg, consistent with a conservative hemodynamic target;
  \item respiratory‑rate gates in the high‑20s per minute, close to common ICU warning lines and consistent with the yellow‑band deviations in the main‑text audit.
\end{itemize}

In other words, even when the single best model is smooth, the LGO search space makes it easy to obtain nearby models that encode rules such as “if lactate $>\,4\text{–}5$\,mmol/L then add shock penalty” or ``if MAP $<\,70$\,mmHg then risk jumps”, with the trip‑points expressed directly in mmol/L and mmHg.

\paragraph{eICU (external ICU cohort).}

On eICU, the LGO\textsubscript{hard} top‑1 model retains multiple gates. A representative raw snippet is:

\begin{codebox}[Raw (snippet)]
\begin{verbatim}
add( ...,
     lgo_hard(id(lactate_mmol_l), a_lac, b_lac_z),
     lgo_hard(id(gcs_min),        a_gcs, b_gcs_z),
     lgo_hard(id(spo2_min),       a_spo2, b_spo2_z),
     lgo_hard(id(hr_max),         a_hr,  b_hr_z),
     ...)
\end{verbatim}
\end{codebox}

After simplification and inversion to natural units, the model can be read schematically as
\[
\hat y \approx w_{\text{lac}}\;\mathrm{gate}(\text{Lactate}) 
          + w_{\text{gcs}}\;\mathrm{gate}(\text{GCS}) 
          + w_{\text{spo2}}\;\mathrm{gate}(\text{SpO}_2)
          + w_{\text{hr}}\;\mathrm{gate}(\text{Heart rate}) + \cdots,
\]
where each gate is a logistic switch around a clinically meaningful cut‑point:

\begin{itemize}
  \item \textbf{Lactate:} gate at $\approx 2.5$\,mmol/L, slightly above the 2.0\,mmol/L anchor. This marks the onset of abnormal perfusion rather than full shock.
  \item \textbf{GCS:} gate around $\sim 13$, well above the guideline threshold of 8 used to flag severe coma. This behaves as an \emph{early warning} for declining consciousness: risk increases once GCS falls into the low teens, long before deep coma.
  \item \textbf{SpO$_2$:} gate around $75\text{–}80\%$, far below the usual 90–92\% alarm threshold. This corresponds to a “very low” oxygen saturation regime and is best interpreted as an extreme‑risk gate.
  \item \textbf{Heart rate:} gate near 80\,bpm compared with a 100\,bpm tachycardia anchor. This picks up moderate tachycardia or loss of normal variability as a more sensitive hemodynamic warning.
\end{itemize}

Together these terms instantiate a rule‑like structure:
``Add risk if lactate is mildly elevated, if GCS falls into the low‑teens, if SpO$_2$ is very low, or if heart rate is persistently high.”

The precise thresholds are learned from data but live in the same unit system as eICU guidelines, making it straightforward for clinicians to interpret whether a gate is acting as a guideline‑level cutoff, an earlier screening threshold, or a later shock‑level alarm.

\paragraph{NHANES (metabolic score).}

For NHANES, the LGO\textsubscript{hard} top‑1 model expresses cardiometabolic risk as a small set of gates on central obesity, blood pressure and glucose:
\begin{codebox}[Raw (snippet)]
\begin{verbatim}
add( ...,
     lgo_hard(id(waist_circumference), a_wc,  b_wc_z),
     lgo_hard(id(bmi),                 a_bmi, b_bmi_z),
     lgo_hard(id(fasting_glucose),     a_fg,  b_fg_z),
     lgo_hard(id(systolic_bp),         a_sbp, b_sbp_z),
     ...)
\end{verbatim}
\end{codebox}

Simplified, the equation can be viewed as
\[
\hat y \approx 
  w_{\text{waist}}\;\mathrm{gate}(\text{Waist}) +
  w_{\text{bmi}}\;\mathrm{gate}(\text{BMI}) +
  w_{\text{fg}}\;\mathrm{gate}(\text{Fasting glucose}) +
  w_{\text{sbp}}\;\mathrm{gate}(\text{SBP}) + \cdots.
\]

The corresponding thresholds, in natural units, connect directly to metabolic‑syndrome criteria:

\begin{itemize}
  \item \textbf{Waist circumference:} gate around $93$–$94$\,cm, essentially matching the IDF central‑obesity cut‑off of 88--94\,cm depending on sex. This gate behaves as a guideline‑level threshold.
  \item \textbf{BMI:} gate near $27$\,kg/m$^2$, slightly above the 25\,kg/m$^2$ “overweight” line and close to the “pre‑obesity” range used in some guidelines.
  \item \textbf{Fasting glucose:} gate around $99$\,mg/dL, very close to the 100\,mg/dL impaired‑fasting threshold. This can be interpreted as a conservative screening gate: risk increases once glucose is in the upper‑normal range.
  \item \textbf{Systolic blood pressure:} gate around $128$–$130$\,mmHg, aligning with stage‑1 hypertension thresholds (ACC/AHA 2017).
\end{itemize}

Thus, the NHANES top‑1 model is nearly a sum of binary rules that a clinician might state in words:
``Risk is high when the patient is centrally obese, overweight, has SBP $\gtrsim 130$\,mmHg, and fasting glucose at or above the impaired‑fasting range.”

Because LGO exposes both the algebra and the thresholds in cm, kg/m$^2$, mmHg and mg/dL, it is straightforward to check each gate against guideline tables and to decide whether it represents a diagnostic cutoff or an earlier ``heads‑up” threshold.

\subsection{Lightweight top-\(k\) audit}
\label{sec:si-topk-audit}

To complement the top‑1 exemplars, we summarize the top‑\(k\) candidate pool per dataset (default \(k=100\)) using the aggregated CSV files shipped with the artifact.  Here we focus on the three clinically oriented cohorts (ICU, eICU, NHANES), where gates have the clearest mechanistic interpretation.

\begin{itemize}
\item \textbf{Gate selectivity and sparsity.}
We report (i) the fraction of models that contain at least one gate (\% with gates) and (ii) the median number of gates per model (zeros included).  Table~\ref{tab:gate-usage} consolidates these statistics for ICU, eICU and NHANES.  The pattern mirrors the main text: LGO\textsubscript{hard} keeps \emph{high} gate usage where thresholding is clinically warranted (ICU, eICU, NHANES) but with many fewer gates per model than LGO\textsubscript{soft}, which tends to gate almost every feature.  In other words, hard gates concentrate capacity on a small set of clinically meaningful “switches”.

\item \textbf{Feature‑wise gate incidence and threshold agreement.}
We list, for each cohort, the most frequently gated features (Top‑3 to Top‑5), the median threshold in natural units (with IQR), and the share of models whose gates fall within \(\le10\%\) / \(\le20\%\) of guideline anchors.  Table~\ref{tab:icu-topk} summarizes ICU and eICU side‑by‑side; Table~\ref{tab:nhanes-topk} summarizes NHANES.  No additional plotting is required; these tables are a lightweight, reproducible view into “what the gates are watching”.
\end{itemize}

Across the three health datasets, LGO\textsubscript{hard} exhibits \emph{selective} gate usage and a \emph{sparser} switching structure than LGO\textsubscript{soft}.  
On ICU and eICU, the most common gates sit on variables that correspond to recognisable clinical states: shock‑level or markedly elevated lactate, impaired renal function (creatinine), low MAP, the use of vasopressors and mechanical ventilation, depressed GCS, tachypnea and very low SpO\textsubscript{2}.  
On NHANES, gates cluster on anthropometric and metabolic variables that define cardiometabolic risk (waist circumference/BMI, systolic blood pressure, fasting glucose, triglycerides), effectively carving out obesity, hypertension and hyperglycaemia regimes.  
The thresholds themselves are mapped back to natural units and compared against guideline values, so each gate can be read as “this model treats lactate \(\ge\) shock‑range” or “waist circumference in the central‑obesity range”, rather than as an opaque change in a continuous score.  
These pool‑level trends corroborate the top‑1 expressions and the main‑text threshold audit.

\begin{table}[h!]
\centering
\caption{Top-\(k\) gate usage and sparsity on the three primary health datasets (ICU, eICU, NHANES; \(k=100\)).  For each experiment, we report the fraction of top‑\(k\) models that contain at least one gate (\% with gates), the median number of gates per model (zeros included), and the median cross‑validation loss and symbolic complexity.  LGO\textsubscript{hard} keeps gate usage high where thresholding is useful while using substantially fewer gates than LGO\textsubscript{soft}. Related to Table~7 and Figure~5.}
\label{tab:gate-usage}
\small
\begin{tabularx}{0.95\textwidth}{@{}l l c R R R R@{}}
\toprule
Dataset & Experiment & Top-\(k\) & Gate usage (\%) & Median \# gates & Complexity (median) & CV loss (median) \\
\midrule
ICU     & base       & 100 &   0.0 &  0.0 & 36.00 & 0.6447 \\
ICU     & lgo\_hard  & 100 & 100.0 &  2.0 & 45.50 & 0.5326 \\
ICU     & lgo\_soft  & 100 & 100.0 & 11.0 & 82.30 & 1.0376 \\
eICU    & base       & 100 &   0.0 &  0.0 & 50.00 & 2.3961 \\
eICU    & lgo\_hard  & 100 & 100.0 &  4.5 & 60.00 & 2.5654 \\
eICU    & lgo\_soft  & 100 & 100.0 & 22.0 &130.35 & 2.4859 \\
NHANES  & base       & 100 &   0.0 &  0.0 & 36.00 & 0.5941 \\
NHANES  & lgo\_hard  & 100 & 100.0 &  5.0 & 53.00 & 0.4583 \\
NHANES  & lgo\_soft  & 100 & 100.0 & 12.5 & 72.15 & 0.7449 \\
\bottomrule
\end{tabularx}
\end{table}

\begin{table}[h!]
\centering
\caption{ICU and eICU (Top-$k$) most frequently gated features with unit-aware thresholds (natural units). For each feature we report how often it is gated in the top-$k$ candidate pool for LGO\textsubscript{hard} (Gate cnt, number of models $N$, and percentage), together with the median gate threshold and interquartile range in natural units, and the gate type. On ICU, gates concentrate on shock- and organ-failure–relevant variables (lactate, creatinine, MAP, vasopressor use); on eICU they additionally highlight ventilation status and hypoxemia (mechanical ventilation, SpO$_2$). These patterns match clinical expectations about which variables define high-risk regimes. Related to Table~4 and Figure~3.}
\label{tab:icu-topk}
\small
\setlength{\tabcolsep}{4pt}
\begin{tabularx}{0.95\textwidth}{@{}l L l r r r r r r l@{}}
\toprule
{Dataset} & {Feature} & {Unit} & {Gate} & \multicolumn{2}{c}{{Models w/ gate}} & \multicolumn{3}{c}{{Threshold}} & {Gate} \\
\cmidrule(lr){5-6} \cmidrule(lr){7-9}
& & & {cnt} & {N} & {\%} & {Median} & {Q1} & {Q3} & {type} \\
\midrule
ICU   & creatinine\_mg\_dl        & mg/dL     & 44 & 100 & 44 & 1.71  & 1.54  & 4.86  & lgo\_thre \\
ICU   & lactate\_mmol\_l          & mmol/L    & 52 & 100 & 52 & 4.72  & 2.23  & 5.31  & lgo\_thre \\
ICU   & map\_mmhg                 & mmHg      & 11 & 100 & 11 & 69.95 & 69.08 & 69.98 & lgo\_thre \\
ICU   & vasopressor\_use\_std     & (std)     & 22 & 100 & 22 & 0.59  & 0.59  & 0.59  & lgo\_thre \\
eICU  & lactate\_mmol\_l          & mmol/L    & 77 & 100 & 77 & 2.52  & 1.07  & 3.44  & lgo\_thre \\
eICU  & mechanical\_ventilation\_std & (std)   & 57 & 100 & 57 & 0.52  & 0.48  & 0.60  & lgo\_thre \\
eICU  & vasopressor\_use\_std     & (std)     & 52 & 100 & 52 & 0.40  & -0.18 & 0.40  & lgo\_thre \\
eICU  & creatinine\_mg\_dl        & mg/dL     & 25 & 100 & 25 & 1.32  & 0.98  & 1.67  & lgo\_thre \\
eICU  & spo2\_min                 & \%        & 20 & 100 & 20 & 76.00 & 75.67 & 76.59 & lgo\_thre \\
\bottomrule
\end{tabularx}
\end{table}

Notably, the learned gating structures revealed institution-specific clinical workflows: MIMIC-IV ICU models preferentially gated on shock-level lactate and conservative MAP thresholds indicative of hemodynamic surveillance, whereas eICU models exhibited earlier lactate activation, profoundly hypoxemic SpO\textsubscript{2} cutoffs, and recurrent gating on mechanical ventilation and vasopressor status---collectively reflecting the intervention-intensive dynamics of acute resuscitation.

\begin{table}[h!]
\centering
\caption{NHANES (Top-\(k\)) most frequently gated features with unit‑aware thresholds (natural units).  Gates fall on classic cardiometabolic risk factors: central obesity (waist circumference), elevated systolic blood pressure, impaired fasting glucose, low HDL, and related demographic covariates.  Thresholds are expressed directly in cm, mmHg, mg/dL, etc., so each gate can be read as an explicit rule such as ``waist circumference in the abdominal‑obesity range” or ``fasting glucose in the pre‑diabetes range”. Related to Table~4 and Figure~3.}
\label{tab:nhanes-topk}
\small
\setlength{\tabcolsep}{4pt}
\begin{tabularx}{0.95\textwidth}{@{}l L l r r r r r r l@{}}
\toprule
{Dataset} &  {Feature} &  {Unit} &  {Gate} & \multicolumn{2}{c}{ {Models w/ gate}} & \multicolumn{3}{c}{ {Threshold}} &  {Gate} \\
\cmidrule(lr){5-6} \cmidrule(lr){7-9}
& & &  {cnt} &  {N} &  {\%} &  {Median} &  {Q1} &  {Q3} &  {type} \\
\midrule
NHANES & waist\_circumference & cm    & 31 & 31 & 100 & 93.94  & 91.67  & 98.02  & lgo\_thre \\
NHANES & systolic\_bp         & mmHg  & 28 & 40 & 100 & 128.34 & 128.34 & 128.89 & lgo\_thre \\
NHANES & fasting\_glucose     & mg/dL & 24 & 37 & 100 & 85.45  & 74.54  & 95.40  & lgo\_thre \\
NHANES & gender\_std          & (std) & 21 & 20 & 100 & 0.27   & 0.27   & 0.27   & lgo\_thre \\
NHANES & hdl\_cholesterol     & mg/dL &  3 &  3 & 100 & 39.65  & 39.65  & 39.65  & lgo\_thre \\
\bottomrule
\end{tabularx}
\end{table}

\clearpage
\section{Anchors and unit metadata}
\label{sec:si-anchors}

\textbf{Purpose.} 
To support \emph{unit‑aware audit} across all datasets, we curate anchors (reference cut‑points) and units in a single YAML file, \texttt{config/guidelines.yaml}, contains entries for the three main health datasets (ICU, eICU, NHANES) as well as the UCI benchmarks.  Anchors allow us to compare learned gates against domain references such as MAP 65\,mmHg, lactate 2.0\,mmol/L, GCS~8, SpO\textsubscript{2}~92\%, or fasting glucose 100/126\,mg/dL. Representative sources include Surviving Sepsis Campaign guidance for ICU/eICU anchors and ACC/AHA / ADA / IDF criteria for cardiometabolic anchors in NHANES.  Exact references are listed in the main text.

\textbf{Schema.}
Entries are grouped by dataset name, with an optional \texttt{global} section for anchors reused across datasets.  Each feature entry binds a processed feature name to a unit string, anchor value, and an optional note:

\begin{codebox}[]
\begin{verbatim}
datasets:
  ICU_composite_risk_score:
    map_mmhg:
      unit: "mmHg"
      anchor: 65.0
      note: "MAP target (Surviving Sepsis Campaign)"
    lactate_mmol_l:
      unit: "mmol/L"
      anchor: 2.0
      note: "Hyperlactatemia threshold"
  eICU_composite_risk_score:
    gcs_min:
      unit: "score"
      anchor: 8.0
      note: "Severe neurologic impairment (GCS $\le$ 8)"
    spo2_min:
      unit: "%"
      anchor: 92.0
      note: "Hypoxemia alarm threshold"
  NHANES_metabolic_score:
    systolic_bp:
      unit: "mmHg"
      anchor: 130.0
      note: "ACC/AHA elevated BP"
    fasting_glucose:
      unit: "mg/dL"
      anchor: 100.0
      note: "ADA impaired fasting glucose"
\end{verbatim}
\end{codebox}

Unit strings follow the conventions used in the main text (e.g., mmHg, mmol/L, mg/dL, cm).  Standardized indicators (e.g., \texttt{gender\_std}) are documented with \texttt{"(std)"} as unit and a short note explaining the coding.

\textbf{Curation and provenance.}
For ICU and eICU, anchors derive from critical‑care practice (hemodynamic targets, neurologic cut‑points, hypoxemia alarms, renal dysfunction thresholds). Examples include MAP 65\,mmHg, GCS~8, SpO\textsubscript{2}~92\%, respiratory rate 24\,min$^{-1}$, creatinine 1.2–1.5\,mg/dL, lactate 2.0\,mmol/L.
For NHANES, anchors follow ACC/AHA blood‑pressure guidance, ADA glucose criteria and IDF/ATP‑III definitions of central obesity and dyslipidemia (e.g., waist circumference, HDL, triglycerides).  UCI benchmarks use anchors only where a stable, dataset‑independent reference exists (e.g., blood‑pressure levels, obvious physical ranges); we deliberately \emph{avoid} constant anchors for categorical codes, counts, or clearly age‑indexed normals (e.g., some Cleveland features), and mark those omissions in the YAML notes.

\textbf{Integration in audit.}
The threshold inversion and traffic-light deviation calculation are detailed in STAR Methods (``Threshold recovery and audit''). Here we document the anchor 
curation decisions and coverage validation.

\textbf{Coverage and maintenance.}
We validate that every anchor in \texttt{guidelines.yaml} refers to an existing feature in the processed dataset and emit a coverage report listing (i) anchored features and (ii) present but unanchored features.  Anchors are versioned within the repository; updates require a short provenance note and citation in the YAML.  This makes explicit when a guideline change (e.g., updated hypertension stages) is responsible for a shift in the audit, rather than a change in the learned model.

\textbf{Limitations.}
Anchors encode \emph{reference} cut‑points, which may differ from \emph{operational} alerting thresholds or locally preferred practice.  For example, ICU respiratory‑rate alarms may be set around 24\,min$^{-1}$, while clinical concern can start earlier; GCS~8 marks severe impairment, but clinicians often respond to more subtle deteriorations.  We therefore always present both the anchor and the learned threshold distribution (median/IQR), and we interpret yellow/red cells either as (i) intentionally conservative or extreme‑risk gates (e.g., shock‑range lactate, very low SpO\textsubscript{2}, high BMI/waist), or as a signal that the underlying anchor is itself context‑dependent (e.g., age‑indexed normals in Cleveland).  These judgments are discussed explicitly in the main‑text threshold audit and case‑study sections. 

\clearpage
\section{Pareto and complexity}
\label{sec:si-pareto}

\textbf{Definitions.}
\emph{Complexity} counts typed primitive nodes (operators, variables, constants) in the final expression.  \emph{CV loss} denotes the shared cross‑validation proxy used during search (folds, warm‑up, and subsampling exactly as in the main configuration).  For each method and dataset we form Pareto fronts in the two‑dimensional space (CV loss, complexity).

\begin{figure}[h!]
  \centering
  \includegraphics[width=\textwidth]{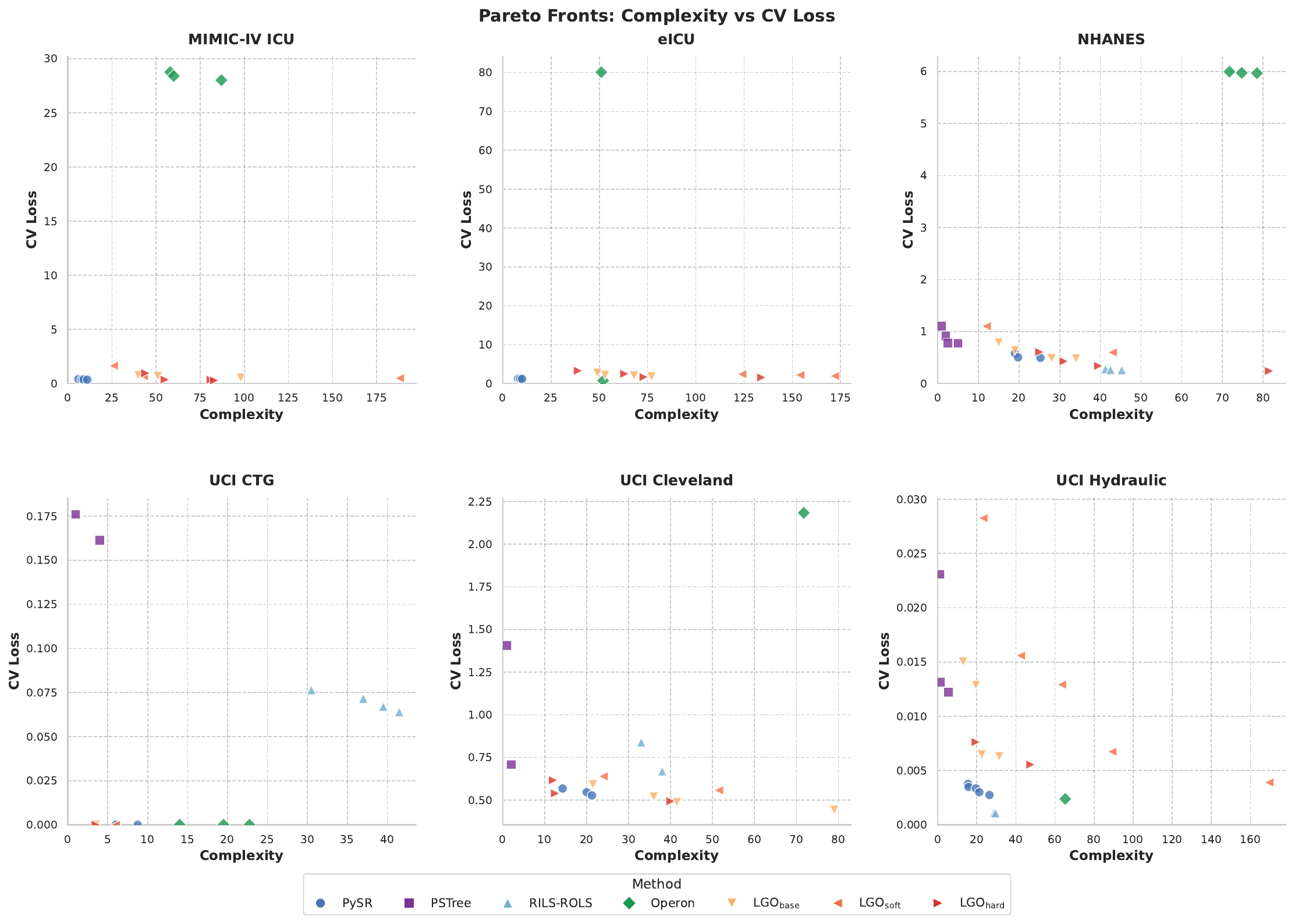}
  \caption{Pareto fronts (CV loss vs.\ symbolic complexity) for all six datasets.  Each panel shows non‑dominated candidates aggregated across seeds; colors denote methods.  On ICU, eICU, and NHANES, LGO\textsubscript{hard} and LGO\textsubscript{base} occupy the low‑loss region at intermediate complexity, whereas LGO\textsubscript{soft} tends to be more complex for similar loss and Operon occasionally produces high‑loss outliers.  On CTG, all engines reach near‑zero CV loss and essentially share the Pareto front.  On Cleveland, LGO\textsubscript{hard} attains the lowest CV losses at higher complexity while PySR provides lower‑complexity alternatives.  On Hydraulic, smooth relations favour PySR and RILS‑ROLS; LGO variants either prune gates and align with these baselines or appear as higher‑complexity points with slightly worse loss. Related to Figure~2.}
  \label{fig:pareto}
\end{figure}

\textbf{Protocol.}
For every dataset (ICU, eICU, NHANES, CTG, Cleveland, Hydraulic) and every method, we collect the top‑$k$ candidates ($k=100$) by the engine’s internal score, recompute CV loss with the shared proxy, and then discard dominated points.  The surviving points form a combined Pareto front across methods.  In Figure~\ref{fig:pareto}, marker color encodes the method and the axes show complexity and CV loss; all methods are evaluated under the same CV proxy and complexity definition.

\textbf{Reading guide.}
Points toward the lower‑left corner (low CV loss at low complexity) represent more favorable accuracy–simplicity trade‑offs.

On the three \emph{health} datasets (ICU, eICU, NHANES), LGO\textsubscript{hard} and LGO\textsubscript{base} populate the low‑loss region at \emph{intermediate} complexities: they sit between very simple but higher‑loss PSTree models and more complex arithmetic baselines, while LGO\textsubscript{soft} tends to produce longer expressions for comparable loss.  In particular, on NHANES, hard‑gated models achieve near‑front CV losses at substantially lower complexity than the outlying high‑loss, high‑complexity Operon candidates.

On CTG, all methods reach near‑zero CV loss and the fronts are essentially saturated, reflecting the near‑separable nature of the task.  On Cleveland, LGO\textsubscript{hard} extends the front to the lowest CV losses at moderate–high complexity, complementing PySR points that offer slightly higher loss at lower complexity.  On Hydraulic, relations are predominantly smooth: PySR and RILS‑ROLS trace the leading front at low complexity, while LGO variants either prune gates and behave like their arithmetic base or occupy higher‑complexity points with slightly worse loss.

Overall, these fronts confirm the picture from the main text: LGO\textsubscript{hard} does not chase the extreme low‑loss / very‑high‑complexity corner, but instead offers a compact region of models that trade a small amount of accuracy for sparse, gate‑based structure and auditable thresholds.

\clearpage
\section{UCI benchmarks}
\label{sec:si-uci}

\textbf{Task characterization.}
The three UCI benchmarks probe different regimes.  CTG (NSPbin) is a nearly linearly separable binary task based on histogram features.  Cleveland (num) is a moderate‑signal regression problem with mixed clinical covariates and a relatively small sample size.  HydraulicSys fault score is a regression task dominated by smooth sensor–response relationships.

\textbf{Summary of results.}
On \textbf{CTG}, multiple engines (PySR, Operon, all LGO variants) saturate AUROC and AUPRC at~$\approx 1.0$ with identical Brier scores (Table~\ref{tab:uci-ctg}), confirming near‑perfect separability under simple formulas.  In this regime, LGO behaves like an arithmetic SR method: it discovers compact expressions and uses very few gates, consistent with the fact that almost any monotone score suffices.

On \textbf{Cleveland}, LGO\textsubscript{hard} is competitive with the strongest SR baselines and attains the best mean $R^2$ ($0.48\pm0.12$), slightly outperforming Operon ($0.48\pm0.11$) and PySR ($0.42\pm0.11$) under the shared budget (Table~\ref{tab:uci-cleveland}).  The corresponding expressions place gates on clinically interpretable quantities (e.g., resting blood pressure, ST‑segment depression, exercise‑induced angina), yielding audit‑ready cut‑points, while purely arithmetic baselines encode similar information through smoother nonlinearities.

On \textbf{Hydraulic}, the underlying relations are predominantly smooth.  As expected, RILS‑ROLS and PySR dominate the Pareto front with high $R^2$ (0.95 and 0.74 on average) and low complexity (Table~\ref{tab:uci-hydraulicsys-fault-score}).  LGO\textsubscript{base} and LGO\textsubscript{hard} reach moderate performance but generally underperform smooth baselines; when gates appear, they focus on a small number of physical variables and can still be interpreted as operating‑range limits, but they are not required for pure predictive accuracy in this system.

Taken together, the UCI benchmarks reinforce the main message from the clinical datasets: LGO\textsubscript{hard} is most beneficial when regime switching and auditability are important (Cleveland), and it gracefully reduces to arithmetic behaviour on near‑separable (CTG) or globally smooth (Hydraulic) problems where thresholds are not the primary mechanism.

\begin{table}[h!]
\centering
\begin{threeparttable}
\caption{UCI CTG NSPbin: mean $\pm$ std (10 seeds). Related to Figure~2.}
\label{tab:uci-ctg}
\small
\begin{tabularx}{0.65\textwidth}{@{}llXXX@{}}
\toprule
method & experiment & AUROC$\uparrow$ & AUPRC$\uparrow$ & Brier$\downarrow$ \\
\midrule
PySR       & base                      & $1.000 \pm 0.000$ & $1.000 \pm 0.000$ & $0.211 \pm 0.000$\\
PSTree     & base                      & $0.766 \pm 0.014$ & $0.526 \pm 0.035$ & $0.386 \pm 0.018$\\
RILS-ROLS  & base                      & $0.517 \pm 0.135$ & $0.288 \pm 0.110$ & $0.333 \pm 0.164$\\
Operon     & base                      & $1.000 \pm 0.000$ & $1.000 \pm 0.000$ & $0.211 \pm 0.000$\\
LGO        & base                      & $1.000 \pm 0.000$ & $1.000 \pm 0.000$ & $0.211 \pm 0.000$\\
LGO        & LGO\textsubscript{soft}   & $1.000 \pm 0.000$ & $1.000 \pm 0.000$ & $0.212 \pm 0.002$\\
LGO        & LGO\textsubscript{hard}   & $1.000 \pm 0.000$ & $1.000 \pm 0.000$ & $0.211 \pm 0.000$\\
\bottomrule
\end{tabularx}
\end{threeparttable}
\end{table}

\begin{table}[h!]
\centering
\begin{threeparttable}
\caption{UCI Heart Cleveland num: mean $\pm$ std (10 seeds). Related to Figure~2.}
\label{tab:uci-cleveland}
\small
\begin{tabularx}{0.65\textwidth}{@{}llXXX@{}}
\toprule
method & experiment & R$^2\uparrow$ & RMSE$\downarrow$ & MAE$\downarrow$ \\
\midrule
PySR       & base                      & $0.416 \pm 0.105$ & $0.909 \pm 0.085$ & $0.638 \pm 0.070$\\
PSTree     & base                      & $0.075 \pm 0.275$ & $1.132 \pm 0.122$ & $0.711 \pm 0.096$\\
RILS-ROLS  & base                      & $0.392 \pm 0.127$ & $0.928 \pm 0.109$ & $0.645 \pm 0.071$\\
Operon     & base                      & $0.480 \pm 0.107$ & $0.856 \pm 0.075$ & $0.600 \pm 0.059$\\
LGO        & base                      & $0.436 \pm 0.112$ & $0.892 \pm 0.084$ & $0.574 \pm 0.067$\\
LGO        & LGO\textsubscript{soft}   & $0.397 \pm 0.177$ & $0.914 \pm 0.108$ & $0.583 \pm 0.076$\\
LGO        & LGO\textsubscript{hard}   & $0.484 \pm 0.118$ & $0.852 \pm 0.090$ & $0.531 \pm 0.061$\\
\bottomrule
\end{tabularx}
\end{threeparttable}
\end{table}

\begin{table}[h!]
\centering
\begin{threeparttable}
\caption{UCI HydraulicSys fault score: mean $\pm$ std (10 seeds). Related to Figure~2.}
\label{tab:uci-hydraulicsys-fault-score}
\small
\begin{tabularx}{0.65\textwidth}{@{}llXXX@{}}
\toprule
method & experiment & R$^{2}\uparrow$ & RMSE$\downarrow$ & MAE$\downarrow$ \\
\midrule
PySR      & base & $0.738 \pm 0.031$ & $0.089 \pm 0.004$ & $0.069 \pm 0.004$\\
PSTree    & base & $0.538 \pm 0.064$ & $0.116 \pm 0.014$ & $0.090 \pm 0.013$\\
RILS-ROLS & base & $0.946 \pm 0.007$ & $0.040 \pm 0.004$ & $0.028 \pm 0.002$\\
Operon$^{*}$ & base & $0.822 \pm 0.049$ & $0.072 \pm 0.010$ & $0.055 \pm 0.008$\\
LGO       & base & $0.609 \pm 0.119$ & $0.106 \pm 0.017$ & $0.084 \pm 0.015$\\
LGO       & LGO\textsubscript{soft} & $0.397 \pm 0.215$ & $0.131 \pm 0.024$ & $0.106 \pm 0.021$\\
LGO       & LGO\textsubscript{hard} & $0.364 \pm 0.248$ & $0.134 \pm 0.027$ & $0.107 \pm 0.023$\\
\bottomrule
\end{tabularx}
\begin{tablenotes}
\footnotesize
\item Notes: Two seeds excluded due to implausible R$^{2}$ values ($< -1$).
\end{tablenotes}
\end{threeparttable}
\end{table}

\clearpage
\section{Case studies: single-feature sanity checks (extended plots)}

To complement the aggregate threshold audit in the main text (Figure~3), we provide extended single-feature sanity checks for seven representative feature--dataset pairs: ICU lactate, eICU GCS, NHANES waist circumference and triglycerides, UCI CTG MeanHist, UCI Cleveland resting systolic blood pressure, and UCI Hydraulic System PS1 pressure. For each case we plot (A) the feature distribution stratified by the composite-risk (or fault / abnormality) label with guideline and LGO thresholds overlaid, and (B) the ROC curve obtained by using the raw feature as a one-dimensional score, with guideline and LGO operating points marked. These plots make explicit how much ``lift'' a single feature provides, and whether LGO gates choose conservative (screening-like) or stringent (diagnostic-like) operating points relative to the anchors.

\begin{figure}[ht]
  \centering
  \includegraphics[width=1.0\linewidth]{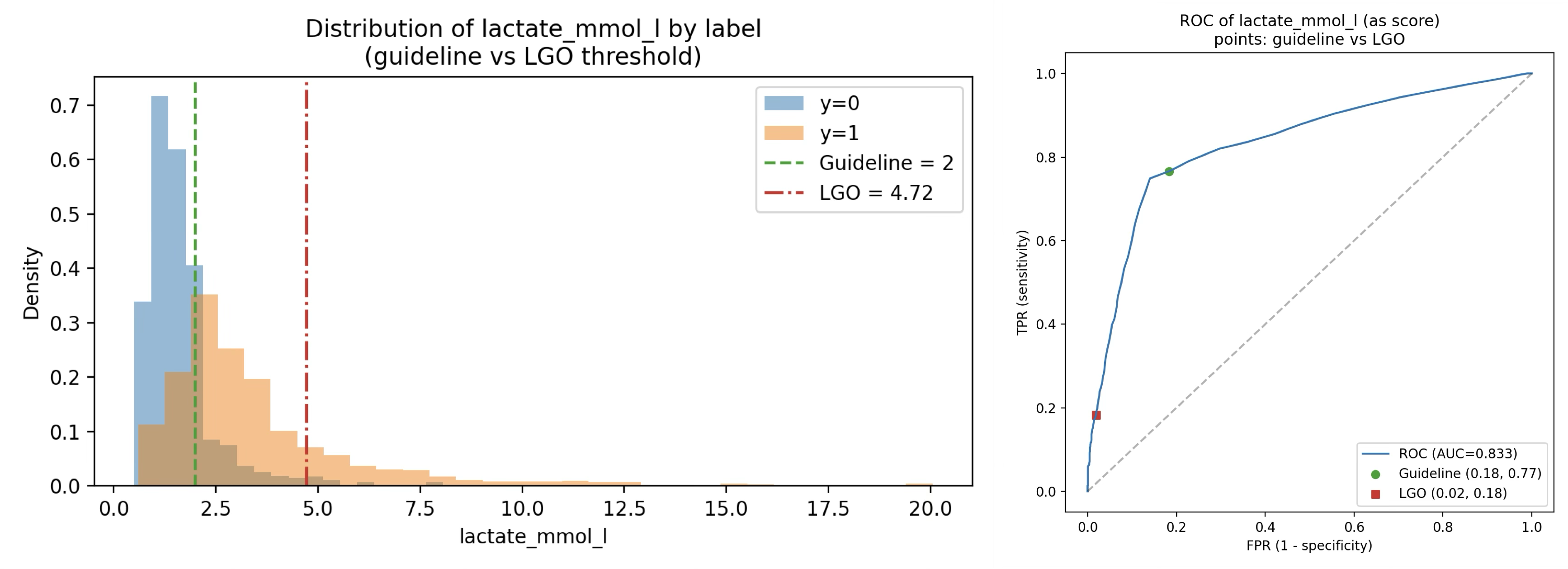}
  \caption{\textbf{ICU lactate (composite risk score $\ge 5$).}
  Left panel: distribution of lactate (mmol/L) for low- vs.\ high-risk patients with vertical lines marking the guideline threshold at 2.0\,mmol/L and the LGO\textsubscript{hard} median threshold (extreme-risk gate) in the higher-lactate tail.
  Right panel: ROC curve using lactate as a continuous score (AUC in the high 0.8 range); circles and squares indicate the guideline and LGO operating points, respectively. The guideline cut-off achieves a balanced trade-off between sensitivity and specificity, whereas the LGO gate sacrifices sensitivity for very high specificity, consistent with an ``escalate only when lactate is clearly high'' rule rather than a screening threshold. Related to Figure~3 and Table~4.}
  \label{fig:si-icu-lactate}
\end{figure}

\begin{figure}[ht]
  \centering
  \includegraphics[width=1.0\linewidth]{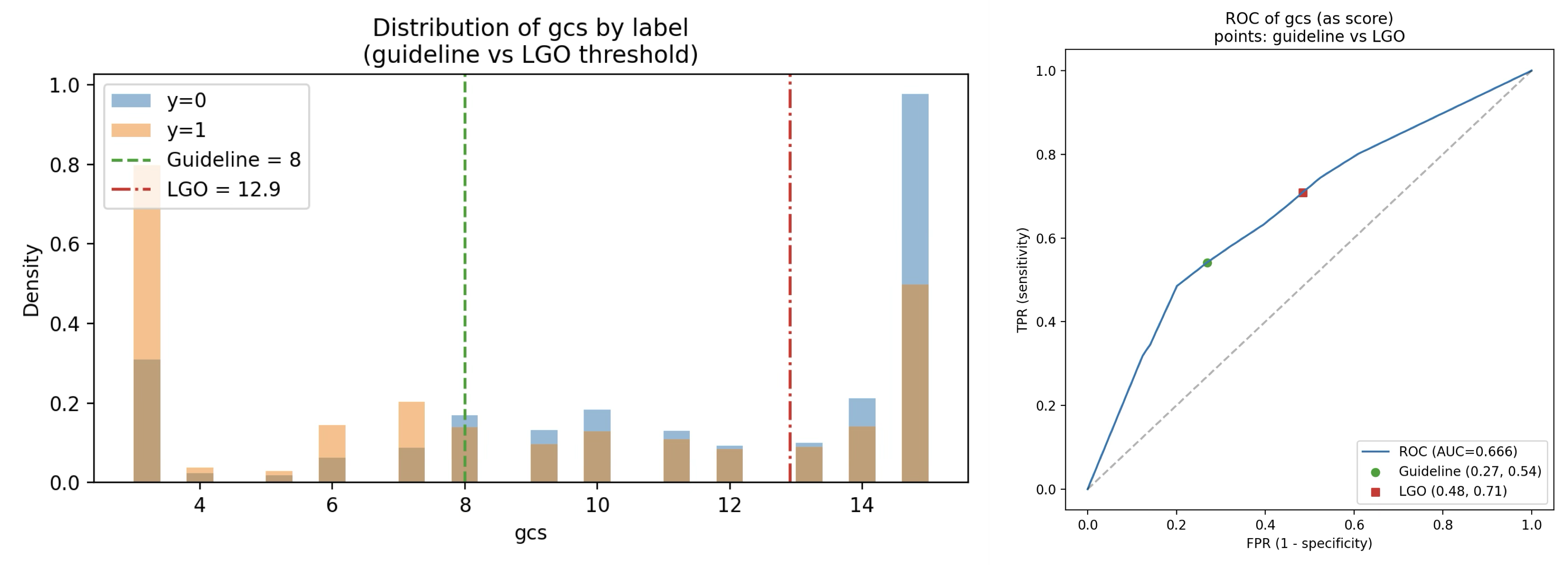}
  \caption{\textbf{eICU GCS (composite risk score $\ge 8$).}
  Left panel: distribution of Glasgow Coma Scale (GCS) scores by risk group; vertical lines show the severe-impairment anchor at GCS~$\le 8$ and the LGO\textsubscript{hard} median gate around the low-teens.
  Right panel: ROC curve for GCS (moderate AUC). The guideline cut-off operates at a relatively specific but less sensitive point, whereas the LGO gate at $\sim 13$ increases sensitivity at the cost of specificity. This pattern is consistent with a ward-level early-warning use case where ``any noticeable drop in consciousness'' should trigger attention, even before the patient reaches the classical GCS~$\le 8$ boundary. Related to Figure~3 and Table~4.}
  \label{fig:si-eicu-gcs}
\end{figure}

\begin{figure}[ht]
  \centering
  \includegraphics[width=1.0\linewidth]{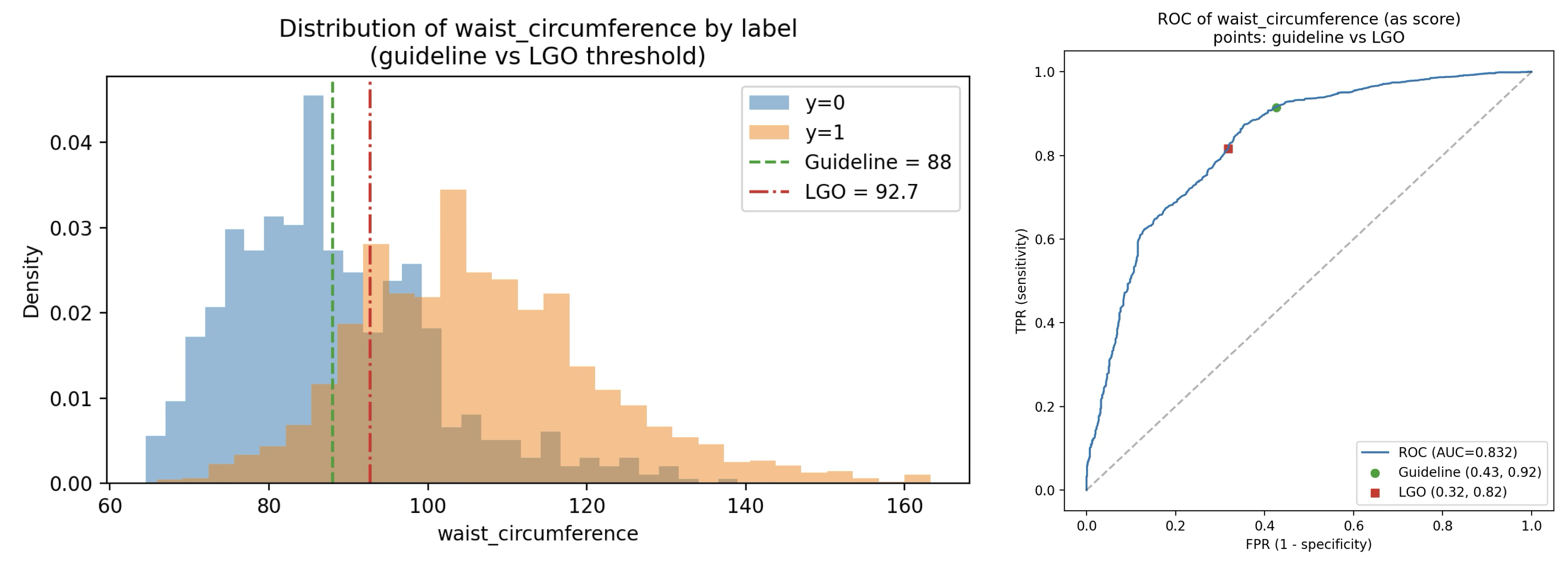}
  \caption{\textbf{NHANES waist circumference (metabolic score $\ge 2$).}
  Left panel: distribution of waist circumference (cm) by metabolic-risk status; vertical lines mark the International Diabetes Federation anchor at 88\,cm and the LGO\textsubscript{hard} median threshold in the low-90s.
  Right panel: ROC curve using waist circumference as a score (good AUC). Both the anchor and the LGO gate sit near the steep part of the curve, and their operating points have similar balanced accuracies, indicating that data-driven and guideline cut-points broadly agree for central adiposity. Related to Figure~3 and Table~4.}
  \label{fig:si-nhanes-waist}
\end{figure}

\begin{figure}[ht]
  \centering
  \includegraphics[width=1.0\linewidth]{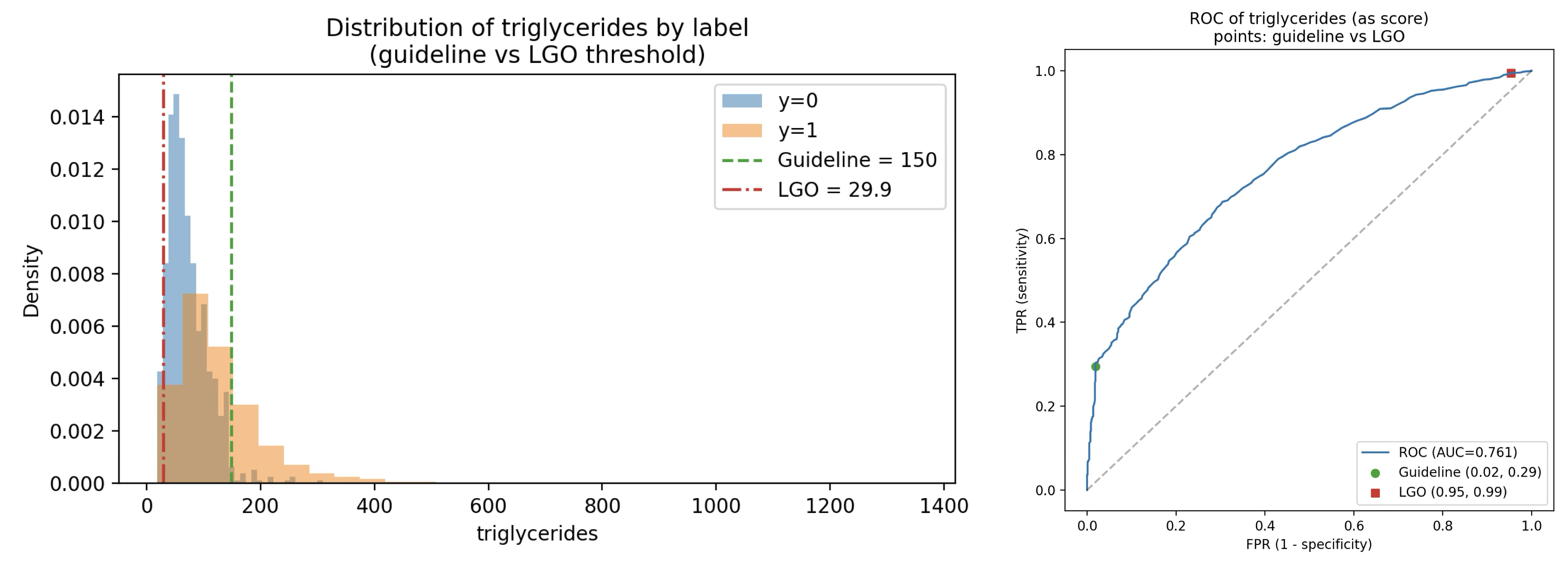}
  \caption{\textbf{NHANES triglycerides (metabolic score $\ge 2$).}
  Left panel: distribution of triglycerides (mg/dL) in low- vs.\ high-risk participants with the 150\,mg/dL hypertriglyceridemia anchor and the LGO\textsubscript{hard} gate overlaid.
  Right panel: ROC curve (moderate AUC). The guideline cut-off sits slightly to the right of the LGO gate: LGO tends to pick a somewhat lower triglyceride threshold, trading a small loss in specificity for higher sensitivity. This again matches a screening interpretation---flag individuals with emerging lipid risk before they cross the full diagnostic boundary. Related to Figure~3 and Table~4.}
  \label{fig:si-nhanes-tg}
\end{figure}

\begin{figure}[ht]
  \centering
  \includegraphics[width=1.0\linewidth]{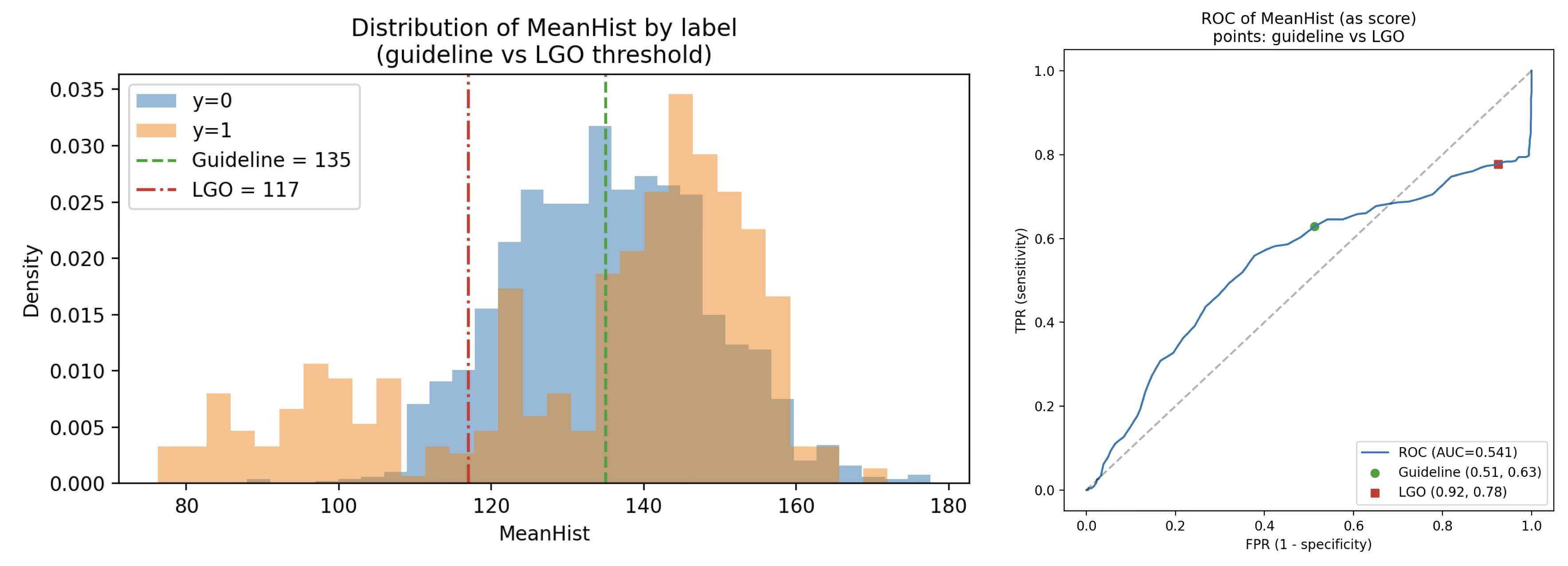}
  \caption{\textbf{UCI CTG (NSP binary) -- MeanHist.}
  Left panel: distribution of the histogram mean of fetal heart rate (MeanHist) in normal vs.\ abnormal traces. The anchor is derived from the curated guideline table (around 135\,bpm), with the LGO\textsubscript{hard} gate lying close to this region.
  Right panel: ROC curve for MeanHist (high AUC, reflecting the near-separable nature of CTG). Both the guideline and LGO operating points lie on the plateau of the ROC, where many thresholds achieve ceiling performance. In contrast, analyses of features like \texttt{FM} (fetal movements; plots not shown) highlight that some anchors inherited from ground-truth tables are poorly calibrated (e.g., an anchor at 0), and LGO gates move toward more informative cut-points, effectively flagging issues with anchor design rather than with the model. Related to Figure~3.}
  \label{fig:si-ctg-meanhist}
\end{figure}

\begin{figure}[ht]
  \centering
  \includegraphics[width=1.0\linewidth]{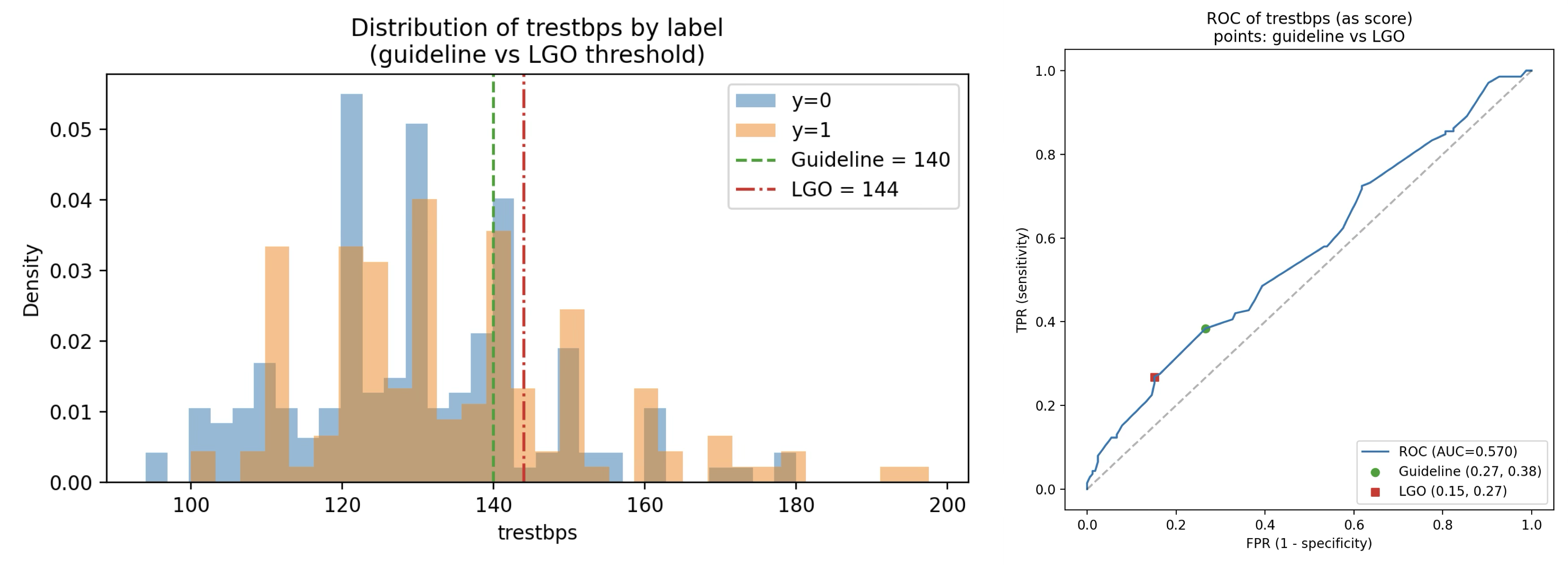}
  \caption{\textbf{UCI Cleveland (fault score / risk) -- resting systolic blood pressure.}
  Left panel: distribution of resting systolic blood pressure (\texttt{trestbps}, mmHg) by outcome. The anchor around 140\,mmHg reflects hypertension guidance; the LGO\textsubscript{hard} gate typically falls in a similar range, sometimes slightly lower.
  Right panel: ROC curve for \texttt{trestbps} (modest but non-trivial AUC). The LGO operating point usually achieves a small gain in balanced accuracy relative to the 140\,mmHg anchor, while staying in a clinically familiar range. By contrast, sanity checks on categorical encodings such as \texttt{cp}, \texttt{slope}, or \texttt{oldpeak} (results not shown) illustrate how coarse coding constrains the possible thresholds: LGO can only place gates on integer-coded levels, so deviations from anchors there can reflect limitations of the feature representation more than a true disagreement with domain knowledge. Related to Figure~3.}
  \label{fig:si-cleveland-trestbps}
\end{figure}

\begin{figure}[ht]
  \centering
  \includegraphics[width=1.0\linewidth]{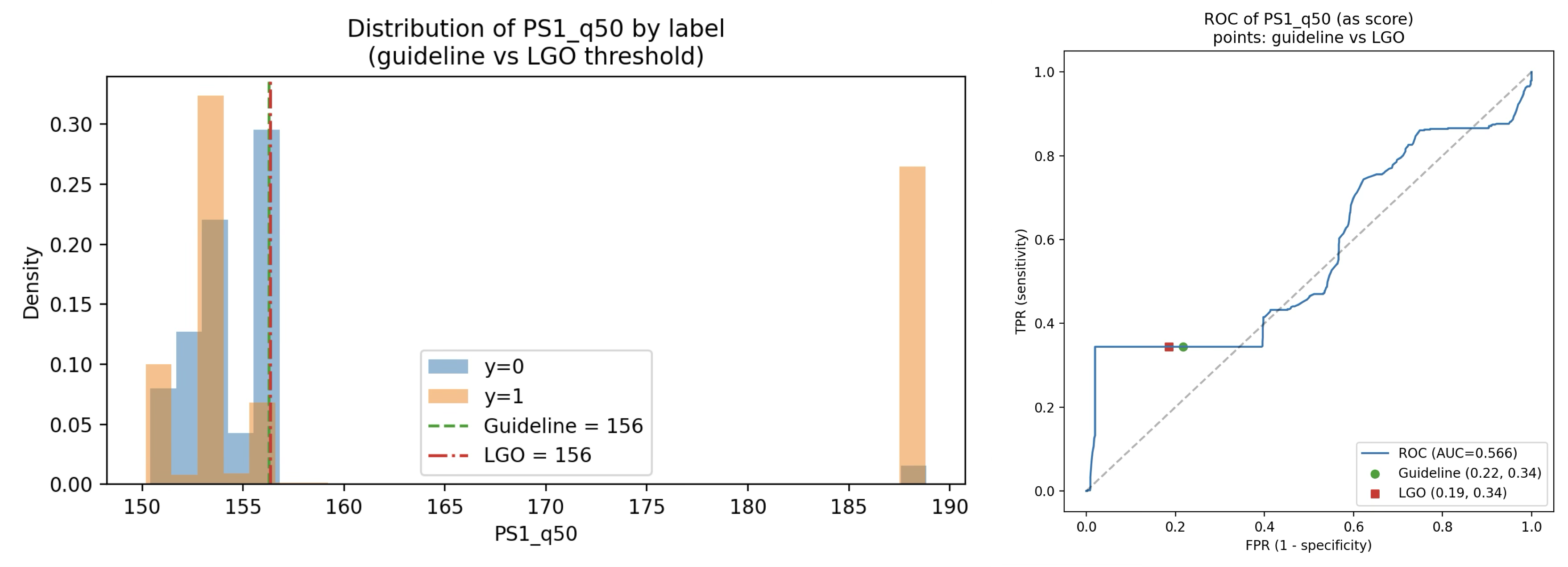}
  \caption{\textbf{UCI Hydraulic System (fault score) -- PS1 pressure.}
  Left panel: distribution of the primary pressure sensor statistic (e.g., \texttt{PS1\_mean} or \texttt{PS1\_q50}, in bar) across low- vs.\ high-fault regimes, with the engineering anchor from the guideline table and the LGO\textsubscript{hard} gate overlaid. The gate typically lands close to the suggested operating envelope, confirming that the learned cut-point is physically plausible.
  Right panel: ROC curve for this pressure feature (AUC in the moderate range). The slope is relatively smooth, and moving from the guideline to the LGO gate yields only marginal changes in TPR/TNR. This is consistent with our main-text observation that Hydraulic is dominated by globally smooth relationships: LGO thresholds align with engineering set-points but add limited predictive gain, underscoring that gates are most beneficial on genuinely regime-switching problems. Related to Figure~3.}
  \label{fig:si-hydraulic-ps1}
\end{figure}

Taken together, these extended plots reinforce the story in the main text. On our primary clinical cohorts (ICU and eICU), LGO gates either recover standard escalation thresholds (MAP, lactate) or intentionally shift toward earlier warning points (GCS, SpO$_2$). On NHANES, LGO gates for waist circumference and triglycerides align with cardiometabolic screening cutoffs. On CTG and Cleveland, sanity checks highlight both agreement with meaningful anchors (MeanHist, \texttt{trestbps}) and cases where legacy anchors or coarse encodings are questionable (e.g., \texttt{FM}, \texttt{oldpeak}/\texttt{slope}). On Hydraulic, LGO gates respect engineering envelopes but have limited impact on smooth sensor--response ROC curves, illustrating that threshold primitives are most informative when the underlying mechanism genuinely involves regime transitions.

\clearpage
\section{Extended comparison with clinical scoring and EBM baselines}

This section provides the full numerical results for the comparisons between LGO\textsubscript{hard}, AutoScore, and Explainable Boosting Machines (EBM) on the three primary health datasets. All experiments reuse the same train–test splits as in the main text (ten random seeds). For ICU and eICU we binarize the composite score at clinically motivated thresholds (ICU: score $\ge 5$; eICU: score $\ge 8$); for NHANES we follow the metabolic‑syndrome convention (score $\ge 2$).

\subsection*{ICU composite risk (binary, score $\ge 5$)}

Table~S\textit{ICU‑A} reports mean $\pm$~std over 10 seeds for AUROC, AUPRC, Brier score, F1, accuracy, and training time, comparing LGO\textsubscript{hard} to AutoScore under a 500k‑evaluation search budget. LGO\textsubscript{hard} achieves a slightly higher AUROC and F1 than AutoScore, with a noticeably lower Brier score (better calibration), at the expense of substantially higher computation time. Table~S\textit{ICU‑B} provides the analogous comparison between LGO\textsubscript{hard} and EBM under a lighter 30k‑evaluation budget, highlighting EBM’s near‑perfect discrimination on this engineered composite endpoint.

\begin{table}[h]
\centering
\small
\caption{ICU composite risk (binary, score $\ge 5$): LGO\textsubscript{hard} vs AutoScore (500k evaluations). Mean $\pm$ std over 10 seeds. Related to Figure~5.}
\label{tab:si-icu-autoscore}
\begin{tabular}{lcccccc}
\toprule
Method & AUROC $\uparrow$ & AUPRC $\uparrow$ & Brier $\downarrow$ & F1 $\uparrow$ & Accuracy $\uparrow$ & Train time [s] $\downarrow$ \\
\midrule
AutoScore      & $0.864 \pm 0.010$ & $0.904 \pm 0.007$ & $0.146 \pm 0.005$ & $0.833 \pm 0.007$ & $0.784 \pm 0.008$ & $0.26 \pm 0.08$ \\
LGO\textsubscript{hard} & $0.884 \pm 0.150$ & $0.905 \pm 0.117$ & $0.088 \pm 0.074$ & $0.901 \pm 0.088$ & $0.878 \pm 0.121$ & $151.1 \pm 59.8$ \\
\bottomrule
\end{tabular}
\end{table}

\begin{table}[h]
\centering
\small
\caption{ICU composite risk (binary, score $\ge 5$): LGO\textsubscript{hard} vs EBM (30k evaluations). Mean $\pm$ std over 10 seeds. Related to Figure~6.}
\label{tab:si-icu-ebm}
\begin{tabular}{lcccccc}
\toprule
Method & AUROC $\uparrow$ & AUPRC $\uparrow$ & Brier $\downarrow$ & F1 $\uparrow$ & Accuracy $\uparrow$ & Train time [s] $\downarrow$ \\
\midrule
EBM           & $0.998 \pm 0.001$ & $0.999 \pm 0.001$ & $0.010 \pm 0.004$ & $0.992 \pm 0.003$ & $0.990 \pm 0.004$ & $1.12 \pm 0.50$ \\
LGO\textsubscript{hard} & $0.843 \pm 0.131$ & $0.874 \pm 0.102$ & $0.198 \pm 0.050$ & $0.795 \pm 0.065$ & $0.685 \pm 0.122$ & $6.28 \pm 1.71$ \\
\bottomrule
\end{tabular}
\end{table}

For robustness, Table~S\textit{ICU‑C} additionally reports medians and interquartile ranges (IQR) for AUROC, AUPRC, Brier, F1, and accuracy for each method. LGO\textsubscript{hard} shows a higher AUROC median than AutoScore (approximately $0.94$ vs.\ $0.87$), confirming that the lower mean is driven by a single degenerate run with AUROC close to 0.5.

\subsection*{eICU composite risk (binary, score $\ge 8$)}

Table~S\textit{eICU} summarizes LGO\textsubscript{hard} and EBM on the eICU composite‑risk endpoint under a 30k‑evaluation budget. EBM again saturates discrimination metrics (AUROC and AUPRC $\approx 1.0$; Brier $\approx 6\times 10^{-4}$; F1 and accuracy $\approx 1.0$), while LGO\textsubscript{hard} achieves high case‑level performance (AUPRC $\approx 0.90$, F1 $\approx 0.95$, accuracy $\approx 0.90$) but more variable AUROC across seeds. As in ICU, these results illustrate that LGO\textsubscript{hard} provides fewer degrees of freedom than EBM but still attains clinically reasonable performance, while yielding explicit gating structure.

\begin{table}[h]
\centering
\small
\caption{eICU composite risk (binary, score $\ge 8$): LGO\textsubscript{hard} vs EBM (30k evaluations). Mean $\pm$ std over 10 seeds. Related to Figure~6.}
\label{tab:si-eicu-ebm}
\begin{tabular}{lcccccc}
\toprule
Method & AUROC $\uparrow$ & AUPRC $\uparrow$ & Brier $\downarrow$ & F1 $\uparrow$ & Accuracy $\uparrow$ & Train time [s] $\downarrow$ \\
\midrule
EBM           & $1.000 \pm 0.000$ & $1.000 \pm 0.000$ & $0.0007 \pm 0.0008$ & $0.9996 \pm 0.0004$ & $0.9993 \pm 0.0008$ & $3.94 \pm 0.78$ \\
LGO\textsubscript{hard} & $0.515 \pm 0.294$ & $0.897 \pm 0.078$ & $0.0921 \pm 0.0041$ & $0.945 \pm 0.000$ & $0.896 \pm 0.001$ & $6.86 \pm 0.87$ \\
\bottomrule
\end{tabular}
\end{table}

\subsection*{NHANES metabolic risk (binary, score $\ge 2$)}

Table~S\textit{NHANES} reports the same set of metrics for the NHANES metabolic score. EBM again yields near‑optimal AUROC and well‑calibrated probabilities, while LGO\textsubscript{hard} attains high overall accuracy but substantially lower AUPRC and F1, reflecting a conservative operating point on this heavily imbalanced outcome. In the main text we therefore focus on NHANES as a case study for threshold recovery and agreement with cardiometabolic anchors, and treat EBM as a performance upper bound rather than a direct competitor.

\begin{table}[h]
\centering
\small
\caption{NHANES metabolic risk (binary, score $\ge 2$): LGO\textsubscript{hard} vs EBM (30k evaluations). Mean $\pm$ std over 10 seeds. Related to Figure~6.}
\label{tab:si-nhanes-ebm}
\begin{tabular}{lcccccc}
\toprule
Method & AUROC $\uparrow$ & AUPRC $\uparrow$ & Brier $\downarrow$ & F1 $\uparrow$ & Accuracy $\uparrow$ & Train time [s] $\downarrow$ \\
\midrule
EBM           & $0.996 \pm 0.003$ & $0.913 \pm 0.054$ & $0.0092 \pm 0.0030$ & $0.788 \pm 0.080$ & $0.988 \pm 0.004$ & $0.82 \pm 0.37$ \\
LGO\textsubscript{hard} & $0.709 \pm 0.221$ & $0.134 \pm 0.125$ & $0.0299 \pm 0.0020$ & $0.011 \pm 0.035$ & $0.968 \pm 0.003$ & $6.93 \pm 1.14$ \\
\bottomrule
\end{tabular}
\end{table}

Together, these extended results reinforce the main‑text message: LGO\textsubscript{hard} is competitive with AutoScore on ICU, and retains clinically reasonable performance on ICU and eICU when compared to a strong EBM upper bound, while providing executable, unit‑aware thresholds and symbolic formulas rather than point tables or per‑feature shape plots.

\subsection*{Baseline coverage.}
\begin{table}[htbp]
\centering
\small
\caption{Baseline Selection for Comparative Evaluation. Related to Discussion.}
\label{tab:baseline-selection}
\begin{tabularx}{1.0\textwidth}{@{}p{4.2cm}p{3.5cm}>{\centering\arraybackslash}p{1.5cm}X@{}}
\toprule
Method Category & Representative Baseline & Included? & Rationale \\
\midrule
Clinical Scoring Systems & AutoScore & \checkmark & Point-based risk scores \\
Additive Models / SET Proxy & EBM (GA\textsuperscript{2}M) & \checkmark & Shape functions with extractable thresholds \\
Integer Scoring (MIP-based) & RiskSLIM & \texttimes & CPLEX dependency, scalability issues \\
Symbolic Regression & PySR, Operon, PSTree, RILS-ROLS & \checkmark & Direct SR comparisons \\
\bottomrule
\end{tabularx}
\end{table}

\clearpage
\section{Formal statistical comparison with AutoScore baselines}

We compared LGO\textsubscript{hard} to AutoScore on the high-risk classification tasks for ICU, eICU and NHANES (score thresholds as in the main text). For each dataset and evaluation budget, we report mean$\pm$std AUROC and AUPRC across 10 resampled train–test splits, the difference in AUROC (LGO--AutoScore), a non-parametric paired Wilcoxon signed-rank test, and effect size (Cohen’s $d$).

\begin{table}[htbp]
\centering
\caption{Performance comparison of LGO\textsubscript{hard} and AutoScore across computational budgets (MIMIC-IV ICU, $n=10$). Related to Figure~4.}
\label{tab:lgo_autoscore_icu}
\small
\begin{threeparttable}
\begin{tabular}{lcccccc}
\toprule
Budget & LGO AUROC & AutoScore AUROC & $\Delta$AUROC [95\% CI] & $p$ & Cohen's $d$ & Win \\
\midrule
30k  & $0.927 \pm 0.064$ & $0.864 \pm 0.009$ & $+0.063$ [0.022, 0.098] & 0.027\tnote{*} & 1.31 & 7/10 \\
100k & $0.910 \pm 0.062$ & $0.864 \pm 0.009$ & $+0.045$ [0.006, 0.083] & 0.084 & 0.96 & 6/10 \\
200k & $0.908 \pm 0.062$ & $0.864 \pm 0.009$ & $+0.044$ [0.005, 0.081] & 0.084 & 0.95 & 6/10 \\
300k & $0.859 \pm 0.137$ & $0.864 \pm 0.009$ & $-0.006$ [$-0.100$, 0.066] & 0.846 & $-0.06$ & 4/10 \\
500k & $0.884 \pm 0.142$ & $0.864 \pm 0.009$ & $+0.019$ [$-0.079$, 0.091] & 0.275 & 0.18 & 6/10 \\
\bottomrule
\end{tabular}
\begin{tablenotes}
\footnotesize
\item $\ast$: $p<0.05$ (Wilcoxon signed-rank test).
\item Win: number of seeds where LGO AUROC $>$ AutoScore AUROC.
\item CI: bootstrap percentile interval (10,000 resamples).
\end{tablenotes}
\end{threeparttable}
\end{table}

\begin{table}[htbp]
\centering
\caption{Performance comparison of LGO\textsubscript{hard} and AutoScore across computational budgets (eICU, $n=10$). Related to Figure~4.}
\label{tab:lgo_autoscore_eicu}
\small
\begin{threeparttable}
\begin{tabular}{lcccccc}
\toprule
Budget & LGO AUROC & AutoScore AUROC & $\Delta$AUROC [95\% CI] & $p$ & Cohen's $d$ & Win \\
\midrule
30k  & $0.880 \pm 0.078$ & $0.798 \pm 0.018$ & $+0.082$ [0.030, 0.125] & 0.020\tnote{*} & 1.37 & 8/10 \\
100k & $0.874 \pm 0.066$ & $0.798 \pm 0.018$ & $+0.076$ [0.028, 0.119] & 0.020\tnote{*} & 1.49 & 9/10 \\
200k & $0.943 \pm 0.019$ & $0.798 \pm 0.018$ & $+0.145$ [0.130, 0.160] & 0.002\tnote{**} & 7.38 & 10/10 \\
300k & $0.851 \pm 0.130$ & $0.798 \pm 0.018$ & $+0.053$ [$-0.039$, 0.117] & 0.105 & 0.54 & 8/10 \\
\bottomrule
\end{tabular}
\begin{tablenotes}
\footnotesize
\item $\ast$: $p<0.05$; $\ast\ast$: $p<0.01$ (Wilcoxon signed-rank test).
\item Win: number of seeds where LGO AUROC $>$ AutoScore AUROC.
\item CI: bootstrap percentile interval (10,000 resamples).
\end{tablenotes}
\end{threeparttable}
\end{table}

\clearpage
\section{Train--test performance and overfitting}
\label{sec:si-train-test}

To quantify generalization and diagnose potential overfitting, we computed each model’s performance on the training splits and compared it with the held‑out test performance.  For the regression tasks we report the mean (±~standard deviation across methods and seeds) of the difference \(\Delta = \text{Train} - \text{Test}\) for each metric.  Negative values indicate that the test error is slightly higher than the training error (a desirable sign of good generalization); positive values indicate the opposite.  As summarized in Table~\ref{tab:si-train-test}, these differences are extremely small (typically \(|\Delta|\lesssim 0.01\)) for ICU, eICU, NHANES and Hydraulic datasets.  The UCI Cleveland dataset shows a slightly larger gap (\(\Delta\mathrm{RMSE}\approx0.08\), \(\Delta\mathrm{MAE}\approx0.05\), \(\Delta R^2\approx0.15\)), which is still modest relative to inter‑method variability.  For the CTG classification task we report differences in AUROC, AUPRC and Brier score in Table~\ref{tab:si-train-test-ctg}; the metrics are virtually identical on train and test splits.

\begin{table}[h]
\centering
\small
\caption{Train--test performance differences ($\Delta$ = Train$-$Test) on regression datasets.  Values are mean $\protect\pm$ std across all methods and seeds. Related to Results (Generalization subsection).}
\label{tab:si-train-test}
\begin{tabular}{lccc}
\toprule
Dataset & $\Delta$RMSE & $\Delta$MAE & $\Delta R^2$ \\
\midrule
ICU        & $-0.01 \pm 0.02$ & $-0.01 \pm 0.02$ & $+0.01 \pm 0.02$ \\
eICU       & $-0.01 \pm 0.02$ & $-0.01 \pm 0.02$ & $+0.00 \pm 0.01$ \\
NHANES     & $-0.01 \pm 0.02$ & $-0.01 \pm 0.02$ & $+0.01 \pm 0.02$ \\
Cleveland  & $+0.08 \pm 0.03$ & $+0.05 \pm 0.02$ & $+0.15 \pm 0.05$ \\
Hydraulic  & $-0.01 \pm 0.01$ & $-0.01 \pm 0.01$ & $+0.00 \pm 0.01$ \\
\bottomrule
\end{tabular}
\end{table}

\begin{table}[h]
\centering
\small
\caption{Train--test performance differences on the CTG classification dataset (mean $\pm$ std across methods and seeds).  The classification metrics match almost exactly on train and test splits, indicating no overfitting. Related to Table~S19.}
\label{tab:si-train-test-ctg}
\begin{tabular}{lccc}
\toprule
Dataset & $\Delta$AUROC & $\Delta$AUPRC & $\Delta$Brier \\
\midrule
CTG & $0.00 \pm 0.00$ & $0.00 \pm 0.00$ & $-0.00 \pm 0.00$ \\
\bottomrule
\end{tabular}
\end{table}
